\documentclass{article}

\usepackage[preprint]{neurips_2026}
\usepackage{xspace}
\usepackage[utf8]{inputenc} 
\usepackage[T1]{fontenc}    
\usepackage{hyperref}       
\usepackage{url}            
\usepackage{booktabs}       
\usepackage{amsfonts}       
\usepackage{nicefrac}       
\usepackage{microtype}      
\usepackage[table]{xcolor}
\usepackage{enumitem}
\usepackage{booktabs}
\usepackage{multirow} 
\usepackage{subcaption}
\newcommand{\method}{{\textsc{AOT-POT}}\xspace}
\usepackage{wrapfig}
\usepackage{graphicx}
\usepackage{bm}
\usepackage{amsmath}
\usepackage{amssymb}
\usepackage{makecell}
\usepackage{pifont}
\setcitestyle{numbers,square}
\definecolor{tableblue}{HTML}{AFC2FF}
\definecolor{tablelightblue}{HTML}{E1E8FF}
\definecolor{mydarkblue}{rgb}{0,0.08,0.45}
\definecolor{mydarkgreen}{RGB}{0, 139, 69}
\definecolor{mycyan}{cmyk}{.3,0,0,0}
\definecolor{dkgreen}{rgb}{0,0.6,0}
\definecolor{dkred}{rgb}{0.6,0,0}
\definecolor{dkblue}{rgb}{0,0,0.6}
\definecolor{purple}{rgb}{0.5,0,0.5}
\definecolor{azure(colorwheel)}{HTML}{007FFF}

\definecolor{uclablue}{rgb}{0.15, 0.45, 0.68}
\hypersetup{
    breaklinks,
    citecolor=uclablue,
    colorlinks=true,
    linkcolor=uclablue
}

\title{AOT-POT: Adaptive Operator Transformation for Large-Scale PDE Pre-training}

\addtocontents{toc}{\protect\setcounter{tocdepth}{-1}}

\author{%
  Qitan Lv$^{1,2}$\thanks{Equal contribution.} \quad
  Hong Wang$^{1}$\footnotemark[1] \quad
  Zhongkai Hao$^{3}$ \quad
  Wen Wu$^{2}$ \quad
  Xuenan Xu$^{2}$ \\
  \textbf{Bowen Zhou}$^{2,3}$ \quad 
  \textbf{Feng Wu}$^{1}$ \quad
  \textbf{Chao Zhang}$^{2,3}$\thanks{Corresponding author.} \\
  \\
  $^1$University of Science and Technology of China \quad
  $^2$Shanghai AI Laboratory \\
  $^3$Department of Electronic Engineering, Tsinghua University, China \\
  \texttt{\{qitanlv, wanghong1700\}@mail.ustc.edu.cn} \quad
  \texttt{fengwu@ustc.edu.cn} \\
  \texttt{\{wuwen, xuxuenan, zbw, zhangchao\}@pjlab.org.cn} \\
  \texttt{hzj21@mails.tsinghua.edu.cn}
}

\begin{document}

\maketitle

\begin{abstract}

Pre-training neural operators on diverse partial differential equation (PDE) datasets has emerged as a promising direction for building general-purpose surrogate models in scientific machine learning. However, the \textbf{inherent complexity} and \textbf{structural diversity} of PDE solution operators make multi-PDE pre-training fundamentally challenging. Existing methods mainly address this by increasing model capacity, while leaving the target solution operators unchanged. Inspired by classical numerical analysis, we instead propose to transform complex and diverse solution operators into simpler, better-aligned forms that are easier to model jointly. Since the optimal transformation varies across PDE types, it must be \emph{adaptive and input-dependent}, allowing a single neural operator to approximate an entire family of operators.
We instantiate this idea as \method (\textbf{a}daptive \textbf{o}perator-\textbf{t}ransformation for \textbf{p}re-training \textbf{o}perator \textbf{t}ransformer), which expands hidden representations into multiple parallel streams, adaptively aggregates and redistributes them before and after each sub-layer, and mixes streams through Sinkhorn-projected doubly stochastic matrices for stable training. These mechanisms together reshape diverse solution operators into a unified form that can be effectively modeled by a single architecture.
Empirically, \method achieves state-of-the-art performance on 12 PDE benchmarks with only 3\% additional parameters, reducing relative L2 error by up to 77.6\% (40.9\% on average). Fine-tuning \method further reduces L2 error by up to 92\% on in-domain PDEs and 89\% on out-of-domain PDEs (unseen types during pre-training), demonstrating that adaptive operator transformation is an effective and complementary direction for advancing PDE foundation models beyond simply scaling model capacity.
 
\end{abstract}

\section{Introduction}

Learning solution operators for partial differential equations (PDEs) is a fundamental task in scientific machine learning~\citep{karniadakis2021physics, li2020fourier}. It leverages parameterized PDE datasets to learn infinite-dimensional mappings between input spaces and solution function spaces, enabling generalization to unseen inputs~\citep{zachmanoglou1986introduction}. 
Acting as surrogate models, neural operators learn the time-evolution operator of PDEs and roll it out autoregressively to generate full trajectories, providing orders of magnitude faster than traditional solvers~\citep{pathak2022fourcastnet}. While architectures such as DeepONet~\citep{lu2021learning}, Fourier neural Operators~\citep{li2020fourier}, and neural-operator Transformers~\citep{hao2023gnot} have demonstrated remarkable success in specific applications like weather forecasting~\citep{pathak2022fourcastnet} and fluid dynamics~\citep{li2022fourier}, recent efforts are moving beyond single-domain models~\citep{hao2024dpot}. 
To build general-purpose surrogate models, pre-training neural operators on heterogeneous PDEs has emerged as a promising paradigm by mixing large amounts of diverse PDE data to train a unified model capable of handling multiple PDE types simultaneously~\citep{hao2024dpot, herde2024poseidon}.

\begin{wrapfigure}{l}{0.45\textwidth}

    \centering
    \includegraphics[width=\linewidth]{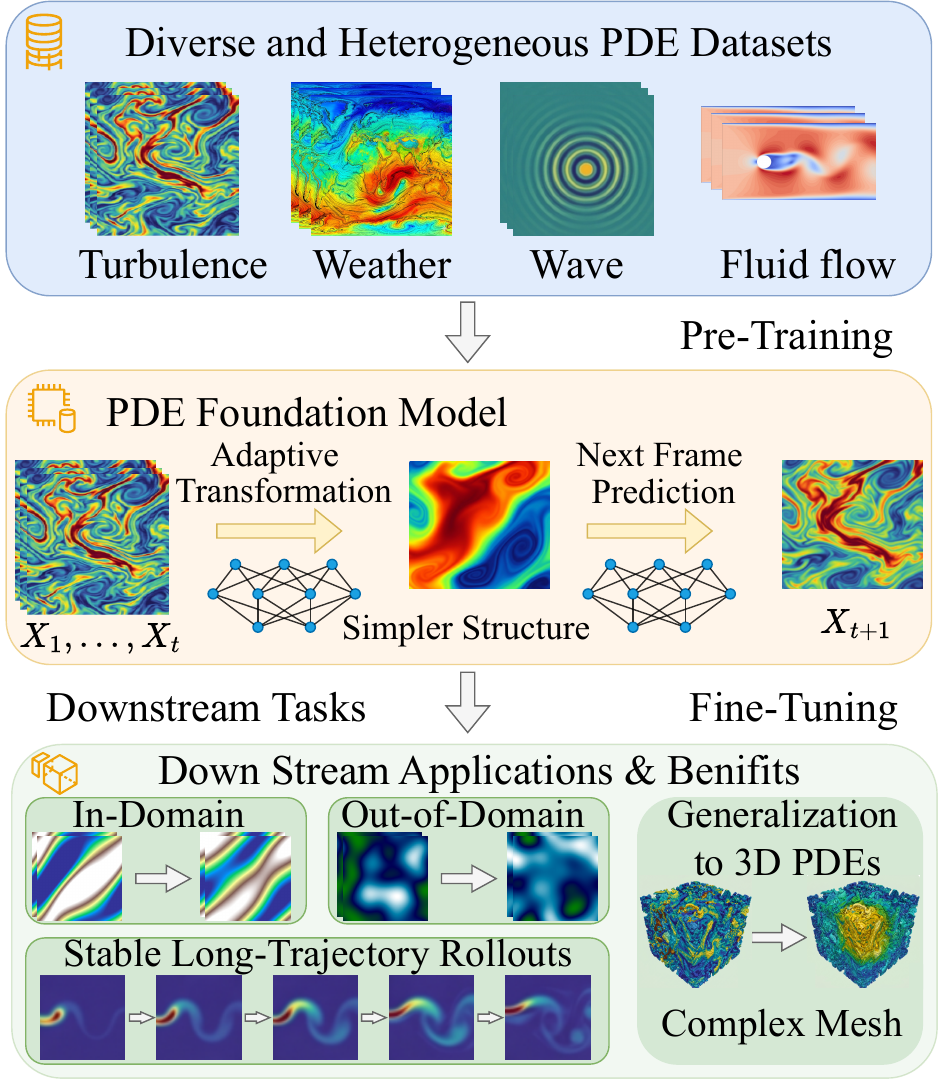}
    \caption{An illustration of pre-training a PDE foundation model using extensive data from diverse datasets. The model is then fine-tuned for diverse downstream operator learning tasks to handle complex scenarios.}
    \label{fig:intro_schematic}

\end{wrapfigure}

Despite this progress, multi-PDE pre-training is still highly challenging due to the heterogeneity of PDE solution operators: their underlying time-evolution
operators are intricately complex and differ substantially across PDEs, so forcing a single network to approximate all of them simultaneously
inevitably incurs large errors~\citep{wang2025mixture}. Existing methods
mainly enlarge model capacity through wider layers, deeper
networks~\citep{hao2024dpot}, or sparse expert
sub-networks~\citep{wang2025mixture}, yet do not exploit PDE-specific
characteristics to simplify the target solution operators themselves.
To tackle this challenge, we draw inspiration from a classical idea in
computational mathematics: a problem-tailored \emph{operator transformation}
can map a heterogeneous family of complex PDEs to a different space that is easier to solve with a given algorithm. For
example, under the Fourier
basis, the Laplacian reduces to pointwise multiplication, turning Poisson's
equation into a pointwise algebraic
relation~\citep{herde2024poseidon}. Inspired by this, we
introduce an analogous transformation directly into the neural-operator
architecture, so that the neural operator effectively learns from a transformed,
easier-to-approximate target rather than the original heterogeneous ones.

\textbf{Our insight.} Given a target solution operator
$\mathcal{G}: \mathcal{U} \to \mathcal{V}$ that a neural operator
$\mathcal{G}_w$ is trained to approximate, we insert into
$\mathcal{G}_w$ a learnable, input-dependent module that mimics an operator transformation, so that the backbone is effectively
trained against a simpler equivalent
$\tilde{\mathcal{G}}$ rather than $\mathcal{G}$ itself. Since different
PDEs benefit from different reformulations, this
 mechanism should be \emph{adaptive and
input-dependent}, so that a single model can handle diverse PDEs
in a unified architecture.




We instantiate this idea as \method, in which an adaptive operator
transformation is used at every layer through input-dependent,
multi-stream connections. Specifically, \method implements this
mechanism through three complementary components: (1)~it expands the hidden
representation into $n$ parallel streams, providing multiple basis components
for the latent function space; (2)~before each sub-layer, it adaptively
aggregates the streams with input-dependent weights, and after processing,
redistributes the output back to all streams with separately learned
weights---together playing a role analogous to a per-input change of basis
on the latent solution operator; (3)~at each layer, it mixes information
across streams via a doubly stochastic matrix obtained by Sinkhorn--Knopp
projection~\citep{sinkhorn1967concerning}, yielding a volume-preserving
transformation with bounded spectral norms for stable training. With only
3\% additional parameters, \method reduces relative L2 error by up to 77.6\%
over strong baselines, while exhibiting strong generalizability to downstream
tasks and improved stability for long-trajectory rollouts.
Our contributions are summarized as follows:
\begin{itemize}[itemsep=0pt, leftmargin=*]
    \item \textbf{New Perspective.} Inspired by the role of operator transformations in classical numerical methods, we extend this idea to neural-operator architectures as a new axis for improving multi-PDE pre-training, orthogonal to the prevailing paradigm of scaling model capacity. We show that suitable architectural modules can act as learnable, input-dependent operator transformations, effectively recasting the target solution operator into an equivalent but substantially simpler form that is easier to approximate while keeping the underlying backbone architecture largely unchanged.

    \item \textbf{Novel Architecture.} We propose \method, a novel architecture
    that generalizes standard residual connections to dynamically computed,
    doubly stochastic multi-stream transformations. By emulating an adaptive
    operator transformation at every layer, \method makes solution
    operators substantially easier to approximate with only 3\% additional
    parameters.

    \item \textbf{Consistent Empirical Gains.} \method achieves
    state-of-the-art performance on 12 PDE benchmarks, consistently reducing
    relative L2 error by up to 77.6\%. Fine-tuning the
    pre-trained model further reduces L2 error by up to 92\% on in-domain
    PDEs and up to 89\% on out-of-domain PDEs unseen during pre-training,
    demonstrating strong generalizability.


\end{itemize}

\section{Preliminaries}

\subsection{Time-dependent PDE Problem}

We consider parameterized time-dependent PDEs whose solutions $\bm{u}(x, t) \in \mathbb{R}^m$ are defined on a spatial domain $\Omega \subset \mathbb{R}^d$ over a time interval $[0, T] \subset \mathbb{R}$:
\begin{align}
  &\frac{\partial \bm{u}}{\partial t} - \mathcal{F}[\bm{u}; \theta](x, t) = 0,  &(x, t) \in \Omega \times [0, T] \subset \mathbb{R}^{d+1}, \\
  &\bm{u}(x, 0) = \bm{u}^0(x), \quad x \in \Omega,  &\mathcal{B}[\bm{u}](x, t) = 0, \quad x \in \partial \Omega, \nonumber
\end{align}
where $\mathcal{F}[\bm{u}; \theta]$ is a differential operator, $\theta \in \bm{\Theta}$ parameterizes the PDE type and coefficients, $\bm{u}^0(x)$ is the initial condition, and $\mathcal{B}[\bm{u}]$ specifies the boundary condition.
Each PDE solution is sampled on a spatial mesh of $\Omega$ and at time steps $0 = t_0 < \cdots < t_T \le T$, yielding a discretized trajectory $\bm{u}_i = (\bm{u}_i^1, \ldots, \bm{u}_i^T)$. In many real-world scenarios, the parameters $\theta$ are inaccessible, and the model must infer the underlying dynamics from the observed frames $(\bm{u}_i^1, \ldots, \bm{u}_i^T)$.

\subsection{Auto-regressive Denoising Pre-training}
\label{sec:pretrain}

Following DPOT~\citep{hao2024dpot}, we take the previous $T$ frames as input and predicts the next frame:
\begin{equation}
  \bm{u}^T = \mathcal{G}_w(\bm{u}^0, \ldots, \bm{u}^{T-1}).
\end{equation}
To mitigate error accumulation during long-trajectory rollouts~\citep{brandstetter2022message}, Gaussian noise $\bm{\varepsilon} \sim \mathcal{N}(0, \epsilon \|\bm{u}^{<t}\| I)$ is injected into inputs. The pre-training objective over the mixed dataset $\mathcal{D}$ is:
\begin{equation}
  \min_w \; \mathcal{L} = \mathbb{E}_{\bm{u} \sim p(\mathcal{D})} \sum_{1 \leq t \leq T} \left\| \mathcal{G}_w(\bm{u}^{<t} + \bm{\varepsilon}) - \bm{u}^t \right\|_2^2,
\end{equation}
where $\bm{u}^{<t} = (\bm{u}^0, \ldots, \bm{u}^{t-1})$ and $p(\cdot)$ is a balanced sampling strategy~\citep{hao2024dpot}.

\section{Motivated Experiments} \label{sec:motivation}

We motivate \method from a classical numerical-analysis perspective: reformulating a complex PDE solution operator into a simpler equivalent often makes it substantially easier to approximate. This view yields two concrete hypotheses we verify empirically below.

\paragraph{Operator transformation in classical numerical PDEs.}
In classical numerical methods, an \emph{operator transformation} $\mathcal{T}$ reformulates a solution operator $\mathcal{G}$ into a simpler equivalent $\tilde{\mathcal{G}}$---e.g., the Fourier basis for the Laplacian~\citep{herde2024poseidon}, or preconditioning for linear operators~\cite{saad2003iterative}. Since different PDEs induce operators of different structure, the appropriate $\mathcal{T}$ is itself PDE-specific.

\paragraph{From classical operator transformation to neural operators.}
Porting the view to a neural operator $\mathcal{G}_w$ approximating $\mathcal{G}$, prepending and appending a learnable operator transformation lets the backbone fit a simpler transformed target $\tilde{\mathcal{G}}$. We augment $\mathcal{G}_w$ with pointwise linear layers $\bm{W}_{\text{in}}, \bm{W}_{\text{out}} \in \mathbb{R}^{C\times C}$ so that $\bm{W}_{\text{out}} \circ \mathcal{G}_w \circ \bm{W}_{\text{in}}$ targets $\mathcal{G}$, while $\mathcal{G}_w$ only fits $\tilde{\mathcal{G}}$. We test two hypotheses: \textbf{H1}---a learnable operator transformation makes a solution operator easier to approximate; \textbf{H2}---the optimal transformation is PDE-specific: one that simplifies an operator may complicate another.

\paragraph{Setup: pointwise linear transformation.}
We test H1 and H2 with DPOT-Tiny ($7$M params)~\citep{hao2024dpot} on four PDE families: NS ($\nu\!=\!10^{-4}$, FNO~\citep{li2020fourier}), CNS ($\eta\!=\!\zeta\!=\!0.1$), SWE, and DR (PDEBench~\citep{takamoto2022pdebench}). The identity-initialized $\bm{W}_{\text{in}}, \bm{W}_{\text{out}} \in \mathbb{R}^{C\times C}$ add only $2C^2\!+\!2C$ parameters ($40$ for $C\!=\!4$). We compare {\texttt{Vanilla}} (base DPOT); {\texttt{Learned}} ($\bm{W}_{\text{in}}, \bm{W}_{\text{out}}$ trained with DPOT); and {\texttt{Frozen}} (linear layers pre-trained on PDE $k$ and frozen, then a fresh DPOT retrained from scratch on PDE $j$).

\begin{figure}[t]

    \centering
    \includegraphics[width=0.9\textwidth]{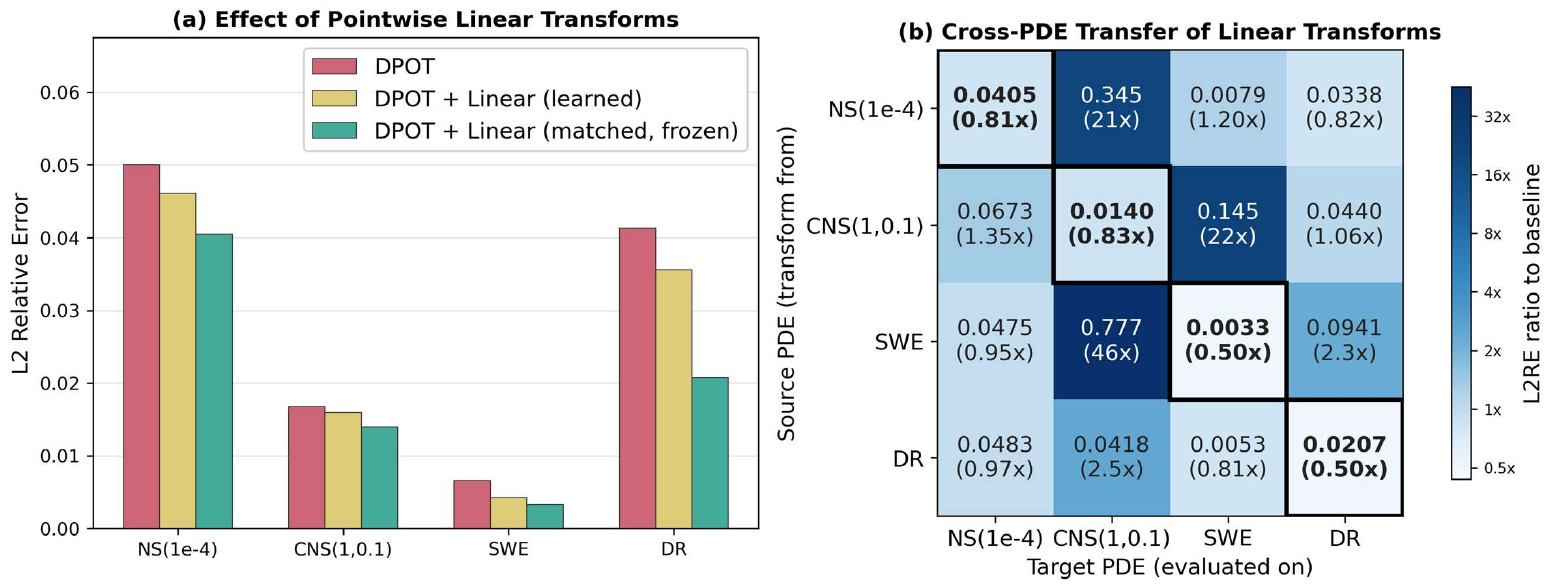}
    \caption{%
    \textbf{Different PDEs benefit from different operator transformations.}
    \textbf{(a)}~Adding pointwise linear layers consistently reduces L2RE across all four PDE
    families; freezing a matched transform and retraining the backbone from
    scratch outperforms joint optimization.
    \textbf{(b)}~Cross-PDE transfer of frozen transforms: rows index the
    source PDE (where the transform was trained), columns the target PDE
    (where the backbone is retrained). Diagonals (bordered) improve over
    Baseline (ratio $<\!1$); off-diagonals can degrade by up to $46\times$,
    showing the optimal operator transformation is strongly PDE-specific. Full 12-dataset results are in Appendix~\ref{app:motivation-full}.
    }
    \label{fig-motivation}

\end{figure}

\paragraph{A learnable operator transformation reduces approximation error.}
Figure~\ref{fig-motivation}(a) shows {\texttt{Learned}} beats {\texttt{Vanilla}} on all families despite adding only $\sim\!40$ of $\sim\!7$M parameters, which extends to all 12 datasets (Appendix~\ref{app:motivation-full}). A minimal learnable operator transformation reformulates the target easier to approximate. Matched {\texttt{Frozen}} beats {\texttt{Learned}} on all families; since freezing ensures the same capacity, the gain cannot come from capacity---a good transformation gives a better learning target. The {\texttt{Learned}}--Matched {\texttt{Frozen}} gap signals that a single $C \times C$ transformation is insufficient, motivating the richer per-layer, input-dependent family that \method realizes.

\paragraph{The optimal transformation is PDE-specific.}
In the cross-PDE transfer matrix (Figure~\ref{fig-motivation}(b)), diagonal entries consistently reduce L2RE relative to {\texttt{Vanilla}}, while off-diagonals can be catastrophic: CNS$\to$SWE worsens error by $22\times$, SWE$\to$CNS by $46\times$. A transformation that simplifies one solution operator may complicate another. Therefore, different PDEs need different transformations.

\paragraph{Implications for \method.}
First, operator transformation is a genuine axis for improving neural operators, with gains at as few as $\sim\!40$ parameters. Second, a multi-PDE pre-trained model must therefore learn \emph{adaptive, input-dependent} transformations, inferring the PDE-appropriate one from the input itself.
Based on this idea, \method introduces input-dependent multi-stream connections, including aggregation, redistribution, and Sinkhorn-projected doubly stochastic mixing. Together, these components mimic adaptive operator transformations and help the model effectively approximate diverse solution operators.

\section{Method}\label{sec:method}

\begin{figure}[t]
    \centering
    \includegraphics[width=0.9\textwidth]{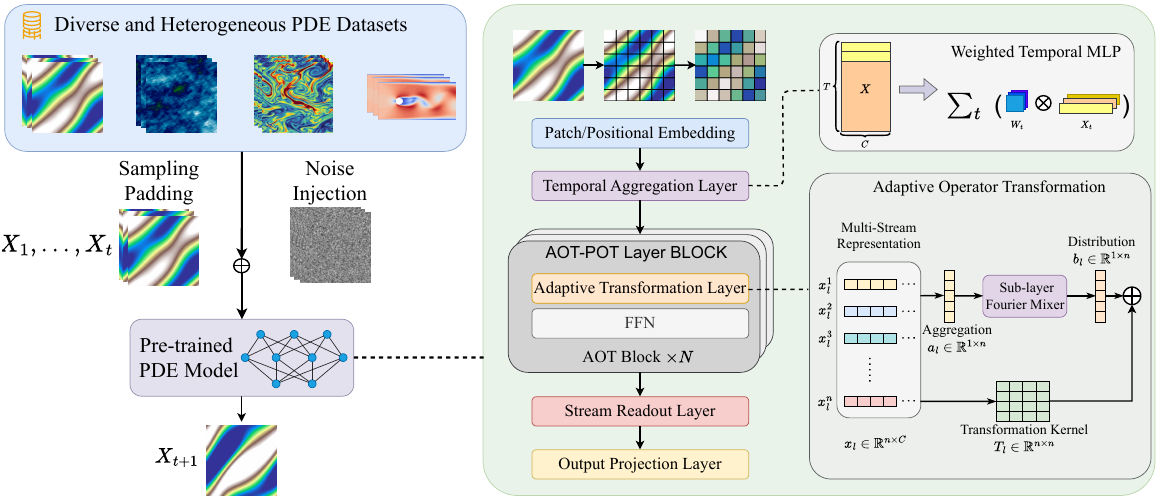}
    \caption{%
    \textbf{Overall architecture of \method.} After patchification and temporal
    aggregation, latent features are lifted into $n$ parallel streams and
    refined by $N\times$AOT blocks, each
    wrapping a Fourier attention layer. Inside an
    AOT block, the  aggregation $\bm{a}_l$, redistribution
    $\bm{d}_l$, and transformation kernel $\bm{T}_l$ together perform a
    per-input change of basis, mimicking an adaptive operator transformation.
    A gated readout then collapses the streams to predict the next
    timestep $\bm{u}^{t+1}$.}
    \label{fig:model}

\end{figure}

\paragraph{Overview.}
The architecture of \method is in Figure~\ref{fig:model}. The input PDE is reduced by a patchification and aggregation layer~\citep{hao2024dpot}, then refined by $N$ AOT blocks, each wrapping a Fourier attention layer~\citep{guibas2021adaptive} with change-of-basis mappings at every layer that mimic adaptive operator transformation.

\paragraph{Spatial encoding and temporal aggregation.}
  For input $\bm{u}^{<T} \in \mathbb{R}^{H \times W \times T \times C}$, where         
  $H$ and $W$ denote the spatial height and width,                               
  $T$ is the observed time step, and                                  
  $C$ is physical channels,                       
  we embed per-timestep patches via a convolutional layer $\mathcal{P}$ with           
  coordinate-aware positional encodings~\citep{dosovitskiy2020image}:  
\begin{equation}
    Z_p^t = \mathcal{P}(\bm{u}^t + \bm{p}^t), \quad t = 1, \ldots, T,
\end{equation}
where $p_{i,j}^t = W_p(x_i, y_j, t)$ with $W_p \in \mathbb{R}^{n \times 3}$, yielding patch features $Z_p^t \in \mathbb{R}^{H/p \times W/p \times C}$. To fuse the temporal axis, neighboring frames are aggregated via a learnable MLP $W_t$ by a constant $\bm{\gamma} \in \mathbb{R}^C$:
\begin{equation}
    \bm{z}_{\text{agg}} = \sum_t W_t \cdot \bm{z}_p^t \, e^{-i \bm{\gamma} t},
\end{equation}
where $\bm{z}_p^t \in \mathbb{R}^C$ is the per-node feature in $Z_p^t$, allowing the model to capture PDE temporal dynamics.

\paragraph{Fourier mixer.}
We adopt a multi-head Fourier attention layer to model spatial interactions. Let $z^l(x) \in \mathbb{R}^{d_z}$ denote the feature at location $x$ in block $l$, and define a kernel integral operator $\mathcal{K}_\phi$:
\begin{equation}
    (\mathcal{K}_\phi z^l)(x) = \int_\Omega \kappa(x, y; \phi) \, z^l(y) \, \mathrm{d}y.
\end{equation}
Under translation invariance $\kappa(x, y; \phi) = \kappa(x - y; \phi)$, this convolution is computed efficiently via:
\begin{equation}
    (\mathcal{K}_\phi z^l)(x) = \mathcal{F}^{-1}\!\left[R_\phi \cdot \mathcal{F}[z^l]\right]\!(x),
\end{equation}
with $R_\phi(k) \in \mathbb{C}^{d_z \times d_z}$ a learnable frequency-dependent weight. We split channels into $h$ heads, $z^l = \text{Concat}(z^l_1, \ldots, z^l_h)$ with $z^l_i(k) \in \mathbb{R}^{d_z / h}$, and process each through a two-layer complex MLP:
\begin{equation}
    z^l_{0,i}(x) = \mathcal{F}^{-1}\!\left[W^l_{2,i} \cdot \sigma\!\left(W^l_{1,i} \cdot \mathcal{F}[z^l_i] + b^l_{1,i}\right) + b^l_{2,i}\right]\!(x),
\end{equation}
where $W^l_{1,i}, W^l_{2,i} \in \mathbb{R}^{d_z/h \times d_z/h}$ and $b^l_{1,i}, b^l_{2,i} \in \mathbb{R}^{d_z/h}$, with outputs $z^l_0 = \text{Concat}(z^l_{0,1}, \ldots, z^l_{0,h})$.

\paragraph{Adaptive operator transformation block.}
A standard transformer block composes sub-layers as: $\bm{x}_{l+1} = \bm{x}_l + \mathcal{F}(\bm{x}_l, \mathcal{W}_l)$, where $\bm{x}_l \in \mathbb{R}^C$ is the hidden state and $\mathcal{F}$ is a sub-layer. This single fixed pathway limits the model's adaptation to the diverse operators of heterogeneous PDEs. To equip the latent representation with multiple basis components and enable a per-input change of basis, we draw inspiration upon~\citep{xie2025mhc} and propose the AOT block: $n$ parallel streams $\bm{x}_l \in \mathbb{R}^{n \times C}$ are maintained as components of an over-complete basis, and updated via three input-dependent linear mappings:

\begin{equation}
    \bm{x}_{l+1} = \bm{T}_l \, \bm{x}_l + \bm{d}_l^{\top} \, \mathcal{F}\!\left(\bm{a}_l \, \bm{x}_l, \mathcal{W}_l\right),
    \label{eq:aot}
\end{equation}

where each mapping plays a distinct role: (1) the \textit{aggregation} $\bm{a}_l \in \mathbb{R}^{1 \times n}$ selects a  combination of the $n$ basis components as the sub-layer input; (2) the \textit{redistribution} $\bm{d}_l \in \mathbb{R}^{1 \times n}$ scatters the output back to all streams, reshaping which components receive the update; (3) the \textit{transformation kernel} $\bm{T}_l \in \mathbb{R}^{n \times n}$ mixes information across streams, applying a layer-wise change of basis.

\paragraph{Stability via doubly stochastic transformation kernels.}
While Eq.~\eqref{eq:aot} enriches representational capacity,
unconstrained $\bm{T}_l$ leads to severe training instability: the
composite residual mapping $\prod_{i=1}^{L-l} \bm{T}_{L-i}$ may amplify or
attenuate signals exponentially with depth~\citep{xie2025mhc}. To prevent
this, we constrain the transformation kernel to the Birkhoff polytope---the
set of doubly stochastic matrices:
\begin{equation}
    \mathcal{M} = \left\{
        \bm{T}_l \in \mathbb{R}^{n \times n}
        \;\middle|\;
        \bm{T}_l \bm{1}_n = \bm{1}_n, \;
        \bm{1}_n^\top \bm{T}_l = \bm{1}_n^\top, \;
        \bm{T}_l \geq 0
    \right\}.
    \label{eq:birkhoff}
\end{equation}
This constraint endows the kernel with two
properties that make it well-suited as a change-of-basis operator:
(1) \textit{Norm preservation:} the spectral norm of a doubly stochastic
matrix is bounded by 1, preventing gradient explosion across layers;
(2) \textit{Compositional closure:} the product of doubly stochastic
matrices remains doubly stochastic, guaranteeing stability at arbitrary
depth.

\paragraph{Input-dependent parameterization.}
We parameterize $\bm{a}_l, \bm{d}_l, \bm{T}_l$ as input-dependent functions of the hidden state, so that the AOT block adjusts its change of basis to the current PDE. Given the hidden state $\bm{x}_l \in \mathbb{R}^{n \times C \times H' \times W'}$ ($H', W'$ are the post-patchification latent dimensions), we apply global average pooling $\bar{\bm{x}}_l = \text{GAP}_{H',W'}(\bm{x}_l) \in \mathbb{R}^{n \times C}$, flatten to $\text{vec}(\bar{\bm{x}}_l) \in \mathbb{R}^{1 \times nC}$, and compute:
\begin{equation}
\begin{aligned}
    \text{vec}(\bar{\bm{x}}_l)' &= \text{RMSNorm}(\text{vec}(\bar{\bm{x}}_l)), &
    \tilde{\bm{a}}_l &= \alpha^{a}_l \cdot (\text{vec}(\bar{\bm{x}}_l)' \, \phi^{a}_l) + \bm{b}^{a}_l, \\[6pt]
    \tilde{\bm{d}}_l &= \alpha^{d}_l \cdot (\text{vec}(\bar{\bm{x}}_l)' \, \phi^{d}_l) + \bm{b}^{d}_l, \quad &
    \tilde{\bm{T}}_l &= \alpha^{T}_l \cdot \text{mat}\!\left(\text{vec}(\bar{\bm{x}}_l)' \, \phi^{T}_l\right) + \bm{b}^{T}_l,
\end{aligned}
\label{eq:param}
\end{equation}
where $\phi^{a}_l, \phi^{d}_l \in \mathbb{R}^{nC \times n}$ and $\phi^{T}_l \in \mathbb{R}^{nC \times n^2}$ are learnable projections, $\text{mat}(\cdot)$ reshapes $\mathbb{R}^{1 \times n^2}$ to $\mathbb{R}^{n \times n}$, and $\alpha^{a}_l, \alpha^{d}_l, \alpha^{T}_l$ are learnable gating scalars. $\bm{b}^{a}_l, \bm{b}^{d}_l$ are zero-initialized and $\bm{b}^{T}_l$ is identity-initialized, so training starts from a stable state and gradually diversifies. The final mappings are:
\begin{equation}
    \bm{a}_l = \frac{\sigma(\tilde{\bm{a}}_l)}{\bm{1}_n^{\!\top} \sigma(\tilde{\bm{a}}_l)}, \quad
    \bm{d}_l = 2\,\sigma(\tilde{\bm{d}}_l), \quad
    \bm{T}_l = \text{Sinkhorn-Knopp}(\tilde{\bm{T}}_l),
\label{eq:constraint}
\end{equation}
where $\sigma(\cdot)$ is the Sigmoid function. Post-normalization places $\bm{a}_l$ on the probability simplex ($\bm{a}_l \geq 0$, $\bm{1}_n^{\!\top} \bm{a}_l = 1$), so $\bm{a}_l \bm{x}_l$ is a convex combination of $n$ basis components.
The Sinkhorn--Knopp operator~\citep{sinkhorn1967concerning} projects $\tilde{\bm{T}}_l$ onto the Birkhoff polytope by iteratively normalizing rows and columns.

\paragraph{Stream readout.}
After the final AOT block, the $n$ parallel streams
$\bm{x}_N^{(1)}, \ldots, \bm{x}_N^{(n)}$ will be collapsed back into a
single output stream by the gated readout:
\begin{equation}
    \bm{x}_{\text{out}}
    = \sum_{i=1}^{n} g_i \, \bm{x}_N^{(i)},
    \quad
    \bm{g} = \text{softmax}(\bm{w}),
    \;
    \bm{w} \in \mathbb{R}^{n},
    \;
    \bm{w} \overset{\text{init}}{=} \bm{0}.
    \label{eq:readout}
\end{equation}
The zero initialization gives a uniform mixing $g_i = 1/n$ at the start
of training, after which the model learns to weight basis components
according to their global utility for the prediction task. 

\paragraph{Loss function.}
We adopt an auto-regressive denoising objective:
\begin{equation}
    \mathcal{L} = \sum_{1 \leq t \leq T}
    \left\| \mathcal{G}_w(\bm{u}^{<t} + \bm{\varepsilon}) - \bm{u}^t \right\|_2^2,
    \label{eq:loss}
\end{equation}
where $\mathcal{G}_w$ is the model's prediction function, $\bm{u}^{<t}$ denotes the
input from previous timesteps, and
$\bm{\varepsilon} \sim \mathcal{N}(0, \epsilon \|\bm{u}^{<t}\| I)$ is Gaussian noise
injected for robustness.

\paragraph{Interpretation as an adaptive operator-transformation.}
The $n$ streams form an over-complete basis of the latent space. Input-dependent $\bm{a}_l, \bm{d}_l$ select and re-weight basis components, while doubly stochastic $\bm{T}_l$ applies a volume-preserving change of basis on the residual path. Together they mimic a layer-wise, per-input operator transformation, so the Fourier attention sub-layer learns a simpler, easier-to-approximate solution operator. Since all mappings are computed dynamically, the AOT block adapts to PDE-specific structure within a unified architecture without explicit PDE types.

\section{Experiments}\label{sec:experiments}


We evaluate \method on:
(1)~main results of \method on 12 datasets;
(2)~downstream knowledge transfer;
(3)~long-trajectory rollout stability;
(4)~fine-tuning sample efficiency;
(5)~ablation of the AOT block;
(6)~interpretability of the transformation layer.
Appendices \ref{app:downstream} to \ref{app:stability} additionally include: scaling law of \method, hyperparameter sensitivity, comparison with Poseidon, dataset heterogeneity, inference time, exploration on the combination of MoE and AOT, and training stability.

\subsection{Experiment Setups}

\paragraph{Datasets and baselines.}
We pre-train \method on 12 datasets from four benchmarks---
FNO~\citep{li2020fourier}, PDEBench~\citep{takamoto2022pdebench},
PDEArena~\citep{gupta2022towards}, and CFDBench~\citep{luo2023cfdbench}. All data is preprocessed following DPOT~\citep{hao2024dpot}. We compare against (1)~\textit{small models} trained individually per
dataset: FNO/Geo-FNO~\cite{li2020fourier,li2022fourier},
UNet~\cite{ronneberger2015u}, FFNO~\cite{tran2021factorized},
GK-Transformer~\cite{cao2021choose}, OFormer~\cite{li2022transformer},
GNOT~\cite{hao2023gnot}; and (2)~\textit{pre-trained models}: FNO-m,
MPP~\cite{mccabe2023multiple}, DPOT-(T/S/M)~\cite{hao2024dpot}, and
MoE-POT-(T/S/M)~\cite{wang2025mixture}. Details are in Appendix~\ref{app:preprocessing}.

\paragraph{Training.}
We train \method-Tiny/Small/Medium with AdamW for 1000 epochs on 8 NVIDIA
H200 GPUs (140\,GB) with a total batch size of 160. The model
auto-regressively takes $10$ frames to predict the next. We report
the $l_2$ relative error (L2RE), the standard metric in operator
learning~\cite{li2020fourier}.

\subsection{Main Experiments}

\begin{table*}[t]

\centering
\setlength{\tabcolsep}{1pt}
\caption{Results of \method on 12 PDE datasets. We \textbf{bold} the best results in each
part and highlight the globally best results using \colorbox{tableblue}{blue}
and the results of our \method using \textcolor{dkred}{darkred}.}\label{tb-main}
\resizebox{\textwidth}{!}{%
\setlength{\tabcolsep}{4pt}
\renewcommand{\arraystretch}{1.3}
\begin{tabular}{ll|cccccccccccc}
\toprule
L2RE & Params & \multicolumn{3}{c|}{FNO-$\nu$} & \multicolumn{6}{c|}{PDEBench CNS-$(\eta, \zeta)$,DR,SWE} & \multicolumn{2}{c|}{PDEArena} & CFDBench \\
Subset & -- & 1e-5 & 1e-4 & \multicolumn{1}{c|}{1e-3} & 1,0.1 & 1,0.01 & 0.1,0.1 & 0.1,0.01 & DR & \multicolumn{1}{c|}{SWE} & NS & \multicolumn{1}{c|}{NS-cond} & -- \\ \hline
\multicolumn{2}{c|}{Small Model} &  &  &  &  &  &  &  &  &  &  &  &  \\
FNO & 0.5M & 0.156 & 0.0834 & 0.0128 & 0.0980 & 0.0960 & 0.360 & 0.170 & 0.120 & \textbf{0.00440} & 0.0912 & \textbf{0.319} & 0.00761 \\
UNet & 25M & 0.198 & 0.119 & 0.0245 & 0.334 & 0.291 & 0.569 & 0.357 & 0.0971 & 0.0521 & 0.102 & 0.337 & 0.209 \\
FFNO & 1.3M & \textbf{0.121} & 0.0503 & 0.00990 & \textbf{0.0212} & 0.0520 & 0.162 & 0.0452 & 0.0571 & 0.0116 & \textbf{0.0839} & 0.602 & \textbf{0.00714} \\
GK-T & 1.6M & 0.134 & 0.0792 & \textbf{0.00980} & 0.0341 & \textbf{0.0377} & 0.0274 & 0.0366 & 0.0359 & 0.00692 & 0.0952 & 0.423 & 0.0105 \\
GNOT & 1.8M & 0.157 & \textbf{0.0443} & 0.0125 & 0.0325 & 0.0420 & \textbf{0.0228} & 0.0341 & 0.0311 & 0.00678 & 0.172 & 0.325 & 0.00877 \\
Oformer & 1.9M & 0.171 & 0.0645 & 0.0104 & 0.0417 & 0.0625 & 0.0254 & \textbf{0.0205} & \textbf{0.0192} & 0.00717 & 0.135 & 0.332 & 0.0102 \\ \midrule
\multicolumn{2}{c|}{Pre-trained} &  &  &  &  &  &  &  &  &  &  &  &  \\
FNO-m & 7M & {0.116} & 0.0922 & 0.0156 & 0.151 & 0.108 & 0.230 & 0.0760 & 0.0321 & 0.00912 & 0.210 & 0.384 & 0.0274 \\
MPP-Ti & 7M & -- & -- & -- & -- & -- & -- & -- & 0.0168 & 0.00660 & -- & -- & -- \\
MPP-S & 30M & -- & -- & -- & -- & -- & -- & -- & {0.0112} & {0.00240} & -- & -- & -- \\
MPP-L & 400M & -- & -- & -- & -- & -- & -- & -- & \textbf{0.00980} & 0.00220 & -- & -- & -- \\

{DPOT-T} & 7M & 0.0976 & 0.0606 & 0.00954 & 0.0173 & 0.0397 & 0.0132 & 0.0220 & 0.0321 & 0.00560 & 0.125 & 0.384 & 0.00952 \\
{DPOT-S} & 30M & 0.0553 & 0.0442 & 0.0131 & 0.0153 & 0.0337 & 0.0119 & 0.0188 & 0.0379 & 0.00657 & 0.0991 & 0.316 & 0.00696 \\
{DPOT-M} & 122M & 0.0409 & 0.0285 & 0.00474 & 0.0116 & 0.0238 & 0.00866 & 0.0129 & 0.0292 & 0.00290 & 0.0812 & 0.276 & 0.00752 \\

{MoE-POT-T} & 17M & 0.0965 & 0.0588 & 0.00896 & 0.0160 & 0.0395 & 0.0126 & 0.0215 & 0.0310 & 0.00527 & 0.134 & 0.367 & 0.0103 \\
{MoE-POT-S} & 90M & 0.0527 & 0.0417 & 0.0126 & 0.0143 & 0.0323 & 0.0113 & 0.0181 & 0.0354 & 0.00629 & 0.106 & 0.298 & 0.00748 \\
{MoE-POT-M} & 288M & 0.0385 & 0.0273 & 0.00442 & 0.0112 & 0.0226 & 0.00826 & 0.0121 & 0.0276 & 0.00279 & 0.0856 & 0.262 & 0.00820 \\

\textcolor{dkred}{\method-T} & 7M & 0.0610 & 0.0438 & 0.00604 & 0.0110 & 0.0280 & 0.00880 & 0.0130 & 0.0241 & 0.00433 & 0.0985 & 0.345 & 0.00776 \\
\textcolor{dkred}{\method-S} & 31M & 0.0376 & 0.0333 & 0.00293 & \textbf{0.00767} & 0.0204 & 0.00833 & 0.00965 & 0.0146 & 0.00266 & 0.0784 & 0.261 & \textbf{0.00493} \\
\textcolor{dkred}{\method-M} & 124M & \textbf{0.0312} & \textbf{0.0257} & \textbf{0.00236} & 0.00906 & \textbf{0.0168} & \textbf{0.00788} & \textbf{0.00954} & 0.0186 & \textbf{0.00200} & \textbf{0.0775} & \textbf{0.242} & 0.00543 \\

\midrule

\multicolumn{2}{c|}{Finetune} &  &  &  &  &  &  &  &  &  &  &  &  \\

{DPOT-T} & 7M & 0.0520 & 0.0367 & 0.00580 & 0.0112 & 0.0195 & 0.0174 & 0.0138 & 0.0148 & 0.00241 & 0.0910 & 0.280 & 0.00391 \\
{DPOT-S} & 30M & 0.0322 & 0.0237 & 0.00437 & 0.0129 & 0.0167 & 0.0152 & 0.0126 & 0.0129 & 0.00235 & 0.0867 & 0.268 & 0.00382 \\
{DPOT-M} & 122M & 0.0229 & 0.0126 & 0.00335 & 0.00998 & 0.0146 & 0.0161 & 0.00947 & 0.0103 & 0.00227 & 0.0294 & 0.172 & 0.00373 \\

{MoE-POT-T} & 17M & 0.0588 & 0.0352 & 0.00546 & 0.0108 & 0.0185 & 0.0163 & 0.0131 & 0.0140 & 0.00232 & 0.0985 & 0.263 & 0.00417 \\
{MoE-POT-S} & 90M & 0.0355 & 0.0221 & 0.00419 & 0.0122 & 0.0161 & 0.0145 & 0.0118 & 0.0123 & 0.00221 & 0.0936 & 0.259 & 0.00403 \\
{MoE-POT-M} & 288M & 0.0239 & 0.0119 & 0.00314 & 0.00956 & 0.0141 & 0.0152 & 0.00903 & 0.00959 & 0.00220 & 0.0313 & 0.163 & 0.00406 \\

\textcolor{dkred}{\method-T} & 7M & 0.0302 & 0.0158 & 0.00277 & 0.00489 & 0.00655 & 0.0120 & 0.00388 & 0.0106 & 0.00151 & 0.0440 & 0.218 & 0.00299 \\
\textcolor{dkred}{\method-S} & 31M & 0.0223 & 0.00758 & 0.00221 & \colorbox{tableblue}{\textbf{0.00427}} & \colorbox{tableblue}{\textbf{0.00491}} & \colorbox{tableblue}{\textbf{0.00909}} & 0.00356 & 0.00686 & 0.00116 & 0.0291 & 0.170 & 0.00239 \\
\textcolor{dkred}{\method-M} & 124M & \colorbox{tableblue}{\textbf{0.0145}} & \colorbox{tableblue}{\textbf{0.00608}} & \colorbox{tableblue}{\textbf{0.00181}} & 0.00510 & 0.00943 & 0.0158 & \colorbox{tableblue}{\textbf{0.00323}} & \colorbox{tableblue}{\textbf{0.00642}} & \colorbox{tableblue}{\textbf{0.000720}} & \colorbox{tableblue}{\textbf{0.0236}} & \colorbox{tableblue}{\textbf{0.152}} & \colorbox{tableblue}{\textbf{0.00216}} \\

\bottomrule
\end{tabular}}

\end{table*}

Table~\ref{tb-main} reports L2RE on the 12 pre-training datasets in three
parts: individually trained small models, zero-shot pre-trained models,
and pre-trained models fine-tuned per subset (lower is better).

\method consistently outperforms DPOT and MoE-POT at all three scales.
\method-S reduces L2RE by 40.9\% on average and up to 77.6\% (on
FNO-$\nu\!=\!10^{-3}$) over DPOT-S with only 3\% extra parameters. \method-S (31M) surpasses DPOT-M (122M) on 11 of 12 datasets,
demonstrating that adaptive operator transformations are far more
parameter-efficient than simply scaling up model size. \method
attains the best zero-shot result on 11 of 12 datasets; the only exception
is PDB-DR, where MPP-L (400M, $3\times$ larger than \method-M) retains a
slight edge. On PDB-SWE, \method-M still surpasses MPP-L using less than one-third of parameters.
After fine-tuning, \method achieves the best results on all 12 datasets at
every scale. Combined with the zero-shot results, \method
attains SOTA on 11/12 datasets without fine-tuning and on all
12 with fine-tuning, demonstrating the effectiveness of adaptive and input-dependent operator transformation for different PDEs.

\subsection{Downstream Knowledge Transfer Experiments}

\begin{wraptable}{r}{0.5\textwidth}

    \centering
    \caption{Downstream task transferability results. $^*$Relative
    median $L^1$ error following~\citep{raonic2023convolutional}.}
    \label{tb-downstream}
    \scriptsize
    \setlength{\tabcolsep}{3pt}
    \renewcommand{\arraystretch}{1.1}
    \resizebox{0.48\textwidth}{!}{%
    \begin{tabular}{l|cccc}
        \toprule
        Method & Turb.\ & PDB-3D & Steady$^*$ & Kolm.\ \\
        \midrule
        (Geo-)FNO & 0.193 & 0.410 & 0.036 & 0.231 \\
        MPP-FT & 0.152 & -- & -- & 0.135 \\
        \midrule
        DPOT-S & 0.168 & 0.269 & 0.028 & 0.146 \\
        DPOT-FT-S & 0.129 & 0.254 & 0.019 & 0.089 \\
        MoE-POT-S & 0.155 & 0.251 & 0.022 & 0.154 \\
        MoE-POT-FT-S & 0.103 & 0.228 & 0.016 & 0.078 \\
        \textcolor{dkred}{\method-S} & 0.087 & 0.103 & 0.009 & 0.038 \\
        \textcolor{dkred}{\method-FT-S} & \textbf{0.041} & \textbf{0.057} & \textbf{0.005} & \textbf{0.026} \\
        \bottomrule
    \end{tabular}}

\end{wraptable}

We evaluate transferability on four unseen downstream tasks:
(1)~high-resolution 2D turbulence from PDEBench~\cite{takamoto2022pdebench}
(different resolutions/Reynolds from training);
(2)~3D Navier--Stokes from PDEBench, testing cross-dimensional generalization;
(3)~steady-state PDEs from CNO~\cite{raonic2023convolutional}, requiring
temporal-to-time-independent transfer; and
(4)~chaotic Kolmogorov turbulence~\cite{hao2024dpot}, challenging long-horizon
prediction. Following DPOT~\cite{hao2024dpot},  methods with ``FT'' suffix are
fine-tuned from pre-trained weights and methods without ``FT'' suffix are from
scratch, both for 500 epochs. As shown in Table~\ref{tb-downstream}, \method from
scratch outperforms DPOT and MoE-POT from scratch on all
tasks, indicating that the AOT block alone
provides a strong inductive bias. Also, \method
from scratch surpasses pre-trained DPOT-FT and MoE-POT-FT on all
tasks, suggesting that
adaptive operator transformations capture properties that
other models cannot fully provide.
Fine-tuning yields the strongest results: \method-FT-S, reducing errors over DPOT-FT-S by up to 77.6\% on the PDB-3D where distribution shift is most
severe. Consistent trends hold at Tiny and Medium scales in
Appendix~\ref{app:downstream}.

\begin{wraptable}{r}{0.5\textwidth}

    \centering
    \caption{Long-trajectory rollout results.}
    \label{tb-rollout}
    \scriptsize
    \setlength{\tabcolsep}{3pt}
    \renewcommand{\arraystretch}{1.2}
    \resizebox{0.5\textwidth}{!}{%
    \begin{tabular}{l|ccccc}
        \toprule
        Method & 20 & 50 & 100 & 200 & 500 \\
        \midrule
        DPOT-S & 0.00152 & 0.00163 & 0.00474 & 0.01689 & 0.06685 \\
        DPOT-FT-S & 0.00121 & 0.00139 & 0.00292 & 0.00933 & 0.02911 \\
        MoE-POT-S & 0.00173 & 0.00199 & 0.00491 & 0.01932 & 0.07015 \\
        MoE-POT-FT-S & 0.00132 & 0.00147 & 0.00275 & 0.01013 & 0.03053 \\
        \textcolor{dkred}{\method-S} & 0.00079 & 0.00095 & 0.00154 & 0.00641 & 0.01543 \\
        \textcolor{dkred}{\method-FT-S} & \textbf{0.00041} & \textbf{0.00049} & \textbf{0.00097} & \textbf{0.00304} & \textbf{0.00401} \\
        \bottomrule
    \end{tabular}}

\end{wraptable}

\subsection{Long-Trajectory Rollout Stability Experiments}

\begin{figure}[t]
    \centering
    \begin{minipage}[t]{0.48\textwidth}
        \centering
        \includegraphics[width=0.9\linewidth]{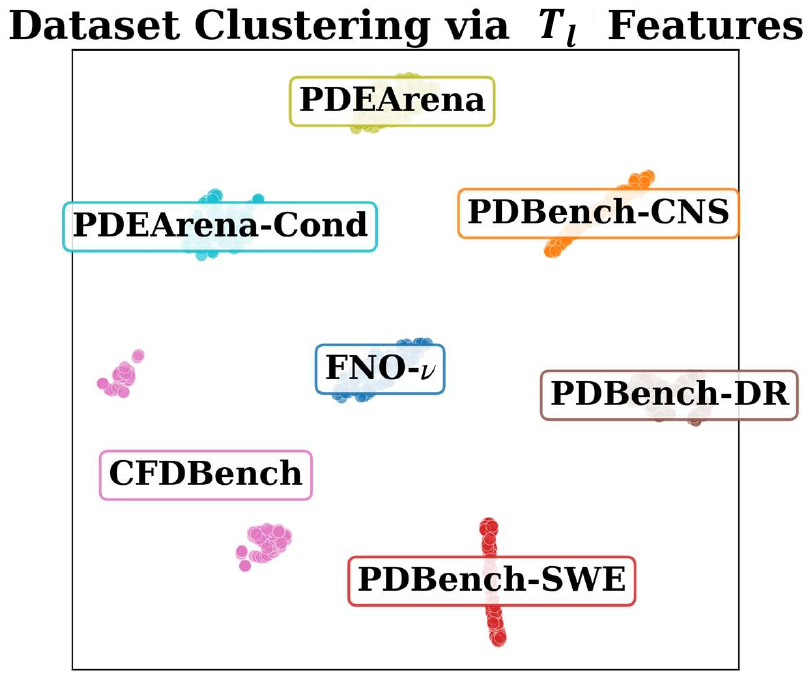}
    \end{minipage}
    \hfill
    \begin{minipage}[t]{0.48\textwidth}
        \centering
        \includegraphics[width=\linewidth]{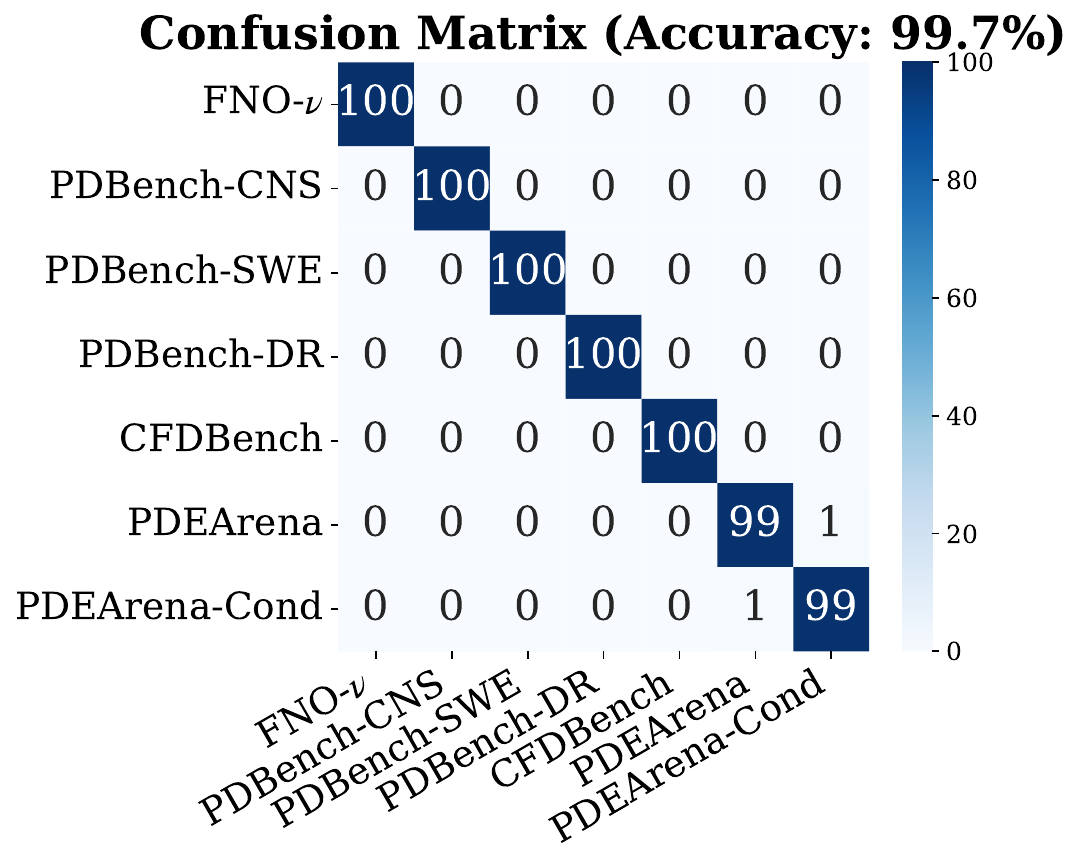}
    \end{minipage}
    \caption{\textbf{Interpretability of $\bm{T}_l$.}
    \textbf{Left:} t-SNE embedding of the concatenated $\bm{T}_l$
    features across all layers.
    \textbf{Right:} Confusion matrix of nearest-neighbor classification.}
    \label{fig-interpret}

\end{figure}

\begin{table}[t]
    \centering
    \caption{L2RE on the in-distribution NS ($10^{-4}$) and the
    out-of-distribution Wave-Layer.}
    \label{tb-fewshot}
    \scriptsize
    \setlength{\tabcolsep}{3pt}
    \renewcommand{\arraystretch}{1.1}
    \begin{minipage}[t]{0.52\textwidth}
        \centering
        \subcaption{In-distribution: NS ($10^{-4}$).}
        \label{tb-fewshot-id}
        \resizebox{\textwidth}{!}{%
        \begin{tabular}{l|cccccc}
            \toprule
            Method & 16 & 32 & 64 & 128 & 512 & 2000 \\
            \midrule
            DPOT-S & 0.36 & 0.31 & 0.29 & 0.20 & 0.094 & 0.048 \\
            DPOT-FT-S & 0.25 & 0.20 & 0.13 & 0.09 & 0.044 & 0.026 \\
            MoE-POT-S & 0.32 & 0.26 & 0.24 & 0.13 & 0.079 & 0.037 \\
            MoE-POT-FT-S & 0.20 & 0.15 & 0.11 & 0.07 & 0.040 & 0.016 \\
            \textcolor{dkred}{\method-S} & 0.22 & 0.15 & 0.12 & 0.07 & 0.013 & 0.008 \\
            \textcolor{dkred}{\method-FT-S} & \textbf{0.11} & \textbf{0.07} & \textbf{0.05} & \textbf{0.02} & \textbf{0.009} & \textbf{0.004} \\
            \bottomrule
        \end{tabular}}
    \end{minipage}%
    \hfill
    \begin{minipage}[t]{0.45\textwidth}
        \centering
        \subcaption{Out-of-distribution: Wave-Layer.}
        \label{tb-fewshot-ood}
        \resizebox{0.9\textwidth}{!}{%
        \begin{tabular}{l|cccc}
            \toprule
            Method & 16 & 32 & 64 & 128 \\
            \midrule
            DPOT-S & 0.59 & 0.53 & 0.47 & 0.39 \\
            DPOT-FT-S & 0.41 & 0.33 & 0.26 & 0.19 \\
            MoE-POT-S & 0.54 & 0.50 & 0.42 & 0.34 \\
            MoE-POT-FT-S & 0.34 & 0.26 & 0.20 & 0.14 \\
            \textcolor{dkred}{\method-S} & 0.30 & 0.27 & 0.22 & 0.15 \\
            \textcolor{dkred}{\method-FT-S} & \textbf{0.21} & \textbf{0.16} & \textbf{0.11} & \textbf{0.07} \\
            \bottomrule
        \end{tabular}}
    \end{minipage}

\end{table}

\begin{table}[t]
\centering
\caption{Ablation study of the three mappings in the AOT block.}
\label{tab-ablation}
\resizebox{\textwidth}{!}{%
\setlength{\tabcolsep}{4pt}
\renewcommand{\arraystretch}{1.3}
\begin{tabular}{l|cccccccccccc}
\toprule
 & \multicolumn{3}{c|}{FNO-$\nu$} & \multicolumn{6}{c|}{PDEBench CNS-$(\eta, \zeta)$,DR,SWE} & \multicolumn{2}{c|}{PDEArena} & CFDBench \\
Variant & 1e-5 & 1e-4 & \multicolumn{1}{c|}{1e-3} & 1,0.1 & 1,0.01 & 0.1,0.1 & 0.1,0.01 & DR & \multicolumn{1}{c|}{SWE} & NS & \multicolumn{1}{c|}{NS-cond} & -- \\
\midrule
w/o $\bm{a}$ & 0.03879 & 0.03349 & 0.00315 & 0.00769 & 0.02125 & 0.00930 & 0.01073 & 0.01470 & 0.00268 & 0.08620 & 0.27276 & 0.00550 \\
w/o $\bm{d}$ & 0.03867 & 0.03355 & 0.00320 & 0.00847 & 0.02065 & 0.00846 & 0.01027 & 0.01536 & 0.00269 & 0.07933 & 0.27247 & 0.00520 \\
w/o $\bm{T}$ & 0.03960 & 0.03653 & 0.00321 & 0.00816 & 0.02267 & 0.00987 & 0.01136 & 0.01734 & 0.00295 & 0.09257 & 0.30991 & 0.00587 \\
\midrule
{\method-S (full)} & {\textbf{0.03755}} & {\textbf{0.03326}} & {\textbf{0.00293}} & {\textbf{0.00767}} & {\textbf{0.02044}} & {\textbf{0.00833}} & {\textbf{0.00965}} & {\textbf{0.01461}} & {\textbf{0.00266}} & {\textbf{0.07842}} & {\textbf{0.26067}} & {\textbf{0.00493}} \\
\bottomrule
\end{tabular}%
}
\end{table}

Auto-regressive prediction suffers from cumulative single-step errors over
long horizons. We generate 1000 trajectories at $128\!\times\!128$
resolution and report L2RE at 20--500 steps using FNO-1e-3 generation codes~\citep{li2020fourier}. As shown in Table~\ref{tb-rollout}, the error of DPOT-FT-S grows $24.1\times$ from step 20 to 500, and MoE-POT-FT-S similarly
$23.1\times$. In contrast, \method-FT-S grows
only $9.8\times$---about $2.5\times$ slower---producing predictions that are both more accurate and more coherent with the underlying dynamics. Even \method without pre-training outperforms DPOT-FT and
MoE-POT-FT at step 500, confirming that
adaptive operator transformations stabilize multi-step rollouts beyond
what pre-training delivers. Consistent trends hold at Tiny and
Medium scales in Appendix~\ref{app:rollout}.

\subsection{Fine-tuning Efficiency Experiments}

To analyze data efficiency, we conduct few-shot fine-tuning on
(1)~NS ($\nu\!=\!10^{-4}$), close to the pre-training data, and
(2)~Wave-Layer from Poseidon~\cite{herde2024poseidon}, a novel unseen PDE 
system. All models fine-tune for 500 epochs while varying the number of
training samples, and we report the final L2RE.

\textbf{In-distribution NS} (Table~\ref{tb-fewshot-id}): \method-FT-S with
only 16 samples already beats DPOT-FT-S with 64 samples.
At 2000 samples, \method-FT-S reaches $0.004$, reducing error over
DPOT-FT-S and MoE-POT-FT-S by 84.6\% and 75.0\%.
\method trained from scratch with 128 samples already matches
MoE-POT-FT-S, showing the AOT block's strong data efficiency even without
pre-training. \textbf{Out-of-distribution Wave-Layer}
(Table~\ref{tb-fewshot-ood}): the advantage is equally pronounced.
\method-FT-S with 16 samples ($0.21$) outperforms DPOT-FT-S with 64
($0.26$), a $4\times$ data reduction; at 128 samples it reaches $0.07$,
reducing error by 63.2\%/50.0\% over DPOT-FT-S/MoE-POT-FT-S, indicating
that \method generalizes effectively to novel systems even in the extreme
low-data regime. Consistent conclusions hold at Tiny and Medium scales
in Appendix~\ref{app:fewshot}.

\subsection{Ablation Experiments}\label{sec:ablation}


We ablate each mapping in the AOT block in Eq.~\eqref{eq:param} by removing one at a time while
keeping the others active ($\bm{a}_l$/$\bm{d}_l$ to uniform aggregation/redistribution,
$\bm{T}_l$ to identity). All variants are pre-trained from scratch under
identical settings. Table~\ref{tab-ablation} shows the full \method achieves the best result
on all 12 datasets, confirming all three mappings contribute. The
transformation kernel $\bm{T}_l$ has the largest impact, degrading average
L2RE by 13.7\% when removed, versus 5.4\% for $\bm{a}_l$ and 4.2\% for
$\bm{d}_l$. This matches our motivation: $\bm{T}_l$ implements the
volume-preserving change of basis that constitutes the operator
transformation, while $\bm{a}_l$ and $\bm{d}_l$ select and redistribute
basis components. Consistent trends hold at Tiny and Medium scales
(Appendix~\ref{app:ablation}); we also study sensitivity to the number of
streams $n$, Sinkhorn iterations, and gating scalar initialization
$\alpha$ in Appendix~\ref{app:sensitivity}.

\subsection{Interpretability Experiments}\label{sec:interpretability}




Since $\bm{T}_l$ is computed dynamically from the input (Eq.~\eqref{eq:param}), a working mechanism should encode PDE-specific signatures that distinguish equation families. We verify this with a nearest-neighbor probe: for each of the 7 pre-training dataset groups (We group the FNO-$\nu$ and PDEBench CNS-$(\eta, \zeta)$ datasets with varying parameters into a single category, respectively.), we sample 100 test instances, concatenate the flattened $\bm{T}_l$ from all 6 layers of \method-S into one feature vector, and classify each sample by smallest Euclidean distance to per-dataset mean templates. The t-SNE embedding (Figure~\ref{fig-interpret}, left) shows clearly separated per-dataset clusters, and the confusion matrix (right) yields 99.7\% accuracy with 5 of 7 groups perfectly classified; the probe further generalizes to unseen Wave-Layer and Wave-Gauss from Poseidon~\citep{herde2024poseidon} at 100\% accuracy each in Appendix~\ref{app:interpretability}. Per-layer $\bm{T}_l$ heatmaps and the emergence of this classification ability over training are in Appendix~\ref{app:interpretability}.

\section{Limitations and Conclusions}\label{sec:conclusion}


\paragraph{Limitations.} We consider a few limitations of \method. First, Section~\ref{sec:motivation} provides empirical evidence that adaptive operator transformations reduce error, but a rigorous approximation-theoretic analysis is lacking. Second, although the kernels $\bm{T}_l$ encode PDE-specific signatures, a  formal link between the patterns and underlying equations remains open and may guide better architecture design.

\paragraph{Conclusions.}
We introduce \method, a novel model that learns \emph{adaptive, input-dependent operator transformations} to recast diverse solution operators into simpler equivalents tractable for a single neural operator. The AOT block maintains $n$ parallel streams as latent basis components and performs a per-input change of basis via aggregation $\bm{a}_l$, redistribution $\bm{d}_l$, and Sinkhorn-projected doubly stochastic mixing $\bm{T}_l$---with only 3\% extra parameters over vanilla DPOT. On 12 PDE benchmarks, \method achieves consistent SOTA performance, cutting relative L2 error by up to 77.6\% zero-shot and up to 92\%/89\% on in-/out-of-domain PDEs after fine-tuning, with markedly improved rollout stability. The learned $\bm{T}_l$ are highly interpretable: a nearest-neighbor classifier on $\bm{T}_l$ reaches 99.7\% PDE-type accuracy and generalizes to unseen PDEs. These results establish adaptive operator transformation as an effective axis for PDE foundation models, orthogonal to scaling model capacity.


\clearpage
\bibliographystyle{unsrtnat}

\clearpage
\appendix
\addtocontents{toc}{\protect\setcounter{tocdepth}{2}}

\renewcommand{\contentsname}{Appendix Table of Contents}

\tableofcontents
\section{Related Work}
\label{sec:related}

Here, we briefly summarize recent advancements on neural operators and pre-training in scientific machine learning.

\subsection{Neural Operators}

A growing body of work aims to learn PDE solution operators directly from data,
bypassing traditional numerical solvers while retaining predictive
accuracy~\citep{li2020fourier, pathak2022fourcastnet, zhao2024recfno,
augenstein2023neural, mukkavilli2023ai, kochkov2024neural, hao2022physics, brunton2024promising, subramanian2023towards}. The central challenge lies in the diversity of PDE families:
different equations exhibit distinct regularity, scales, and geometric structures,
demanding flexible architectural choices.

Among the earliest data-driven approaches, DeepONet~\citep{lu2021learning} factorizes
the operator into separate branch and trunk networks. The Fourier Neural Operator
(FNO)~\citep{li2020fourier} takes a different route by parameterizing the kernel
integral in frequency space, enabling efficient global feature extraction. Follow-up
works---Geo-FNO~\citep{li2022fourier}, NUNO~\citep{liu2023nuno},
GINO~\citep{li2023geometry}---generalize FNO to non-trivial geometries, while
PINO~\citep{li2024physics} and PI-DeepONet~\citep{wang2021learning} blend operator
learning with physics-informed
regularization~\citep{raissi2019physics, brandstetter2022message,
michalowska2023neural} to improve data efficiency.

A parallel line of research builds on transformer
architectures~\citep{hao2023gnot, cao2021choose, li2022transformer, wu2023solving}.
GNOT~\citep{hao2023gnot} applies cross-attention to handle irregular meshes, and
AFNO~\citep{guibas2021adaptive} performs token mixing entirely in the Fourier domain,
yielding memory and compute profiles comparable to
MLP-Mixer~\citep{tolstikhin2021mlp}; the latter has been adopted for global weather
prediction at scale~\citep{pathak2022fourcastnet}.\textbf{ A common limitation of these
methods, however, is that each model must be trained from scratch on a per-task basis,
motivating the search for transferable representations across PDE families.}

\subsection{Pre-training in Scientific Machine Learning}

Learning transferable representations from large corpora has transformed natural
language processing~\citep{radford2019language, brown2020language, chen2024sac, bai2025intern, lv2024coarse, chen2025knowledge, lv-etal-2025-exploiting, yang2025qwen3,liu2024deepseek} and computer
vision~\citep{he2022masked}, and the same idea is gaining traction in the sciences,
from protein folding~\citep{jumper2021highly} and molecular
design~\citep{zhou2023uni} to weather prediction~\citep{nguyen2023climax}.

In the area of PDE data, there have been initial attempts to explore pre-training
across different physical systems. SSL~\citep{mialon2023self} proposes contrastive
learning on PDEs using symmetries.
ICON~\citep{yang2023context} utilizes the structure of large language models to explore
in-context learning for PDE data. MPP~\citep{mccabe2023multiple} proposes an
auto-regressive approach to pre-train on time-dependent PDE datasets. However, these
explorations have so far been limited to specific types of equations and data scales.

Recently, DPOT~\citep{hao2024dpot} has taken a notable leap  by combining
auto-regressive denoising with a Fourier attention backbone and scales over 10+ heterogeneous PDE datasets, establishing strong multi-benchmark
baselines. \mbox{MoE-POT}~\citep{wang2025mixture} extends this line by replacing
dense feedforward layers with sparsely activated mixture-of-experts, so that
different PDE families are routed to dedicated expert sub-networks, maintaining
accuracy while curbing inference overhead. Poseidon~\citep{herde2024poseidon} adopts a
multiscale operator transformer and leverages the semi-group property of
time-dependent PDEs for efficient knowledge transfer.  Walrus~\citep{mccabe2025walrus} performs large-scale distributed pre-training on 15 datasets
  from the Well benchmark~\citep{ohana2024well}, further revealing the vast potential of neural operators to
  simultaneously learn from a wide variety of PDE types.   All of these advances focus on
scaling model capacity or routing computation more efficiently, yet leave the
target solution operators themselves as complex as before. \textbf{Our work takes an
orthogonal path: rather than enlarging or specializing the model, we mimic the
classical idea of \emph{operator transformation} within the neural-operator
architecture through a learnable, input-dependent module, recasting the
heterogeneous PDE solution operators into simpler, easier-to-approximate equivalents
that a single network can jointly approximate.}

\section{Details of Experiment Settings} \label{app:details}

\subsection{Data Preprocessing and Sampling}\label{app:preprocessing}

We follow the data preprocessing and sampling pipeline introduced in
DPOT~\citep{hao2024dpot}, with modifications to ensure compatibility across diverse PDE datasets.

\paragraph{Data padding and masking.}
To standardize spatial resolution across heterogeneous PDE datasets, we fix the
target resolution at $H = 128$, which matches the native resolution of a
significant portion of the datasets. Lower-resolution datasets are upscaled to
$H$ via bilinear interpolation, while higher-resolution datasets are
downscaled through random sub-sampling or interpolation. To unify the channel
dimension, all datasets are padded along the variable axis to match the dataset
with the maximum number of channels, with unused entries filled by a constant
value (e.g., 1). For datasets defined on irregular geometries, an additional
binary mask channel encodes the domain boundary, ensuring that the model can
distinguish valid spatial locations from padded regions.

\paragraph{Noise injection.}
During pre-training, Gaussian noise is injected into the input to improve
autoregressive rollout stability, as described in
Eq.~\eqref{eq:loss}. Specifically, for each input sequence
$\bm{u}^{<t}$, we add noise
$\bm{\varepsilon} \sim \mathcal{N}(0,\, \epsilon \|\bm{u}^{<t}\| I)$, where
$\epsilon$ controls the noise level. This noise injection is applied only
during pre-training to enable the robustness of the model prediction and is disabled during fine-tuning and evaluation.

\paragraph{Balanced data sampling.}
When training on a mixture of PDE datasets with varying sizes, uniform sampling
can lead to imbalanced gradient contributions and poor training efficiency. We
adopt the balanced sampling strategy from DPOT~\citep{hao2024dpot}: each
dataset $\mathcal{D}_k$ ($1 \leqslant k \leqslant K$) is assigned an
importance weight $w_k$, and the probability of sampling a data point from the
$k$-th dataset is
\[
    p_k = \frac{w_k}{K\, |\mathcal{D}_k| \cdot \sum_{k} w_k},
\]
where $|\mathcal{D}_k|$ denotes the number of samples in $\mathcal{D}_k$.
This formulation decouples the sampling probability from dataset size, ensuring
that smaller but important datasets receive adequate representation during
training. In all our experiments, for datasets with long trajectories, such as PDB-DR and PDB-SWE, we assigned a weight of $w=3$, while for others, the weight was set to 1.

\paragraph{Patchification.}
Following standard practice in vision
Transformers~\citep{dosovitskiy2020image}, we tokenize the spatiotemporal input
$\bm{u}^{<T} \in \mathbb{R}^{H \times W \times T \times C}$ by applying a
convolutional embedding layer with kernel size $P \times P$ and stride $P$:
\[
    \text{Conv2D}(C \to d,\; \text{kernel}=P,\; \text{stride}=P),
\]
which partitions the spatial domain into non-overlapping $P \times P$ patches
and maps each patch to a $d$-dimensional embedding vector. This reduces the
spatial token count by a factor of $P^2$ while preserving local structure,
enabling efficient global modeling in the subsequent Fourier attention layers.
We use $P = 8$ in all experiments.

\subsection{Design of the AOT block}\label{app:aot-design}

As detailed in Section~\ref{sec:method}, \method integrates an Adaptive
Operator Transformation (AOT) block---drawn upon the hyper-connection
mechanism of~\citep{xie2025mhc}.
Here we provide additional design details and discuss key implementation
choices.

\paragraph{Block architecture.}
Each \method block contains a Fourier attention layer
(AFNO2D) wrapped by an independent AOT block
with three input-dependent mappings: the aggregation $\bm{a}_l$, the
redistribution $\bm{d}_l$, and the transformation kernel $\bm{T}_l$. The
hidden state is expanded from $\mathbb{R}^{C}$ to $\mathbb{R}^{n \times C}$
at the input of the first block and collapsed back to $\mathbb{R}^{C}$ at
the output of the last block, where $n$ denotes the number of parallel
streams. Within each sub-layer, the update follows Eq.~\eqref{eq:aot}:
$\bm{a}_l$ aggregates the $n$ streams into a single input, the sub-layer
processes it unchanged from the original DPOT design, $\bm{d}_l$
redistributes the output back to all streams, and $\bm{T}_l$ applies a
volume-preserving change of basis on the residual path. Crucially, the
sub-layer itself operates on the standard single-stream
representation and requires no modification, making the AOT block a
drop-in enhancement to any existing DPOT backbone.

\paragraph{Normalization.}
We adopt Sandwich-Norm~\citep{ding2021cogview}, which applies GroupNorm~\citep{wu2018group} both
before and after each sub-layer. This additional post-normalization stabilizes
the multi-stream residual updates by preventing the accumulated outputs of
$\bm{d}_l$ from drifting in scale across layers.

\paragraph{Initialization and convergence.}
All gating scalars $\alpha^{a}_l$, $\alpha^{d}_l$, $\alpha^{T}_l$ are
initialized to 0.01, and the static bias $\bm{b}^{T}_l$ is initialized to
the identity matrix (Eq.~\eqref{eq:param}). This ensures that the model
begins in a state equivalent to the standard residual connection and
gradually learns to diversify the operator transformation during training.
The Sinkhorn-Knopp projection (Eq.~\eqref{eq:constraint}) uses 20
iterations, which we found sufficient for convergence in all configurations.
We also perform a parameter sensitivity analysis for the number of iterations
and the gating scalars $\alpha$ in Appendix~\ref{app:sensitivity}.

\paragraph{Design process.}
The current \method architecture is the result of systematic experimentation.
We explored several design axes: (1)~the number of parallel streams $n$,
(2)~whether to apply the AOT block to all sub-layers or only several 
sub-layers, (3)~normalization strategies (pre-norm only vs.\ Sandwich-Norm), and
(4)~the gating scalar initialization magnitude. Applying the AOT block to both
sub-layers consistently outperformed the single-sub-layer variant, and
Sandwich-Norm proved essential for stable training at the medium scale. \textbf{The
final design presented in this work is the robust and effective configuration
that emerged from this process.} Furthermore, in Appendix~\ref{app:moe_aot}, we explore an approach that combines the MoE architecture with the AOT block to provide deeper insights into the application of operator-transformation-based architectures within the PDE domain.

\paragraph{Parameter overhead.}
The AOT block introduces three small projection matrices per
sub-layer per block: $\phi^{a}_l, \phi^{d}_l \in
\mathbb{R}^{nC \times n}$ and $\phi^{T}_l \in
\mathbb{R}^{nC \times n^2}$, plus bias and gating scalar terms. For $n = 4$
and $C = 512$ (Tiny), this amounts to $2 \times 2 \times (4 \cdot 512 \times 4
+ 4 \cdot 512 \times 16) \approx 0.08\text{M}$ additional parameters per
block---a negligible overhead relative to the AFNO2D and MLP parameters. The
total AOT block overhead across all blocks is less than 3\% of the model
parameters.

\subsection{Model Sizes and Training Details}\label{app:training-details}

\paragraph{Model configurations.}
We evaluate \method at four scales: tiny, small, medium, and large. The
architectural configurations are summarized in Table~\ref{tab-model-size}.
All three scales share the same AOT block hyperparameters: $n = 4$ parallel
streams, 20 Sinkhorn-Knopp iterations, and a gating scalar initialization of
$\alpha = 0.01$. 

\begin{table}[h]
\centering
\caption{Architectural configurations of \method at different scales. The
AOT block hyperparameters ($n$, Sinkhorn iterations, $\alpha$) are shared across
all scales.}
\label{tab-model-size}
\renewcommand{\arraystretch}{1.1}
\resizebox{\textwidth}{!}{%
\begin{tabular}{lccccccccc}
\toprule
\makecell{Scale} &
\makecell{Attention\\dim} &
\makecell{MLP\\dim} &
\makecell{Layers} &
\makecell{AFNO\\blocks} &
\makecell{Streams\\($n$)} &
\makecell{Sinkhorn\\iter} &
\makecell{$\alpha$ init} &
\makecell{Sandwich\\Norm} &
\makecell{Params} \\
\midrule
Tiny   & 512  & 512  & 4  & 4 & 4 & 20 & 0.01 & \ding{51} & 7M   \\
Small  & 1024 & 1024 & 6  & 8 & 4 & 20 & 0.01 & \ding{51} & 31M  \\
Medium & 1024 & 4096 & 12 & 8 & 4 & 20 & 0.01 & \ding{51} & 124M \\
Large & 1536 & 6144 & 24 & 16 & 4 & 20 & 0.01 & \ding{51} & 515M \\
\bottomrule
\end{tabular}%
}
\end{table}

\paragraph{Pre-training.}
All models are pre-trained on 12 PDE datasets simultaneously. We use the Adam
optimizer with a learning rate of $1 \times 10^{-3}$, weight decay of
$1 \times 10^{-6}$, and momentum parameters $(\beta_1, \beta_2) = (0.9, 0.9)$.
Training follows a one-cycle learning rate schedule over 1000 epochs with 200
warm-up epochs. We train on 8 NVIDIA H200 GPUs (140\,GB each) with a total
batch size of 160. The patch size is set to $P = 8$, the number of retained
Fourier modes to 32, and the input sequence length to $T = 10$ timesteps. The
noise injection scale is $\epsilon = 5 \times 10^{-4}$. In all our experiments, for datasets with long trajectories, such as PDB-DR and PDB-SWE, we assigned a weight of $w=3$, while for others, the weight was set to 1.

\paragraph{Fine-tuning.}
For fine-tuning on pre-training subsets, we use the same Adam optimizer with
a learning rate of $1 \times 10^{-3}$ and a one-cycle schedule over 200
epochs with 40 warm-up epochs. For downstream tasks (including
out-of-distribution datasets), we extend training to 500 epochs with 100
warm-up epochs to compensate for the larger distribution shift. Unlike
MoE-based approaches that require freezing the router-gating
network~\citep{wang2025mixture}, \method fine-tunes all parameters end-to-end,
including the AOT block. Since the AOT block mappings are computed
dynamically from the input (Eq.~\eqref{eq:param}), they naturally adapt to the
target distribution without conflicting with pre-trained routing behavior.


\paragraph{Dataset configurations and splits.}
Table~\ref{tab-dataset-size} provides a comprehensive breakdown of the training and testing splits utilized across all pre-training datasets. The pre-training corpus encompasses a diverse range of physical systems with heterogeneous sample sizes, ranging from smaller datasets like PDEBench DR and SWE (900 training samples) to larger collections such as FNO-$\nu$ ($10^{-4}$) and CFDBench (up to 9,800 and 9,000 training samples, respectively). Conversely, Table~\ref{tab-dataset-size-downstream} delineates the data allocation for downstream evaluation. To ensure a standardized comparative analysis, all downstream tasks are uniformly constrained to 2,000 training samples and 200 testing samples. Notably, this evaluation suite includes both tasks that share underlying distributions with the pre-training families—specifically NS($10^{-4}$) and CNS(1,\,0.01)—as well as strictly out-of-distribution scenarios, such as the Kolmogorov flow, NS-3D, and the Wave equations.

\begin{table}[h]
\centering
\caption{Dataset sizes for pre-training (training / test splits).}
\label{tab-dataset-size}
\resizebox{\textwidth}{!}{%
\setlength{\tabcolsep}{4pt}
\begin{tabular}{l ccc cccccc cc c}
\toprule
Splits  & \multicolumn{3}{c|}{FNO-$\nu$} & \multicolumn{6}{c|}{PDEBench CNS-$(\eta, \zeta)$,DR,SWE} & \multicolumn{2}{c|}{PDEArena} & CFDBench \\
  & 1e-5 & 1e-4 & \multicolumn{1}{c|}{1e-3} & 1,0.1 & 1,0.01 & 0.1,0.1 & 0.1,0.01 & DR & \multicolumn{1}{c|}{SWE} & NS & \multicolumn{1}{c|}{NS-cond} & -- \\
\midrule
Train  & 1000 & 9800 & 1000 & 9000 & 9000 & 9000 & 9000 & 900 & 900 & 6500 & 3100 & 9000 \\
Test   &  200 &  200 &  200 &  200 &  200 &  200 &  200 &  60 &  60 &   650  &   200  & 1000 \\
\bottomrule
\end{tabular}%
}
\end{table}

\begin{table}[h]
\centering
\caption{Dataset sizes for downstream tasks (training / test splits). NS($10^{-4}$) and CNS(1,\,0.01) overlap with pre-training dataset families; the remaining tasks are out-of-distribution PDEs that are unseen during pre-training.}
\label{tab-dataset-size-downstream}
\resizebox{\textwidth}{!}{%
\setlength{\tabcolsep}{4pt}
\begin{tabular}{lccccccc}
\toprule
Splits & NS($10^{-4}$) & CNS(1,\,0.01) & PDEArena & Kolmogorov & NS-3D & Wave-Layer & Wave-Gauss \\
\midrule
Train & 2000 & 2000 & 2000 & 2000 & 2000 & 2000 & 2000 \\
Test  &  200 &  200 &  200 &  200 &  200 &  200 &  200 \\
\bottomrule
\end{tabular}%
}
\end{table}

\subsection{Mathematical Forms of Datasets}\label{app:dataset-math}

We list the governing PDEs of the 12 datasets used for pre-training. All
datasets follow the DPOT benchmark suite~\citep{hao2024dpot}.

\begin{itemize}

\item \textbf{FNO-$\boldsymbol{\nu}$}~\citep{li2020fourier}.
The quantity of interest is the vorticity
$w(x,t)$, $(x,t) \in [0,1]^2 \times [0,T]$, governed by the incompressible
Navier--Stokes equations:
\begin{align}
    \partial_t w + \bm{u} \cdot \nabla w &= \nu \,\Delta w + f(x), \\
    \nabla \cdot \bm{u} &= 0,
\end{align}
where $\nu$ is the kinematic viscosity and $f(x)$ is a fixed external forcing.
Three viscosity values are used: $\nu \in \{10^{-5}, 10^{-4}, 10^{-3}\}$,
yielding datasets of increasing turbulence intensity.

\item \textbf{PDEBench-CNS}~\citep{takamoto2022pdebench}.
We predict the velocity, pressure, and density fields
$\bm{u}(x,t)$, $p(x,t)$, $\rho(x,t)$, $(x,t) \in [0,1]^2 \times [0,1]$,
governed by the compressible Navier--Stokes equations:
\begin{align}
    \partial_t \rho + \nabla \cdot (\rho\,\bm{u}) &= 0, \\
    \rho\bigl(\partial_t \bm{u} + \bm{u} \cdot \nabla \bm{u}\bigr) &=
        -\nabla p + \eta\,\Delta \bm{u}
        + \bigl(\zeta + \tfrac{\eta}{3}\bigr)\nabla(\nabla \cdot \bm{u}), \\
    \partial_t \Bigl(\tfrac{3}{2}p + \tfrac{\rho\,|\bm{u}|^2}{2}\Bigr) &=
        -\nabla \cdot \Bigl[\Bigl(\varepsilon + p
        + \tfrac{\rho\,|\bm{u}|^2}{2}\Bigr)\bm{u}
        - \bm{u} \cdot \bm{\sigma}'\Bigr],
\end{align}
where $\eta$ and $\zeta$ are the shear and bulk viscosities. Four parameter
combinations are used:
$(M,\eta,\zeta) \in \{(1,0.1,0.1),\;(1,0.01,0.01),\;(0.1,0.1,0.1),\;(0.1,0.01,0.01)\}$,
where $M$ denotes the Mach number.

\item \textbf{PDEBench-SWE}~\citep{takamoto2022pdebench}.
We predict the water depth $h(x,t)$, $(x,t) \in [-1,1]^2 \times [0,5]$,
governed by the shallow water equations:
\begin{align}
    \partial_t h + \nabla \cdot (h\,\bm{u}) &= 0, \\
    \partial_t(h\,\bm{u}) + \nabla \cdot
        \Bigl(\tfrac{1}{2}\,h\,\bm{u}^2 + \tfrac{1}{2}\,g\,h^2\Bigr) &=
        -g\,h\,\nabla b,
\end{align}
where $g$ is gravitational acceleration and $b(x)$ is the bottom topography.

\item \textbf{PDEBench-DR}~\citep{takamoto2022pdebench}.
We predict the concentration fields $\bm{u}(x,t)$,
$(x,t) \in [-2.5,2.5]^2 \times [0,1]$, governed by the diffusion-reaction
equation:
\begin{equation}
    \partial_t \bm{u} = \bm{D}\,\nabla^2 \bm{u} + \bm{R}(\bm{u}),
\end{equation}
where $\bm{D}$ is the diffusion tensor and $\bm{R}(\bm{u})$ denotes the
nonlinear reaction term.

\item \textbf{PDEArena-NS / NS-Cond}~\citep{gupta2022towards}.
We predict the velocity and pressure fields $\bm{v}(x,t)$, $p(x,t)$,
$(x,t) \in [0,32]^2 \times [0,24]$, governed by the incompressible
Navier--Stokes equations with external forcing:
\begin{align}
    \partial_t \bm{v} &= -\bm{v} \cdot \nabla \bm{v}
        + \mu\,\nabla^2 \bm{v} - \nabla p + \bm{f}, \\
    \nabla \cdot \bm{v} &= 0.
\end{align}
The dataset includes two variants: NS (unconditional) and NS-Cond
(conditioned on forcing $\bm{f}$), totaling two datasets.

\item \textbf{CFDBench}~\citep{luo2023cfdbench}.
We predict the velocity and pressure fields $\bm{u}(x,t)$, $p(x,t)$ on
domains with irregular geometries. The governing equations are the
incompressible Navier--Stokes equations in conservative form:
\begin{align}
    \partial_t(\rho\,\bm{u}) + \nabla \cdot (\rho\,\bm{u} \otimes \bm{u})
        &= -\nabla p + \nabla \cdot
        \bigl[\mu\bigl(\nabla \bm{u} + (\nabla \bm{u})^\top\bigr)\bigr], \\
    \nabla \cdot (\rho\,\bm{u}) &= 0.
\end{align}
This dataset is unique in that it features varying boundary shapes and
conditions, requiring the model to generalize across geometries.

\end{itemize}


\section{Experimental Data and Supplementary Experiments}

\subsection{Downstream Tasks at Additional Scales}
\label{app:downstream}

We provide downstream task results of our \method at the tiny and medium scales to complement
the small-scale results in Table~\ref{tb-downstream}.

\begin{table*}[htbp]
    \centering
    \caption{Downstream task results for \method at the tiny scale.}
    \label{tb-downstream-tiny}
    \scriptsize
    \setlength{\tabcolsep}{3pt}
    \renewcommand{\arraystretch}{1.1}
    \resizebox{0.55\textwidth}{!}{%
    \begin{tabular}{l|cccc}
        \toprule
        Method & Turb.\ & 3D & Steady$^*$ & Kolm.\ \\
        \midrule
        (Geo-)FNO & 0.193 & 0.410 & 0.036 & -- \\
        MPP-FT & 0.152 & -- & -- & -- \\
        \midrule
        DPOT-T & 0.189 & 0.301 & 0.032 & 0.193 \\
        DPOT-FT-T & 0.143 & 0.277 & 0.027 & 0.101 \\
        MoE-POT-T & 0.173 & 0.288 & 0.030 & 0.173 \\
        MoE-POT-FT-T & 0.147 & 0.259 & 0.026 & 0.111 \\
        \textcolor{dkred}{\method-T} & 0.109 & 0.144 & 0.019 & 0.071 \\
        \textcolor{dkred}{\method-FT-T} & \colorbox{tableblue}{\textbf{0.076}} & \colorbox{tableblue}{\textbf{0.098}} & \colorbox{tableblue}{\textbf{0.011}} & \colorbox{tableblue}{\textbf{0.049}} \\
        \bottomrule
    \end{tabular}}
\end{table*}
As shown in Table~\ref{tb-downstream-tiny}, at the Tiny scale, \method-FT-T achieves the
best performance on all four tasks, reducing L2RE by 46.9\% on PDB-Turb, by
64.6\% on the 3D dataset, by 59.3\% on Steady, and by 51.5\% on Kolmogorov
relative to DPOT-FT-T. Notably, \method-T trained from scratch already
outperforms both DPOT-FT-T and MoE-POT-FT-T on all four tasks, confirming
that the architectural advantage of the AOT block is consistent even at small model
scales.

\begin{table*}[htbp]
    \centering
    \caption{Downstream task results for \method at the medium scale.}
    \label{tb-downstream-medium}
    \scriptsize
    \setlength{\tabcolsep}{3pt}
    \renewcommand{\arraystretch}{1.1}
    \resizebox{0.55\textwidth}{!}{%
    \begin{tabular}{l|cccc}
        \toprule
        Method & Turb.\ & 3D & Steady$^*$ & Kolm.\ \\
        \midrule
        (Geo-)FNO & 0.193 & 0.410 & 0.036 & -- \\
        MPP-FT & 0.152 & -- & -- & -- \\
        \midrule
        DPOT-M & 0.142 & 0.233 & 0.019 & 0.115 \\
        DPOT-FT-M & 0.097 & 0.209 & 0.010 & 0.097 \\
        MoE-POT-M & 0.123 & 0.188 & 0.017 & 0.102 \\
        MoE-POT-FT-M & 0.097 & 0.156 & 0.013 & 0.075 \\
        \textcolor{dkred}{\method-M} & 0.063 & 0.069 & 0.005 & 0.021 \\
        \textcolor{dkred}{\method-FT-M} & \colorbox{tableblue}{\textbf{0.021}} & \colorbox{tableblue}{\textbf{0.033}} & \colorbox{tableblue}{\textbf{0.002}} & \colorbox{tableblue}{\textbf{0.014}} \\
        \bottomrule
    \end{tabular}}

\end{table*}

At the Medium scale (Table~\ref{tb-downstream-medium}), \method-FT-M achieves
the strongest results across all scales and tasks, reducing L2RE by 78.4\% on
PDB-Turb ($0.021$ vs.\ $0.097$), by 84.2\% on the 3D dataset ($0.033$ vs.\
$0.209$), by 80.0\% on Steady ($0.002$ vs.\ $0.010$), and by 85.6\% on
Kolmogorov ($0.014$ vs.\ $0.097$) relative to DPOT-FT-M. The consistent
improvements across all three model scales confirm that the benefits of
our \method architecture are robust and scale reliably with model
capacity.

\subsection{Full Results for Motivated Experiments}
\label{app:motivation-full}

This section provides the complete results underlying the motivated experiments
in Section~\ref{sec:motivation}, including the full cross-PDE transfer matrix,
analysis at additional model scales, and visualization of the learned linear
transforms.

\subsubsection{Experimental Protocol}

We use the DPOT-Tiny architecture (7\,M parameters) as the backbone throughout
these experiments. A pointwise linear transform is instantiated as
$\bm{W}_{\text{in}}, \bm{W}_{\text{out}} \in \mathbb{R}^{C \times C}$
(identity-initialized, $C = 4$), applied to the channel dimension at the
input and output of the model, respectively. This adds exactly $2 \times 4^2 =
40$ learnable parameters.

Three configurations are evaluated:
\begin{itemize}
    \item \textbf{DPOT}: Baseline model without linear transforms.
    \item \textbf{DPOT+Linear}: The linear transforms are jointly optimized
          with the DPOT backbone on each dataset independently.
    \item \textbf{DPOT+Fixed-$k$}: The linear transforms trained on dataset $k$
          are loaded and frozen; the DPOT backbone is re-initialized from scratch
          and trained on the target dataset $j$. This isolates the effect of the
          fixed basis transformation.
\end{itemize}

All models are trained with identical hyperparameters (learning rate, schedule,
number of epochs) on each dataset.

\subsubsection{Full Cross-PDE Transfer Matrix}
\label{app:cross-transfer-table}

Table~\ref{tab:full-cross-transfer} presents the complete $14 \times 12$ matrix
of L2 relative errors. The first two rows correspond to the baseline (DPOT) and
jointly learned (DPOT+Linear) configurations. Rows 3--14 show the
Matched-Frozen configuration: each row uses a linear transform trained on the
dataset indicated in the row label, frozen during training on every target
dataset (columns). Diagonal entries (highlighted in \colorbox{tableblue}{blue})
correspond to matched transforms.

\begin{table*}[h]
\centering
\caption{Full cross-PDE transfer matrix (L2 Relative Error, DPOT-Tiny). Each
column is a target dataset. Rows 1--2: baselines. Rows 3--14: DPOT with a
frozen linear transform from the source dataset indicated in the row label.
Diagonal entries (\colorbox{tableblue}{blue}) correspond to matched
source-target pairs. Lower is better.}
\label{tab:full-cross-transfer}
\resizebox{\textwidth}{!}{%
\setlength{\tabcolsep}{3pt}
\renewcommand{\arraystretch}{1.15}
\begin{tabular}{l|ccc|cccc|c|c|c|c|c}
\toprule
\multirow{2}{*}{Configuration} & \multicolumn{3}{c|}{FNO NS ($\nu$)} & \multicolumn{4}{c|}{PDEBench CNS ($\eta$,$\zeta$)} & PDEBench & PDEBench & PDEArena & PDEArena & CFD \\
 & 1e-5 & 1e-4 & 1e-3 & (1,0.1) & (1,0.01) & (0.1,0.1) & (0.1,0.01) & DR & SWE & NS & NS-cond & Bench \\
\midrule
DPOT (baseline) & 0.0687 & 0.0500 & 0.0083 & 0.0167 & 0.0281 & 0.0212 & 0.0267 & 0.0413 & 0.0066 & 0.1363 & 0.3875 & 0.0093 \\
DPOT+Linear & 0.0678 & 0.0461 & 0.0082 & 0.0160 & 0.0280 & 0.0167 & 0.0211 & 0.0356 & 0.0043 & 0.1360 & 0.3768 & 0.0087 \\
\midrule
Fixed: NS(1e-5) & \cellcolor{tableblue}\textbf{0.0558} & 0.0460 & 0.0085 & 0.2288 & 0.0291 & 0.2854 & 0.0292 & 0.0382 & 0.0054 & 0.1361 & 0.3837 & 0.0096 \\
Fixed: NS(1e-4) & 0.0699 & \cellcolor{tableblue}\textbf{0.0405} & 0.0090 & 0.3448 & 0.0350 & 0.4836 & 0.0374 & 0.0338 & 0.0079 & 0.1369 & 0.3858 & 0.0089 \\
Fixed: NS(1e-3) & 0.0671 & 0.0501 & \cellcolor{tableblue}\textbf{0.0070} & 0.7089 & 0.0321 & 0.5082 & 0.0323 & 0.0374 & 0.0060 & 0.1364 & 0.3829 & 0.0089 \\
Fixed: CNS(1,0.1) & 0.9083 & 0.0673 & 0.0379 & \cellcolor{tableblue}\textbf{0.0140} & 0.0269 & 0.0138 & 0.0198 & 0.0440 & 0.1455 & 0.1427 & 0.4122 & 0.0094 \\
Fixed: CNS(1,0.01) & 0.1313 & 0.0529 & 0.0272 & 0.0139 & \cellcolor{tableblue}\textbf{0.0248} & 0.0143 & 0.0197 & 0.0307 & 0.0400 & 0.1402 & 0.3991 & 0.0092 \\
Fixed: CNS(0.1,0.1) & 0.9361 & 0.8624 & 0.4600 & 0.0974 & 0.0405 & \cellcolor{tableblue}\textbf{0.0123} & 0.0227 & 0.8349 & 0.6531 & 0.2594 & 0.8271 & 0.7241 \\
Fixed: CNS(0.1,0.01) & 0.2399 & 0.0562 & 0.0482 & 0.0159 & 0.0288 & 0.0208 & \cellcolor{tableblue}\textbf{0.0158} & 0.0447 & 0.1167 & 0.1415 & 0.4091 & 0.0176 \\
Fixed: DR & 0.0706 & 0.0483 & 0.0095 & 0.0419 & 0.0310 & 0.4499 & 0.0225 & \cellcolor{tableblue}\textbf{0.0207} & 0.0053 & 0.1365 & 0.3561 & 0.0088 \\
Fixed: SWE & 0.0689 & 0.0475 & 0.0097 & 0.7769 & 0.0403 & 0.9403 & 0.0453 & 0.0941 & \cellcolor{tableblue}\textbf{0.0033} & 0.1370 & 0.3543 & 0.0086 \\
Fixed: NS (PDA) & 0.0691 & 0.0464 & 0.0091 & 0.6130 & 0.0385 & 0.5015 & 0.0354 & 0.0547 & 0.0064 & \cellcolor{tableblue}\textbf{0.1103} & 0.3557 & 0.0094 \\
Fixed: NS-cond & 0.0696 & 0.0612 & 0.0093 & 0.5787 & 0.0348 & 0.4939 & 0.0274 & 0.0431 & 0.0062 & 0.1316 & \cellcolor{tableblue}\textbf{0.3636} & 0.0099 \\
Fixed: CFDBench & 0.0687 & 0.0464 & 0.0092 & 0.9592 & 0.0504 & 0.5022 & 0.0394 & 0.0931 & 0.0046 & 0.1374 & 0.3692 & \cellcolor{tableblue}\textbf{0.0054} \\

\bottomrule
\end{tabular}}
\end{table*}

\begin{figure*}[h]
    \centering
    \includegraphics[width=\textwidth]{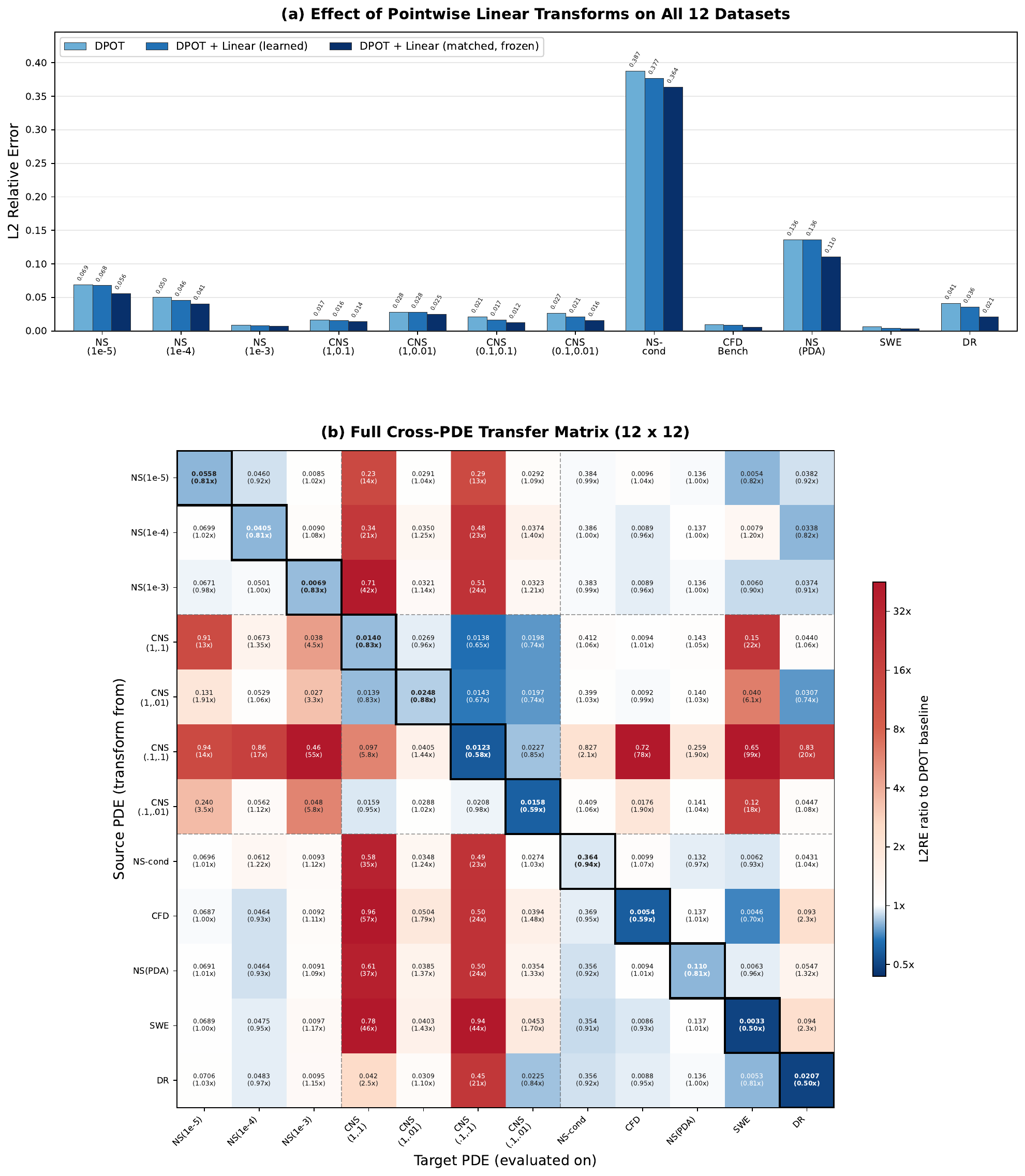}
    \caption{Full motivated experiment results on all 12 datasets (DPOT-Tiny).
    \textbf{(a)}~Bar chart comparing DPOT baseline (light blue), DPOT with
    jointly trained linear layers (medium blue), and DPOT with matched frozen
    linear transforms (dark blue). The matched frozen configuration yields the
    lowest error on every dataset. \textbf{(b)}~$12 \times 12$ cross-PDE
    transfer matrix. Rows indicate the source PDE (transform origin), columns
    the target PDE (evaluation). Diagonal entries (black borders, dark blue)
    show consistent improvement; off-diagonal entries reveal that mismatched
    transforms cause degradation ranging from mild (within the same PDE family)
    to catastrophic (across families, up to $99\times$). Dashed lines separate
    the FNO-NS, PDEBench-CNS, and remaining dataset groups.}
    \label{fig:motivation-all-tiny}
\end{figure*}

Figure~\ref{fig:motivation-all-tiny} provides a visual summary of the same
data: panel (a) shows the bar chart comparison across all 12 datasets, and
panel (b) displays the full $12 \times 12$ cross-transfer heatmap, where
diagonal entries (dark blue) consistently outperform the baseline while many
off-diagonal entries (red) indicate severe degradation.

\paragraph{Analysis.}
Several patterns emerge from Table~\ref{tab:full-cross-transfer} and
Figure~\ref{fig:motivation-all-tiny}:

\begin{enumerate}

\item \textbf{Jointly learned linear transforms universally help.}
Comparing the first two rows, DPOT+Linear improves over the DPOT baseline on
all 12 datasets despite adding only 40 parameters ($< 0.001\%$ of total). This
confirms that even a minimal linear change of basis in the input/output function
space can meaningfully reduce the effective complexity of the solution operator.

\item \textbf{Matched frozen transforms consistently outperform joint
optimization.}
The diagonal entries (blue cells) are lower than the corresponding DPOT+Linear
values on all 12 datasets, yielding reductions of 17--50\% over the DPOT
baseline. The most pronounced gains appear on SWE ($0.0033$ vs.\ baseline
$0.0066$, a 50\% reduction) and DR ($0.0207$ vs.\ $0.0413$, also 50\%). This
demonstrates that a correct PDE-specific basis---when provided to the
backbone---is more effective than allowing the backbone and transform to
co-adapt from random initialization. In other words, the linear transform
captures genuine structural properties of each PDE family that fundamentally
simplify learning.

\item \textbf{Mismatched transforms can be catastrophic.}
Off-diagonal entries reveal dramatic performance degradation when a transform
optimized for one PDE family is applied to another. The most extreme cases
involve the CNS($0.1, 0.1$) transform, which increases L2RE on NS(1e-5) from
$0.069$ to $0.936$ ($14\times$), on CFDBench from $0.009$ to $0.724$
($80\times$), and on DR from $0.041$ to $0.835$ ($20\times$). Conversely,
applying incompressible-flow transforms to compressible-flow datasets
(e.g., NS(1e-3)$\to$CNS(1,0.1): $0.709$, CFDBench$\to$CNS(1,0.1): $0.959$)
also leads to catastrophic failure. These results confirm that the optimal
function-space transformation is strongly PDE-dependent.

\item \textbf{Related PDE families show partial transferability.}
Transforms between PDEs within the same family show milder degradation. For
example, the three FNO NS datasets (differing only in viscosity $\nu$) exhibit
moderate cross-transfer performance, and the four CNS datasets transfer
reasonably among themselves. This is consistent with the intuition that PDEs
sharing similar mathematical structure benefit from similar basis
transformations.
\end{enumerate}

\subsubsection{Visualization of Learned Linear Transforms}
\label{app:transform-visualization}

To further understand what the learned linear transforms capture, we visualize
the deviation from identity, $\bm{W} - I$, for both the input and output
weight matrices learned on each of the 12 datasets, along with the
corresponding bias vectors $\bm{b}$. Since all transforms are initialized as
the identity matrix ($\bm{W} = I$, $\bm{b} = \bm{0}$), the deviation reflects
the PDE-specific adaptation discovered during training.

\begin{figure*}[h]
    \centering
    \includegraphics[width=0.95\textwidth]{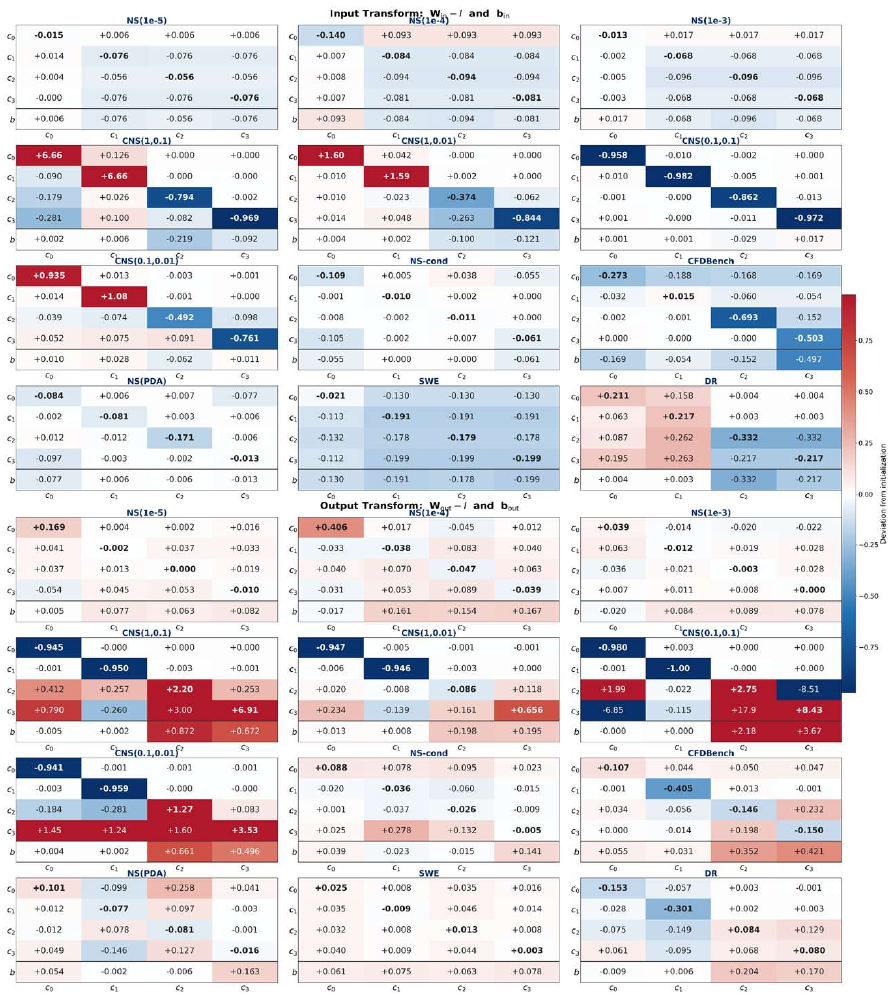}
    \caption{Visualization of the learned pointwise linear transforms across
    all 12 pre-training datasets. Each panel shows the deviation from identity
    $\bm{W} - I$ (top, $4 \times 4$ heatmap) and the learned bias $\bm{b}$
    (bottom, $1 \times 4$ bar) for both the input transform (upper half) and
    output transform (lower half). Blue indicates negative deviation, red
    indicates positive deviation, and white indicates near-zero (close to
    identity initialization). Color scale is clipped at the 95th percentile
    of absolute values for readability; exact values are annotated in each
    cell. Transforms within the same PDE family (e.g., the three FNO NS
    variants) share visually similar structure, while cross-family transforms
    are strikingly different.}
    \label{fig:transform-weights}
\end{figure*}

Figure~\ref{fig:transform-weights} displays the results, color-coded with a
diverging navy--white--red scheme (blue: negative deviation, red: positive,
white: near-zero). Several qualitative observations emerge:

\begin{enumerate}
\item \textbf{Intra-family similarity.}
The three FNO NS transforms (differing only in viscosity $\nu$) exhibit
similar weight patterns with small deviations (order $10^{-1}$), explaining
their mild cross-transfer degradation in Table~\ref{tab:full-cross-transfer}.
The four CNS transforms likewise share a common structural motif---strong
negative diagonal entries in $\bm{W}_{\text{in}}$---while differing in
magnitude and off-diagonal structure.

\item \textbf{Inter-family divergence.}
Transforms for fundamentally different PDE families exhibit markedly different
weight patterns. For instance, CNS(0.1,0.1) requires extreme deviations from
identity (entries exceeding $\pm 17$ in $\bm{W}_{\text{out}}$), whereas FNO
NS transforms stay within $\pm 0.15$. This two-order-of-magnitude difference
explains why applying the CNS(0.1,0.1) transform to incompressible-flow
datasets causes catastrophic failure ($14$--$80\times$ error increase).

\item \textbf{Asymmetry between input and output transforms.}
The input transforms $\bm{W}_{\text{in}}$ and output transforms
$\bm{W}_{\text{out}}$ learn qualitatively different patterns for the same
dataset, suggesting that the optimal input-space and output-space basis changes
serve complementary roles: $\bm{W}_{\text{in}}$ preprocesses the input into a
representation more amenable to the neural operator, while $\bm{W}_{\text{out}}$
maps the output back to the original basis.

\item \textbf{Modest deviations suffice for most PDEs.}
For 10 out of 12 datasets, the deviations remain within $\pm 1$, confirming
that a small change of basis is sufficient to meaningfully improve
approximation quality. The two CNS datasets with low Mach number
(CNS(0.1,0.1) and CNS(0.1,0.01)) are notable exceptions, requiring
substantially larger deviations.
\end{enumerate}

\subsubsection{Results at DPOT-Small Scale}
\label{app:motivation-small}

To verify that the PDE-specific nature of optimal linear transforms is not only a serederpity of the Tiny model scale, we repeat the full cross-PDE transfer
experiment using DPOT-Small. All three configurations
(baseline, jointly learned, and matched-frozen) are evaluated.

\begin{figure*}[h]
    \centering
    \includegraphics[width=\textwidth]{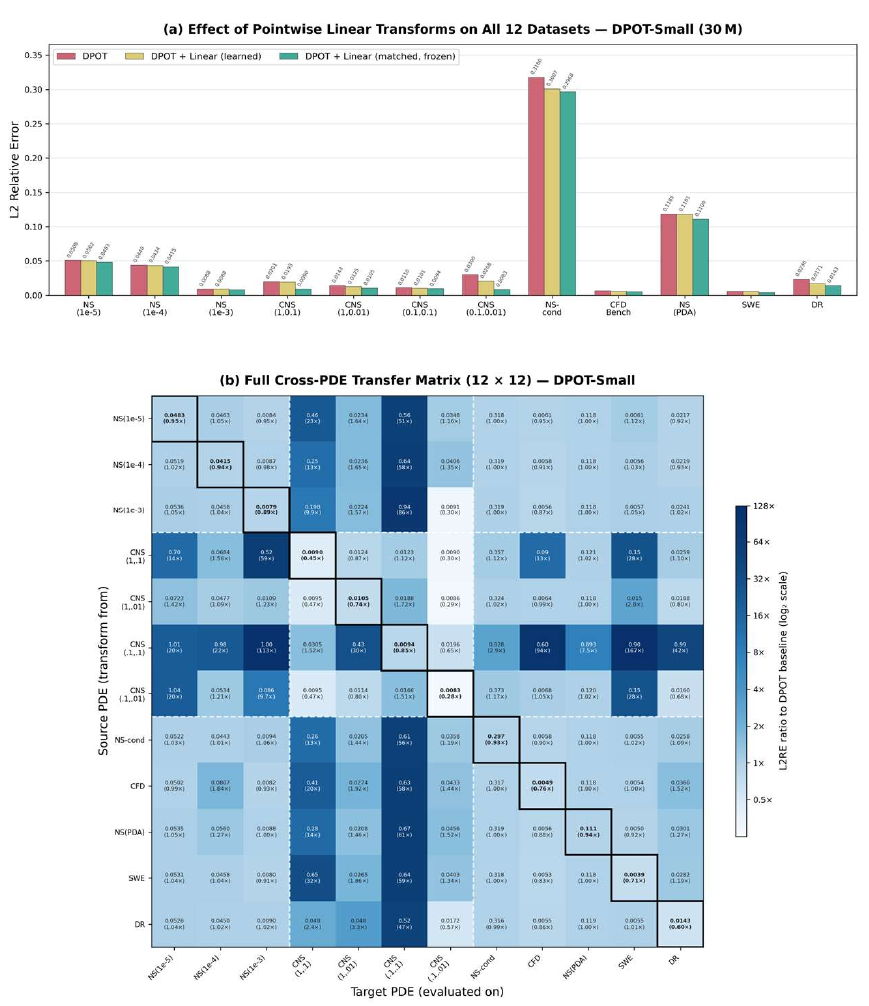}
    \caption{Full motivated experiment results on all 12 datasets (DPOT-Small,
    30\,M parameters).
    \textbf{(a)}~Bar chart comparing DPOT baseline (rose), DPOT with jointly
    trained linear layers (sand), and DPOT with matched frozen linear
    transforms (teal). The matched frozen configuration yields the lowest
    error on every dataset, consistently outperforming both the baseline
    and the jointly learned transform.
    \textbf{(b)}~$12 \times 12$ cross-PDE transfer matrix. Diagonal entries
    (black borders, light blue) show consistent improvement; off-diagonal
    entries reveal that mismatched transforms cause degradation up to
    $167\times$, even more extreme than at the Tiny scale. Dashed lines
    separate the FNO-NS, PDEBench-CNS, and remaining dataset groups.}
    \label{fig:motivation-all-small}
\end{figure*}

Table~\ref{tab:cross-transfer-small} presents the complete $14 \times 12$
transfer matrix. Figure~\ref{fig:motivation-all-small} provides the
corresponding visual summary.

\begin{table*}[h]
\centering
\caption{Full cross-PDE transfer matrix (L2 Relative Error, DPOT-Small, 30\,M).
Rows 1--2: baselines. Rows 3--14: DPOT with a frozen linear transform from
the source dataset indicated in the row label. Diagonal entries
(\colorbox{tableblue}{blue}) correspond to matched source-target pairs.
Lower is better.}
\label{tab:cross-transfer-small}
\resizebox{\textwidth}{!}{%
\setlength{\tabcolsep}{3pt}
\renewcommand{\arraystretch}{1.15}
\begin{tabular}{l|ccc|cccc|c|c|c|c|c}
\toprule
\multirow{2}{*}{Configuration} & \multicolumn{3}{c|}{FNO NS ($\nu$)} & \multicolumn{4}{c|}{PDEBench CNS ($\eta$,$\zeta$)} & PDEBench & PDEBench & PDEArena & PDEArena & CFD \\
 & 1e-5 & 1e-4 & 1e-3 & (1,0.1) & (1,0.01) & (0.1,0.1) & (0.1,0.01) & DR & SWE & NS & NS-cond & Bench \\
\midrule
DPOT (baseline) & 0.0508 & 0.0440 & 0.0088 & 0.0201 & 0.0143 & 0.0110 & 0.0300 & 0.0236 & 0.0054 & 0.1185 & 0.3180 & 0.0064 \\
DPOT+Linear & 0.0502 & 0.0434 & 0.0088 & 0.0193 & 0.0125 & 0.0101 & 0.0208 & 0.0171 & 0.0053 & 0.1181 & 0.3007 & 0.0053 \\
\midrule
Fixed: NS(1e-5) & \cellcolor{tableblue}\textbf{0.0483} & 0.0463 & 0.0084 & 0.4646 & 0.0234 & 0.5562 & 0.0348 & 0.0217 & 0.0061 & 0.1180 & 0.3184 & 0.0061 \\
Fixed: NS(1e-4) & 0.0519 & \cellcolor{tableblue}\textbf{0.0415} & 0.0087 & 0.2528 & 0.0236 & 0.6383 & 0.0406 & 0.0219 & 0.0056 & 0.1184 & 0.3186 & 0.0058 \\
Fixed: NS(1e-3) & 0.0536 & 0.0458 & \cellcolor{tableblue}\textbf{0.0079} & 0.1980 & 0.0224 & 0.9432 & 0.0091 & 0.0241 & 0.0057 & 0.1183 & 0.3189 & 0.0056 \\
Fixed: CNS(1,0.1) & 0.7038 & 0.0684 & 0.5221 & \cellcolor{tableblue}\textbf{0.0090} & 0.0124 & 0.0123 & 0.0090 & 0.0259 & 0.1534 & 0.1213 & 0.3572 & 0.0856 \\
Fixed: CNS(1,0.01) & 0.0723 & 0.0477 & 0.0109 & 0.0095 & \cellcolor{tableblue}\textbf{0.0105} & 0.0188 & 0.0086 & 0.0188 & 0.0152 & 0.1184 & 0.3236 & 0.0064 \\
Fixed: CNS(0.1,0.1) & 1.0067 & 0.9758 & 1.0002 & 0.0305 & 0.4323 & \cellcolor{tableblue}\textbf{0.0094} & 0.0196 & 0.9949 & 0.9044 & 0.8928 & 0.9277 & 0.6033 \\
Fixed: CNS(0.1,0.01) & 1.0412 & 0.0534 & 0.0857 & 0.0095 & 0.0114 & 0.0166 & \cellcolor{tableblue}\textbf{0.0083} & 0.0160 & 0.1511 & 0.1203 & 0.3727 & 0.0068 \\
Fixed: DR & 0.0526 & 0.0451 & 0.0090 & 0.0477 & 0.0477 & 0.5186 & 0.0172 & \cellcolor{tableblue}\textbf{0.0143} & 0.0055 & 0.1188 & 0.3158 & 0.0055 \\
Fixed: SWE & 0.0531 & 0.0458 & 0.0081 & 0.6454 & 0.0265 & 0.6439 & 0.0404 & 0.0282 & \cellcolor{tableblue}\textbf{0.0039} & 0.1185 & 0.3179 & 0.0054 \\
Fixed: NS(PDA) & 0.0535 & 0.0560 & 0.0088 & 0.2772 & 0.0208 & 0.6658 & 0.0456 & 0.0301 & 0.0050 & \cellcolor{tableblue}\textbf{0.1109} & 0.3186 & 0.0056 \\
Fixed: NS-cond & 0.0522 & 0.0443 & 0.0094 & 0.2640 & 0.0206 & 0.6099 & 0.0358 & 0.0258 & 0.0055 & 0.1181 & \cellcolor{tableblue}\textbf{0.2968} & 0.0058 \\
Fixed: CFDBench & 0.0502 & 0.0807 & 0.0082 & 0.4093 & 0.0274 & 0.6344 & 0.0433 & 0.0360 & 0.0054 & 0.1183 & 0.3170 & \cellcolor{tableblue}\textbf{0.0049} \\

\bottomrule
\end{tabular}}
\end{table*}

\paragraph{Analysis.}
The DPOT-Small results reinforce and extend the key findings from DPOT-Tiny:

\begin{enumerate}

\item \textbf{Jointly learned linear transforms universally help.}
Comparing the first two rows of Table~\ref{tab:cross-transfer-small},
DPOT+Linear improves over the DPOT baseline on all 12 datasets despite adding
only 40 parameters. The gains range from marginal on FNO NS datasets
($\sim$1\%) to substantial on DR (28\%), CNS(0.1,0.01) (31\%), and
NS-cond (5\%). This confirms that the benefit of PDE-specific basis changes
extends to larger model scales.

\item \textbf{Matched frozen transforms consistently outperform joint
optimization.}
All 12 diagonal entries are lower than both the DPOT baseline and the
DPOT+Linear values. Improvements over baseline range from 5\% on NS(1e-5) to
72\% on CNS(0.1,0.01). The CNS family benefits most strongly (15--72\%),
while FNO NS datasets show more modest gains (5--11\%). Notably, the matched
frozen transform outperforms joint optimization on every dataset, with
especially large margins on CNS(1,0.1) (53\% vs.\ 4\%), SWE (29\% vs.\ 1\%),
and DR (40\% vs.\ 28\%). This demonstrates that a correct PDE-specific
basis---when provided to the backbone---is more effective than allowing the
backbone and transform to co-adapt, consistent with the Tiny-scale finding.

\item \textbf{Mismatched degradation remains severe---and can be even more
extreme at larger scale.}
The most catastrophic mismatch at DPOT-Small is CNS(0.1,0.1)$\to$SWE:
$167\times$ baseline error, exceeding the corresponding Tiny-scale degradation
of $99\times$. Similarly, CNS(0.1,0.1)$\to$NS(1e-3) reaches $113\times$ and
CNS(0.1,0.1)$\to$CFDBench reaches $94\times$. The CNS(0.1,0.1) transform
remains the most disruptive, consistent with the Tiny-scale observation that
low-Mach-number compressible-flow transforms require extreme deviations from
identity.

\item \textbf{Intra-family transferability persists.}
Within the FNO NS family, cross-transfer degradation remains mild ($<\!1.3\times$
baseline). The four CNS datasets also exhibit partial mutual transferability,
though with more variation than at the Tiny scale. This confirms that PDEs
sharing similar mathematical structure benefit from similar basis
transformations regardless of model scale.

\item \textbf{Scale-dependent nuances.}
Compared to DPOT-Tiny, the diagonal improvements at DPOT-Small are more
variable: FNO NS datasets show smaller gains (5--11\% vs.\ 17--19\% at Tiny),
while CNS(0.1,0.01) shows a larger gain (72\% vs.\ 41\% at Tiny). This
suggests that increased model capacity partially compensates for the lack of
PDE-specific transformations on ``easier'' datasets, but remains insufficient
for more complex PDE families---further motivating an adaptive architecture
that can automatically calibrate its transformations.

\end{enumerate}

\paragraph{Visualization of learned transforms at small scale.}

\begin{figure*}[h]
    \centering
    \includegraphics[width=0.95\textwidth]{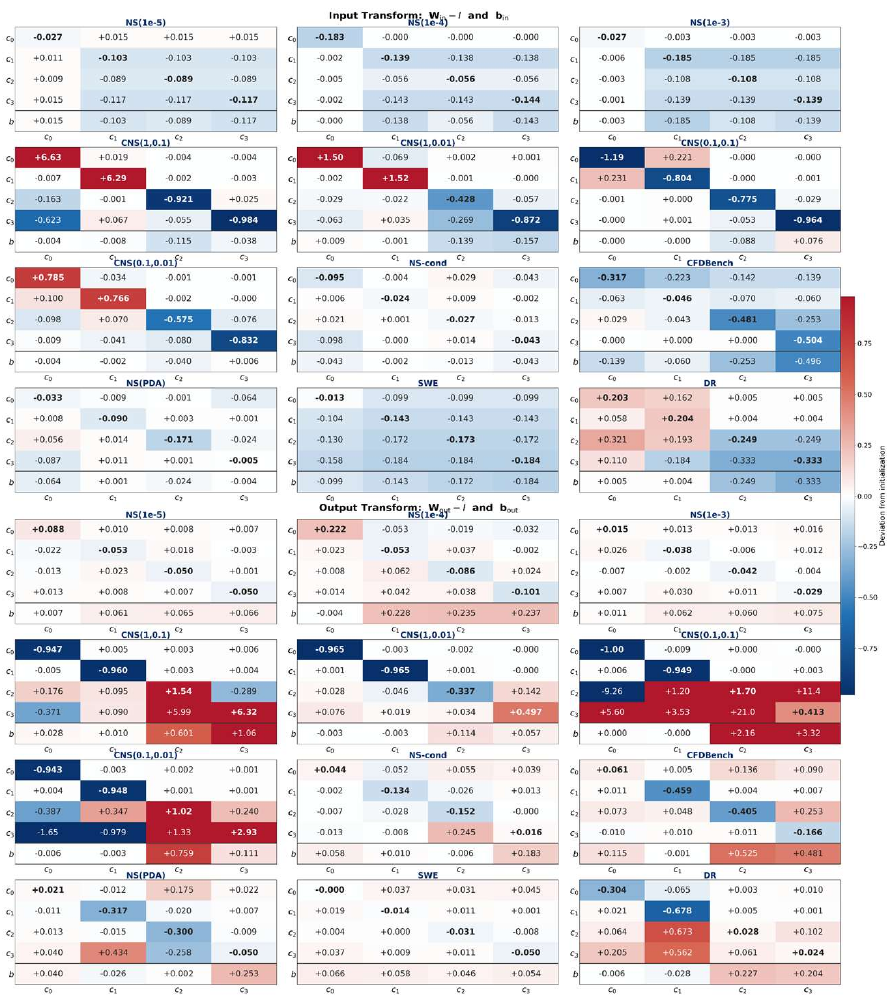}
    \caption{Visualization of the learned pointwise linear transforms at
    DPOT-Small scale (30\,M parameters). Format identical to
    Figure~\ref{fig:transform-weights}. The qualitative structure of the
    transforms is preserved across model scales: FNO NS transforms remain
    near-identity, CNS transforms share a common motif, and low-Mach-number
    CNS datasets exhibit the largest deviations.}
    \label{fig:transform-weights-small}
\end{figure*}

Figure~\ref{fig:transform-weights-small} visualizes the learned linear
transforms at the DPOT-Small scale. The qualitative patterns are consistent
with the Tiny-scale results (Figure~\ref{fig:transform-weights}): intra-family
similarity, inter-family divergence, and the extreme deviations required by
low-Mach-number CNS datasets are all preserved at the larger model scale.

\subsubsection{Results at DPOT-Medium Scale}
\label{app:motivation-medium}

We further extend the cross-PDE transfer experiment to DPOT-Medium (122\,M
parameters), the largest model scale in our study. All three configurations
(baseline, jointly learned, and matched-frozen) are evaluated.

\begin{figure*}[h]
    \centering
    \includegraphics[width=\textwidth]{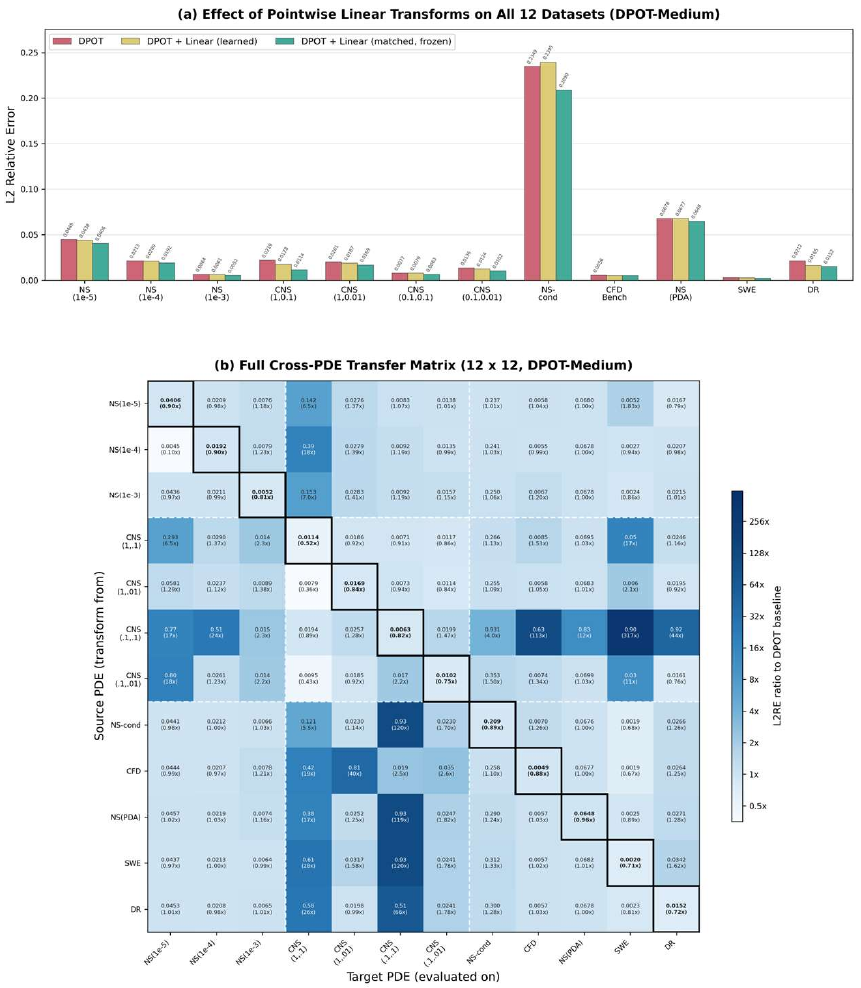}
    \caption{Full motivated experiment results on all 12 datasets (DPOT-Medium,
    122\,M parameters).
    \textbf{(a)}~Bar chart comparing DPOT baseline (rose), DPOT with jointly
    trained linear layers (sand), and DPOT with matched frozen linear
    transforms (teal). The matched frozen configuration yields the lowest
    error on 11 of 12 datasets, consistently outperforming both the baseline
    and the jointly learned transform.
    \textbf{(b)}~$12 \times 12$ cross-PDE transfer matrix. Diagonal entries
    (black borders, light blue) show consistent improvement; off-diagonal
    entries reveal that mismatched transforms cause degradation up to
    $317\times$, the most extreme across all three model scales. Dashed lines
    separate the FNO-NS, PDEBench-CNS, and remaining dataset groups.}
    \label{fig:motivation-all-medium}
\end{figure*}

Table~\ref{tab:cross-transfer-medium} presents the complete $14 \times 12$
transfer matrix. Figure~\ref{fig:motivation-all-medium} provides the
corresponding visual summary.

\begin{table*}[h]
\centering
\caption{Full cross-PDE transfer matrix (L2 Relative Error, DPOT-Medium,
122\,M). Rows 1--2: baselines. Rows 3--14: DPOT with a frozen linear
transform from the source dataset indicated in the row label. Diagonal entries
(\colorbox{tableblue}{blue}) correspond to matched source-target pairs.
Lower is better.}
\label{tab:cross-transfer-medium}
\resizebox{\textwidth}{!}{%
\setlength{\tabcolsep}{3pt}
\renewcommand{\arraystretch}{1.15}
\begin{tabular}{l|ccc|cccc|c|c|c|c|c}
\toprule
\multirow{2}{*}{Configuration} & \multicolumn{3}{c|}{FNO NS ($\nu$)} & \multicolumn{4}{c|}{PDEBench CNS ($\eta$,$\zeta$)} & PDEBench & PDEBench & PDEArena & PDEArena & CFD \\
 & 1e-5 & 1e-4 & 1e-3 & (1,0.1) & (1,0.01) & (0.1,0.1) & (0.1,0.01) & DR & SWE & NS & NS-cond & Bench \\
\midrule
DPOT (baseline) & 0.0449 & 0.0213 & 0.0064 & 0.0219 & 0.0201 & 0.0078 & 0.0136 & 0.0212 & 0.0028 & 0.0678 & 0.2350 & 0.0056 \\
DPOT+Linear & 0.0438 & 0.0209 & 0.0061 & 0.0178 & 0.0187 & 0.0079 & 0.0124 & 0.0165 & 0.0027 & 0.0677 & 0.2395 & 0.0049 \\
\midrule
Fixed: NS(1e-5) & \cellcolor{tableblue}\textbf{0.0406} & 0.0209 & 0.0076 & 0.1416 & 0.0276 & 0.0083 & 0.0138 & 0.0167 & 0.0052 & 0.0680 & 0.2373 & 0.0058 \\
Fixed: NS(1e-4) & 0.0045 & \cellcolor{tableblue}\textbf{0.0192} & 0.0079 & 0.3854 & 0.0279 & 0.0093 & 0.0135 & 0.0207 & 0.0027 & 0.0678 & 0.2411 & 0.0055 \\
Fixed: NS(1e-3) & 0.0436 & 0.0211 & \cellcolor{tableblue}\textbf{0.0052} & 0.1529 & 0.0283 & 0.0092 & 0.0157 & 0.0215 & 0.0024 & 0.0679 & 0.2498 & 0.0067 \\
Fixed: CNS(1,0.1) & 0.2928 & 0.0290 & 0.0145 & \cellcolor{tableblue}\textbf{0.0114} & 0.0186 & 0.0071 & 0.0117 & 0.0246 & 0.0485 & 0.0695 & 0.2661 & 0.0085 \\
Fixed: CNS(1,0.01) & 0.0581 & 0.0237 & 0.0089 & 0.0079 & \cellcolor{tableblue}\textbf{0.0170} & 0.0073 & 0.0114 & 0.0195 & 0.0061 & 0.0683 & 0.2553 & 0.0058 \\
Fixed: CNS(0.1,0.1) & 0.7733 & 0.5091 & 0.0150 & 0.0194 & 0.0257 & \cellcolor{tableblue}\textbf{0.0063} & 0.0199 & 0.9232 & 0.9007 & 0.8306 & 0.9308 & 0.6272 \\
Fixed: CNS(0.1,0.01) & 0.8037 & 0.0261 & 0.0139 & 0.0095 & 0.0185 & 0.0168 & \cellcolor{tableblue}\textbf{0.0102} & 0.0161 & 0.0315 & 0.0699 & 0.3534 & 0.0074 \\
Fixed: DR & 0.0453 & 0.0208 & 0.0065 & 0.5782 & 0.0198 & 0.5131 & 0.0241 & \cellcolor{tableblue}\textbf{0.0152} & 0.0023 & 0.0679 & 0.3001 & 0.0057 \\
Fixed: SWE & 0.0437 & 0.0213 & 0.0064 & 0.6065 & 0.0317 & 0.9280 & 0.0241 & 0.0342 & \cellcolor{tableblue}\textbf{0.0020} & 0.0682 & 0.3118 & 0.0057 \\
Fixed: NS(PDA) & 0.0457 & 0.0219 & 0.0074 & 0.3819 & 0.0252 & 0.9259 & 0.0247 & 0.0271 & 0.0025 & \cellcolor{tableblue}\textbf{0.0648} & 0.2904 & 0.0057 \\
Fixed: NS-cond & 0.0441 & 0.0212 & 0.0066 & 0.1208 & 0.0230 & 0.9273 & 0.0230 & 0.0266 & 0.0019 & 0.0676 & \cellcolor{tableblue}\textbf{0.2090} & 0.0070 \\
Fixed: CFDBench & 0.0444 & 0.0207 & 0.0078 & 0.4208 & 0.8062 & 0.0191 & 0.0354 & 0.0264 & 0.0019 & 0.0677 & 0.2576 & \cellcolor{tableblue}\textbf{0.0049} \\

\bottomrule
\end{tabular}}
\end{table*}

\paragraph{Analysis.}
The DPOT-Medium results reinforce and extend the key findings from smaller
scales:

\begin{enumerate}

\item \textbf{Jointly learned linear transforms broadly help.}
Comparing the first two rows of Table~\ref{tab:cross-transfer-medium},
DPOT+Linear improves over the DPOT baseline on 10 of 12 datasets. The gains
range from marginal on NS(PDA) ($< 1\%$) to substantial on DR (22\%) and
CNS(1,0.1) (19\%). Two datasets---CNS(0.1,0.1) and NS-cond---show slight
regression ($< 2.5\%$), suggesting that at the Medium scale, joint
optimization of the linear transform and a large backbone can occasionally
encounter optimization difficulties on certain PDE families.

\item \textbf{Matched frozen transforms remain universally beneficial.}
All 12 diagonal entries in Table~\ref{tab:cross-transfer-medium} are lower
than the corresponding DPOT baseline values, with improvements ranging from
4.5\% on NS(PDA) to 48\% on CNS(1,0.1). The most pronounced gains appear
on SWE ($0.0020$ vs.\ baseline $0.0028$, a 29\% reduction) and DR
($0.0152$ vs.\ $0.0212$, also 28\%), consistent with smaller scales.

\item \textbf{Matched frozen transforms outperform jointly learned ones on
nearly all datasets.}
The matched frozen transform achieves lower L2RE than DPOT+Linear on 11 of 12
datasets. The sole exception is CFDBench, where the difference is negligible
($0.0049$ vs.\ $0.0049$, a $0.6\%$ gap). This reinforces the conclusion that
a pre-optimized, PDE-specific basis change is more effective than joint
end-to-end optimization, and that decoupling the transform and backbone
training is beneficial.

\item \textbf{Mismatched degradation is the most extreme at this scale.}
The worst mismatch is CNS(0.1,0.1)$\to$SWE: $317\times$ baseline error,
surpassing the $167\times$ at Small and $99\times$ at Tiny. Other severe
cases include CNS(0.1,0.1)$\to$CFDBench ($113\times$) and
CNS(0.1,0.1)$\to$DR ($44\times$). The CNS(0.1,0.1) transform remains the
most disruptive across all scales.

\item \textbf{Remarkably strong intra-family transfer.}
The NS(1e-4) transform applied to NS(1e-5) achieves $0.0045$---a
$10\times$ improvement over the baseline $0.0449$ and even better than the
matched diagonal entry ($0.0406$). This suggests that within the FNO NS
family at the Medium scale, cross-viscosity transfer can be extremely
beneficial, potentially because the NS(1e-4) transform captures a more
general incompressible-flow basis that also benefits the harder NS(1e-5)
regime.

\item \textbf{Consistent findings across three scales.}
The core conclusions---matched frozen transforms universally help, mismatched
transforms can be catastrophic, and the optimal transformation is strongly
PDE-dependent---hold consistently across DPOT-Tiny (7\,M), DPOT-Small
(30\,M), and DPOT-Medium (122\,M). This scale-invariance further motivates
the AOT block in \method, which provides an adaptive, input-dependent
operator transformation that goes beyond a single shared linear layer.

\end{enumerate}

\begin{figure*}[h]
    \centering
    \includegraphics[width=0.95\textwidth]{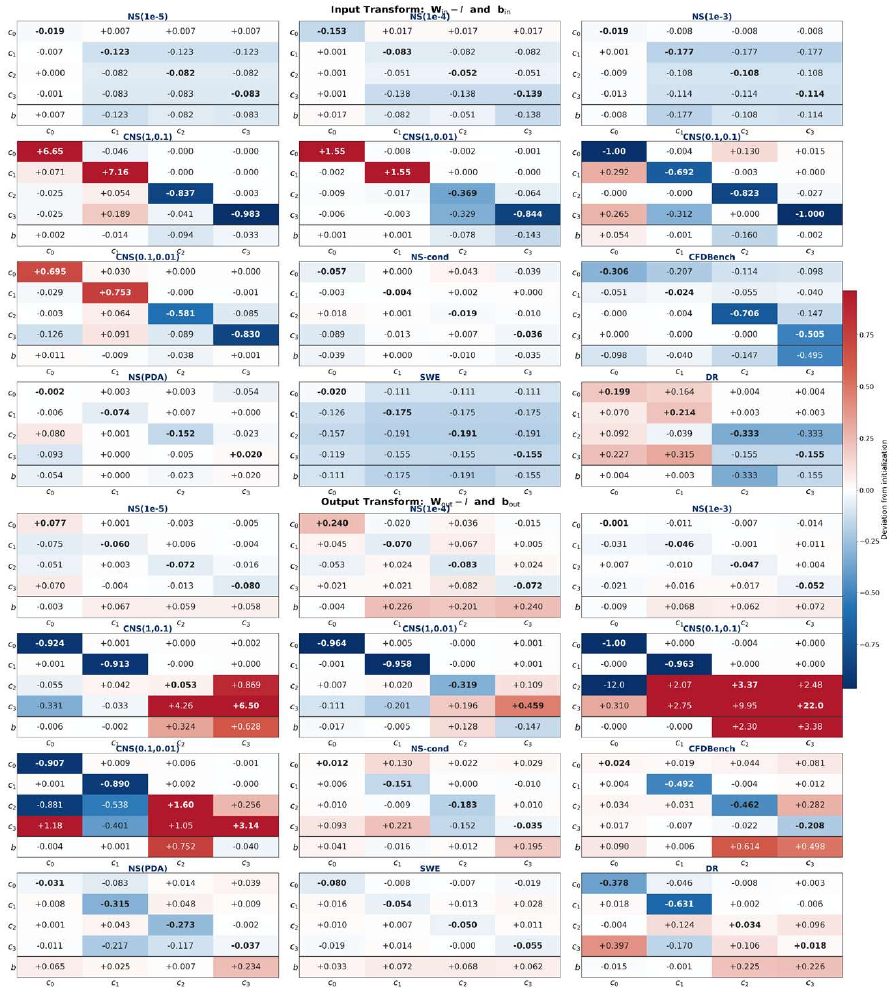}
    \caption{Visualization of the learned pointwise linear transforms at
    DPOT-Medium scale (122\,M parameters). Format identical to
    Figure~\ref{fig:transform-weights}. The qualitative structure of the
    transforms is preserved across model scales: FNO NS transforms remain
    near-identity, CNS transforms share a common motif, and low-Mach-number
    CNS datasets exhibit the largest deviations.}
    \label{fig:transform-weights-medium}
\end{figure*}

\paragraph{Visualization of learned transforms at medium scale.}Figure~\ref{fig:transform-weights-medium} visualizes the learned linear
  transforms at the DPOT-Medium scale. The qualitative patterns observed at
  smaller scales are preserved and, in several cases, amplified:

  \begin{enumerate}

  \item \textbf{Intra-family similarity persists.}
  The three FNO NS input transforms share a consistent pattern of small
  negative diagonal entries (order $10^{-1}$ to $10^{-2}$), with NS(1e-4)
  showing slightly larger deviations than NS(1e-5) and NS(1e-3). The four
  CNS transforms likewise share a common structural motif---strong diagonal
  modifications in $\bm{W}_{\text{in}}$---while differing in magnitude.

  \item \textbf{CNS deviations are even more extreme at the Medium scale.}
  The CNS(1,0.1) input transform exhibits diagonal entries as large as $+7.16$,
  and the CNS(0.1,0.1) output transform reaches $+22.0$ in a single entry---the
  largest deviation observed across all three model scales (compared to
  $\sim\!\pm 17$ at Tiny). This growing magnitude at larger scale is consistent
  with the more extreme mismatch degradation observed in
  Table~\ref{tab:cross-transfer-medium} ($317\times$), and suggests that
  larger models develop increasingly specialized internal representations
  that diverge further from the default basis for complex PDE families.

  \item \textbf{Clear channel-role structure in CNS transforms.}
  For all four CNS datasets, the input transforms show a striking split: the
  first two channels ($c_0, c_1$, corresponding to momentum components) receive
  large positive diagonal entries, while the last two ($c_2, c_3$, corresponding
  to density and energy) receive negative ones. The output transforms exhibit
  the complementary pattern, with extreme off-diagonal entries concentrated
  in the $c_2$--$c_3$ block. This suggests the learned transforms perform a
  physically meaningful rescaling that separates the dynamical roles of
  different conserved quantities.

  \item \textbf{Near-identity transforms for simpler PDEs.}
  NS-cond, SWE, and NS(1e-3) remain very close to identity at both input
  and output (deviations $< 0.2$), indicating that the default representation
  is already well-suited for these PDE families. In contrast, CFDBench and DR
  show moderate deviations (order $10^{-1}$), with DR exhibiting a distinct
  positive-negative split on the diagonal similar to, but much milder than,
  the CNS pattern.

  \item \textbf{Asymmetry between input and output transforms.}
  The input and output weight matrices learn qualitatively different structures
  for the same dataset. For instance, CNS(1,0.1) input features large positive
  diagonal entries ($+6.65$, $+7.16$), while its output features near-complete
  suppression of the first two channels ($-0.924$, $-0.913$) with extreme
  amplification in the last two rows ($+4.26$, $+6.50$). This complementary
  structure confirms that $\bm{W}_{\text{in}}$ and $\bm{W}_{\text{out}}$
  serve distinct roles: the input transform rescales the representation into
  a basis amenable to the neural operator, while the output transform maps
  back with a different, non-trivial projection.

  \end{enumerate}

\subsection{Pre-trained Results of \method at Large Scale}

\begin{table*}[t]
\vspace{-1em}
\centering
\caption{Results of \method at the Large scale.}\label{tb-app-large}
\resizebox{\textwidth}{!}{%
\setlength{\tabcolsep}{4pt}
\renewcommand{\arraystretch}{1.3}
\begin{tabular}{ll|cccccccccccc}
\toprule
L2RE & Params & \multicolumn{3}{c|}{FNO-$\nu$} & \multicolumn{6}{c|}{PDEBench CNS-$(\eta, \zeta)$,DR,SWE} & \multicolumn{2}{c|}{PDEArena} & CFDBench \\
Subset & -- & 1e-5 & 1e-4 & \multicolumn{1}{c|}{1e-3} & 1,0.1 & 1,0.01 & 0.1,0.1 & 0.1,0.01 & DR & \multicolumn{1}{c|}{SWE} & NS & \multicolumn{1}{c|}{NS-cond} & -- \\ \hline

\multicolumn{2}{c|}{Pre-trained} &  &  &  &  &  &  &  &  &  &  &  &  \\

{DPOT-L} & 500M & 0.05500 & 0.02740 & 0.00528 & 0.01000 & 0.02160 & 0.00872 & 0.01150 & 0.02320 & 0.00233 & 0.07980 & 0.24000 & 0.00650 \\

\textcolor{dkred}{\method-L} & 515M & \textbf{0.02934} & \textbf{0.02113} & \textbf{0.00285} & \textbf{0.00862} & \textbf{0.01804} & \textbf{0.00754} & \textbf{0.01000} & \textbf{0.02199} & \textbf{0.00197} & \textbf{0.06372} & \textbf{0.20111} & \textbf{0.00586} \\

\bottomrule
\end{tabular}}

\vspace{-1em}
\end{table*}

To examine whether the benefits of \method persist when the backbone is
scaled up well beyond the configurations reported in
Table~\ref{tb-main}, we additionally pre-train both DPOT and \method
at the Large scale---roughly $4\times$ larger than the M scale---on the same
corpus and report zero-shot L2RE in Table~\ref{tb-app-large}. The
comparison is anchored on the matched-capacity pair DPOT-L (500M)
vs.\ \method-L (515M): \method-L introduces only $\sim\!3\%$ extra
parameters on top of DPOT-L, mirroring the parameter overhead at the S
scale, so any quality gap can be attributed to the adaptive operator
transformation rather than to additional capacity.

\textbf{Zero-shot results at the large scale.}
\method-L delivers consistent improvements over DPOT-L on \emph{every}
dataset. \method-L reduces L2RE by up to
$46.7\%$ on FNO-$\nu\!=\!10^{-5}$ and $46.0\%$ on FNO-$\nu\!=\!10^{-3}$.
The gains are particularly pronounced on the FNO Navier--Stokes subsets,
which span a wide viscosity range and therefore benefit most from
input-dependent operator transformations. 
Combined
with the main-table results, the large-scale experiment confirms that
\method's improvements are an effect of the adaptive operator
transformation rather than of model capacity, and that the design scales
gracefully as the pre-trained backbone is enlarged.

\subsection{Scaling Experiments}\label{sec:scaling}

\begin{wrapfigure}{r}{0.5\textwidth}
    \vspace{-1.2em}
    \centering
    \includegraphics[width=\linewidth]{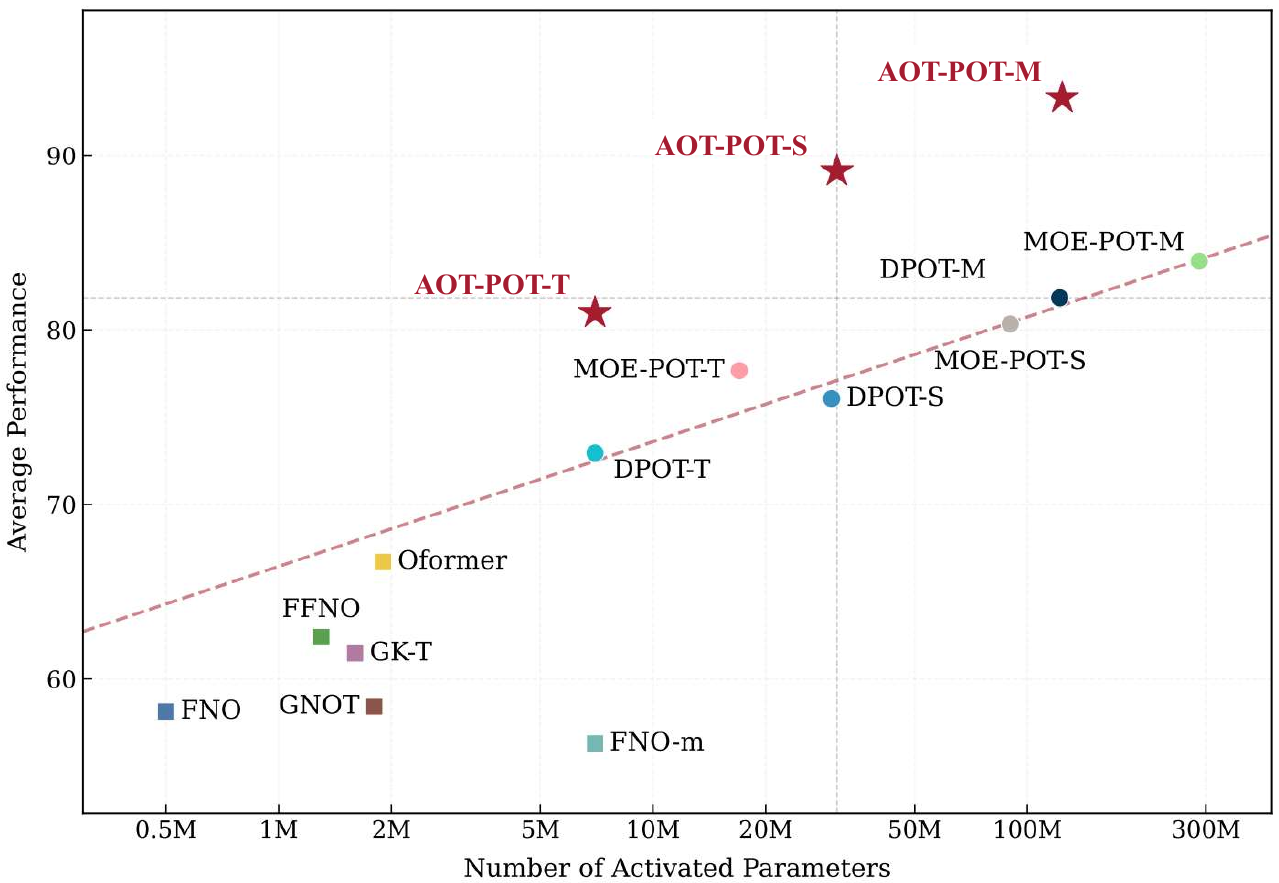}
    \caption{\textbf{Scaling experiments.} Average performance across 12
    pre-training datasets versus activated parameters. The dashed line is the power-law trend fitted to DPOT. Performance is
    $50\cdot\overline{-\log_{10}(\mathrm{L2RE})}$ (higher is better;
    $100$ = L2RE $=0.01$ everywhere).}
    \label{fig-scaling}
    \vspace{-1em}
\end{wrapfigure}

Figure~\ref{fig-scaling} compares \method against
DPOT~\citep{hao2024dpot}, MoE-POT~\citep{wang2025mixture}, and
single-dataset neural operators (FNO, FFNO, GK-T, GNOT, Oformer, FNO-m)
on average zero-shot performance over all 12 datasets versus activated
parameters---those contributing to each forward pass. \method sits above
the DPOT scaling trend at all three scales, with the advantage most
pronounced at small scale: the smallest \method variant rivals the
largest DPOT, indicating a substantial ``parameter-free'' improvement
from the AOT block. \method also clearly outperforms the sparse MoE-POT
while activating far fewer parameters per forward pass. Together, these
results confirm that adaptive operator transformations are a more
effective inductive bias for scaling PDE foundation models than dense
expansion or sparse routing.

\subsection{Ablation at Additional Scales}\label{app:ablation}

In this section, we provide the AOT block ablation results of our \method
at the tiny and medium scales in Tables~\ref{tab-ablation-tiny}
and~\ref{tab-ablation-medium}, complementing the small-scale results in
Table~\ref{tab-ablation}.

\begin{table}[h]
\centering
\caption{Ablation study of the three mappings of the AOT block at the
\textbf{tiny} scale (7M parameters). The full row reports the pretrained
\method-T result from Table~\ref{tb-main}. We \textbf{bold} the best result(s) in each column.}
\label{tab-ablation-tiny}
\resizebox{\textwidth}{!}{%
\setlength{\tabcolsep}{5pt}
\renewcommand{\arraystretch}{1.3}
\begin{tabular}{l ccc cccccc cc c}
\toprule
 & \multicolumn{3}{c}{FNO-$\nu$} & \multicolumn{6}{c}{PDEBench CNS-$(\eta, \zeta)$,DR,SWE} & \multicolumn{2}{c}{PDEArena} & CFDBench \\
 \cmidrule(lr){2-4} \cmidrule(lr){5-10} \cmidrule(lr){11-12} \cmidrule(lr){13-13}
Variant & 1e-5 & 1e-4 & 1e-3 & 1,0.1 & 1,0.01 & 0.1,0.1 & 0.1,0.01 & DR & SWE & NS & NS-cond & -- \\
\midrule
w/o $\bm{a}$ & 0.06581 & 0.04798 & 0.00632 & 0.01143 & 0.02834 & 0.00966 & 0.01348 & 0.02526 & 0.00441 & 0.10348 & 0.35807 & 0.00801 \\
w/o $\bm{d}$ & 0.06314 & 0.04458 & 0.00675 & 0.01181 & 0.02813 & 0.01036 & 0.01328 & 0.02465 & 0.00518 & 0.11233 & 0.35410 & 0.00883 \\
w/o $\bm{T}$ & 0.06369 & 0.04867 & 0.00659 & 0.01299 & 0.03068 & 0.00903 & 0.01467 & 0.02565 & 0.00471 & 0.11817 & 0.36728 & 0.00917 \\
\midrule
\textcolor{dkred}{\method-T (full)} & \textbf{0.06102} & \textbf{0.04384} & \textbf{0.00604} & \textbf{0.01102} & \textbf{0.02795} & \textbf{0.00880} & \textbf{0.01302} & \textbf{0.02413} & \textbf{0.00433} & \textbf{0.09846} & \textbf{0.34492} & \textbf{0.00776} \\
\bottomrule
\end{tabular}%
}
\end{table}

\begin{table}[h]
\centering
\caption{Ablation study of the three mappings of the AOT block at the
\textbf{medium} scale (124M parameters). The full row reports the pretrained
\method-M result from Table~\ref{tb-main}. We \textbf{bold} the best result(s) in each column.}
\label{tab-ablation-medium}
\resizebox{\textwidth}{!}{%
\setlength{\tabcolsep}{5pt}
\renewcommand{\arraystretch}{1.3}
\begin{tabular}{l ccc cccccc cc c}
\toprule
 & \multicolumn{3}{c}{FNO-$\nu$} & \multicolumn{6}{c}{PDEBench CNS-$(\eta, \zeta)$,DR,SWE} & \multicolumn{2}{c}{PDEArena} & CFDBench \\
 \cmidrule(lr){2-4} \cmidrule(lr){5-10} \cmidrule(lr){11-12} \cmidrule(lr){13-13}
Variant & 1e-5 & 1e-4 & 1e-3 & 1,0.1 & 1,0.01 & 0.1,0.1 & 0.1,0.01 & DR & SWE & NS & NS-cond & -- \\
\midrule
w/o $\bm{a}$ & 0.03199 & 0.02753 & 0.00244 & 0.00973 & 0.01781 & 0.00795 & 0.00964 & 0.02028 & 0.00211 & 0.08516 & 0.24904 & 0.00584 \\
w/o $\bm{d}$ & 0.03478 & 0.02900 & 0.00242 & 0.00942 & 0.01729 & 0.00886 & 0.00983 & 0.01910 & \textbf{0.00200} & 0.08379 & 0.25970 & 0.00557 \\
w/o $\bm{T}$ & 0.03136 & 0.02737 & 0.00276 & 0.00916 & 0.01790 & 0.00867 & 0.01078 & 0.01995 & 0.00224 & 0.08447 & 0.28587 & 0.00579 \\
\midrule
\textcolor{dkred}{\method-M (full)} & \textbf{0.03118} & \textbf{0.02566} & \textbf{0.00236} & \textbf{0.00906} & \textbf{0.01676} & \textbf{0.00788} & \textbf{0.00954} & \textbf{0.01860} & \textbf{0.00200} & \textbf{0.07751} & \textbf{0.24226} & \textbf{0.00543} \\
\bottomrule
\end{tabular}%
}
\end{table}

\paragraph{Analysis.}
At the {tiny} scale (Table~\ref{tab-ablation-tiny}), all three ablation
variants exhibit clear degradation relative to the full model, confirming that
no single mapping is redundant even at the smallest model capacity. Removing
$\bm{T}$ causes the largest performance drop on 8 out of 12
datasets, with particularly severe degradation on the CNS subsets (e.g.,
CNS(0.1,\,0.01) increases from 0.01302 to 0.01467, a 12.7\% relative increase)
and PDEArena NS-cond (0.34492 $\to$ 0.36728, +6.5\%). Interestingly,
removing $\bm{a}$ produces a larger error on FNO-$10^{-5}$
(0.06581 vs.\ 0.06369 for w/o~$\bm{T}$), suggesting that the
aggregation mechanism is especially important for highly turbulent flows where
selecting the right combination of stream information is critical.

At the {medium} scale (Table~\ref{tab-ablation-medium}), the absolute
performance gaps between variants narrow---reflecting the increased baseline
capacity of the 124M-parameter model---but the relative ordering remains the
same. Removing $\bm{T}$ still causes the largest degradation
on the majority of datasets, most strikingly on PDEArena NS-cond
(0.24226 $\to$ 0.28587, +18.0\%) and DR (0.01860 $\to$ 0.01995, +7.3\%). A notable
observation at this scale is that removing $\bm{d}$ produces
a relatively larger impact than removing $\bm{a}$ on several
datasets (e.g., FNO-$10^{-5}$: 0.03478 vs.\ 0.03199), reversing the pattern
seen at the Tiny scale. This suggests that as model depth increases (from 4
layers at Tiny to 12 at Medium), the redistribution mapping
$\bm{d}$ becomes more important for correctly routing
sub-layer outputs back into the multi-stream representation across many layers.

Across all three scales, the full \method achieves the best or tied-best result
on every dataset, validating the necessity of the complete three-mapping
design. The consistent dominance of $\bm{T}$ removal as
the most damaging ablation aligns with its central architectural role:
$\bm{T}$ is the operator-transformation kernel itself---it directly governs
inter-stream communication in the residual path, and it is the component
constrained to the Birkhoff polytope (Eq.~\eqref{eq:birkhoff}), which enables
the PDE-specific operator transformations analyzed in
Section~\ref{sec:interpretability}.

\subsection{Hyperparameter Sensitivity Experiment}\label{app:sensitivity}

We study the hyperparameter sensitivity of \method to three key AOT block
hyperparameters: the number of parallel streams $n$, the number of
Sinkhorn-Knopp iterations, and the gating scalar initialization $\alpha$ to provide a more comprehensive understanding of our \method.
For each experiment, we vary one hyperparameter while keeping all others at
their default values ($n=4$, 20 iterations, $\alpha=0.01$) for a fair comparison.

\paragraph{Number of streams $n$.}
 We compare $n \in \{2, 4, 8\}$ across all three
scales in Table~\ref{tab-sens-streams}.

\begin{table}[h]
\centering
\caption{Sensitivity to the number of parallel streams $n$. The $n=4$ row
reports the pretrained \method-(T/S/M) result from Table~\ref{tb-main}. We
\textbf{bold} the best result in each column per scale, respectively.}
\label{tab-sens-streams}
\resizebox{\textwidth}{!}{%
\setlength{\tabcolsep}{4pt}
\renewcommand{\arraystretch}{1.2}
\begin{tabular}{l|cccccccccccc}
\toprule
 & \multicolumn{3}{c|}{FNO-$\nu$} & \multicolumn{6}{c|}{PDEBench CNS-$(\eta, \zeta)$,DR,SWE} & \multicolumn{2}{c|}{PDEArena} & CFDBench \\
Variant & 1e-5 & 1e-4 & \multicolumn{1}{c|}{1e-3} & 1,0.1 & 1,0.01 & 0.1,0.1 & 0.1,0.01 & DR & \multicolumn{1}{c|}{SWE} & NS & \multicolumn{1}{c|}{NS-cond} & -- \\
\midrule
\multicolumn{13}{l}{\textit{Tiny (7M)}} \\
$n=2$ & 0.06580 & 0.04789 & 0.00614 & 0.01175 & 0.02872 & 0.00881 & 0.01392 & \textbf{0.02250} & \textbf{0.00426} & \textbf{0.09629} & \textbf{0.31862} & \textbf{0.00767} \\
$n=4$ & \textbf{0.06102} & \textbf{0.04384} & 0.00604 & \textbf{0.01102} & 0.02795 & 0.00880 & \textbf{0.01302} & 0.02413 & 0.00433 & 0.09846 & 0.34492 & 0.00776 \\
$n=8$ & 0.06438 & 0.04473 & \textbf{0.00588} & 0.01185 & \textbf{0.02689} & \textbf{0.00875} & 0.01362 & 0.02465 & 0.00470 & 0.10437 & 0.35908 & 0.00835 \\
\midrule
\multicolumn{13}{l}{\textit{Small (31M)}} \\
$n=2$ & 0.03940 & 0.03653 & 0.00294 & 0.00812 & 0.01974 & 0.00792 & 0.00963 & 0.01549 & 0.00271 & 0.07869 & 0.25496 & 0.00505 \\
$n=4$ & \textbf{0.03755} & \textbf{0.03326} & 0.00293 & \textbf{0.00767} & 0.02044 & 0.00833 & 0.00965 & \textbf{0.01461} & \textbf{0.00266} & \textbf{0.07842} & 0.26067 & \textbf{0.00493} \\
$n=8$ & 0.03808 & 0.03601 & \textbf{0.00283} & 0.00785 & \textbf{0.01934} & \textbf{0.00766} & \textbf{0.00922} & \textbf{0.01461} & 0.00281 & 0.08604 & \textbf{0.25116} & 0.00501 \\
\midrule
\multicolumn{13}{l}{\textit{Medium (124M)}} \\
$n=2$ & \textbf{0.03109} & \textbf{0.02550} & \textbf{0.00234} & \textbf{0.00906} & 0.01694 & 0.00792 & 0.00964 & 0.01868 & 0.00201 & 0.07800 & 0.24323 & 0.00547 \\
$n=4$ & 0.03118 & 0.02566 & 0.00236 & \textbf{0.00906} & \textbf{0.01676} & \textbf{0.00788} & \textbf{0.00954} & 0.01860 & 0.00200 & 0.07751 & 0.24226 & 0.00543 \\
$n=8$ & 0.03235 & 0.02599 & 0.00236 & 0.00910 & 0.01755 & 0.00798 & 0.00964 & \textbf{0.01835} & \textbf{0.00186} & \textbf{0.07633} & \textbf{0.24129} & \textbf{0.00536} \\
\bottomrule
\end{tabular}%
}
\end{table}

At the {tiny} scale, $n=4$ yields the best results on the FNO and
key CNS subsets that most directly stress the AOT block's mixing
capacity---FNO-$10^{-5}$ (0.06102 vs.\ 0.06580 for $n=2$ and
0.06438 for $n=8$), FNO-$10^{-4}$, CNS(1,\,0.1) (0.01102 vs.\
0.01175 for $n=2$), and CNS(0.1,\,0.01)---4 of 12 datasets in total.
$n=2$ trails on these key benchmarks, indicating that two streams
provide insufficient representational diversity for the
7M-parameter model to distinguish the mixing patterns required by
different PDE families; $n=2$ does retain a small edge on the
structurally simpler downstream benchmarks (DR, SWE, PDEArena NS
and NS-cond, and CFDBench, all within 1--7\% of $n=4$), where the
choice of stream count is less critical. Increasing to $n=8$ does
not improve over $n=4$ on the FNO/CNS subsets and even degrades
performance on FNO-$10^{-5}$ (0.06438) and PDEArena NS (0.10437),
likely because the larger $n$ introduces more AOT block parameters
($\bm{\phi}^{T} \in \mathbb{R}^{nC \times n^2}$ grows quadratically
in $n$) that the small model cannot effectively utilise, leading to
mild overfitting.

At the {small} scale, $n=4$ remains the overall best choice,
winning on 7 of 12 datasets (counting the tie with $n=8$ on
PDB-DR), including the critical FNO-$10^{-5}$ (0.03755) and
PDB-DR (0.01461). $n=8$ takes the lead on FNO-$10^{-3}$ (0.00283),
the three high-viscosity CNS subsets (CNS(1,\,0.01),
CNS(0.1,\,0.1), CNS(0.1,\,0.01)), and PDEArena NS-cond (0.25116);
$n=2$ no longer dominates any column. The gap between all three
variants is notably smaller than at the Tiny scale, suggesting that
the 31M-parameter backbone has sufficient capacity to partially
compensate for a suboptimal stream count.

At the {medium} scale, performance differences become minimal
across all three values of $n$, with no single setting dominating---
the maximum within-column spread stays below 6\% on every dataset
(7.5\% on PDB-SWE only). $n=2$ holds a slim edge on the FNO subsets
(FNO-$10^{-5}$: 0.03109 vs.\ 0.03118 for $n=4$ and 0.03235 for
$n=8$), $n=4$ leads on the four CNS subsets, and $n=8$ takes the
heavier downstream benchmarks (PDB-DR, PDB-SWE, PDEArena NS,
NS-cond, CFDBench). This fan-out indicates that the 124M-parameter
model can effectively learn PDE-specific transformations even with
fewer streams, as its deeper architecture (12 layers) provides
alternative pathways for information mixing. We adopt $n=4$ as the
default for all scales, as it provides a good balance between
representational richness and parameter efficiency on the
AOT-stress (FNO/CNS) subsets without requiring scale-specific
tuning.

\paragraph{Number of Sinkhorn-Knopp iterations.}
 We compare 10, 20, and 40 Sinkhorn-Knopp iterations of our \method in Table~\ref{tab-sens-sinkhorn}.

\begin{table}[h]
\centering
\caption{Sensitivity of \method to the number of Sinkhorn-Knopp iterations
at all three scales. The \texttt{20 iter} row reports the pretrained
\method-(T/S/M) result from Table~\ref{tb-main} (the default configuration).
We \textbf{bold} the best result in each column per scale, respectively.}
\label{tab-sens-sinkhorn}
\resizebox{\textwidth}{!}{%
\setlength{\tabcolsep}{4pt}
\renewcommand{\arraystretch}{1.2}
\begin{tabular}{l|cccccccccccc}
\toprule
 & \multicolumn{3}{c|}{FNO-$\nu$} & \multicolumn{6}{c|}{PDEBench CNS-$(\eta, \zeta)$,DR,SWE} & \multicolumn{2}{c|}{PDEArena} & CFDBench \\
Variant & 1e-5 & 1e-4 & \multicolumn{1}{c|}{1e-3} & 1,0.1 & 1,0.01 & 0.1,0.1 & 0.1,0.01 & DR & \multicolumn{1}{c|}{SWE} & NS & \multicolumn{1}{c|}{NS-cond} & -- \\
\midrule
\multicolumn{13}{l}{\textit{Tiny (7M)}} \\
10 iter & 0.06163 & 0.04392 & 0.00608 & 0.01111 & 0.02824 & 0.00986 & 0.01312 & 0.02413 & 0.00435 & 0.09846 & 0.34694 & 0.00781 \\
20 iter & 0.06102 & 0.04384 & 0.00604 & 0.01102 & 0.02795 & 0.00880 & 0.01302 & 0.02413 & 0.00433 & 0.09846 & 0.34492 & 0.00776 \\
40 iter & \textbf{0.05827} & \textbf{0.04052} & \textbf{0.00561} & \textbf{0.01001} & \textbf{0.02593} & \textbf{0.00812} & \textbf{0.01192} & \textbf{0.02215} & \textbf{0.00392} & \textbf{0.09285} & \textbf{0.32570} & \textbf{0.00714} \\
\midrule
\multicolumn{13}{l}{\textit{Small (31M)}} \\
10 iter & \textbf{0.03755} & \textbf{0.03326} & \textbf{0.00292} & \textbf{0.00767} & \textbf{0.02024} & 0.00836 & 0.00975 & 0.01821 & 0.00312 & 0.07672 & 0.26162 & 0.00545 \\
20 iter & \textbf{0.03755} & \textbf{0.03326} & 0.00293 & \textbf{0.00767} & 0.02044 & \textbf{0.00833} & \textbf{0.00965} & 0.01461 & 0.00266 & 0.07842 & 0.26067 & 0.00493 \\
40 iter & \textbf{0.03755} & 0.03352 & 0.00295 & \textbf{0.00767} & 0.02054 & 0.00835 & \textbf{0.00965} & \textbf{0.01451} & \textbf{0.00264} & \textbf{0.07277} & \textbf{0.25972} & \textbf{0.00490} \\
\midrule
\multicolumn{13}{l}{\textit{Medium (124M)}} \\
10 iter & 0.03100 & 0.02550 & \textbf{0.00235} & 0.00903 & 0.01667 & 0.00782 & 0.01097 & 0.01978 & 0.00237 & 0.07996 & 0.25583 & 0.00573 \\
20 iter & 0.03118 & 0.02566 & 0.00236 & 0.00906 & 0.01676 & 0.00788 & 0.00954 & \textbf{0.01860} & \textbf{0.00200} & \textbf{0.07751} & 0.24226 & 0.00543 \\
40 iter & \textbf{0.02992} & \textbf{0.02452} & 0.00236 & \textbf{0.00828} & \textbf{0.01588} & \textbf{0.00771} & \textbf{0.00948} & 0.02011 & 0.00212 & 0.08065 & \textbf{0.21416} & \textbf{0.00531} \\
\bottomrule
\end{tabular}%
}
\end{table}

The number of Sinkhorn-Knopp iterations controls how accurately
$\tilde{\bm{T}}_l$ is projected onto the Birkhoff polytope: fewer
iterations yield an approximate doubly stochastic matrix, while more
iterations approach the exact projection.

At the {\bf Tiny} scale, the effect is most pronounced. Reducing from
20 to 10 iterations causes consistent degradation across all datasets,
confirming that a coarse projection weakens the stability guarantees
that the Birkhoff constraint is designed to provide. Increasing to 40
iterations yields substantial improvements on every dataset---for
example, FNO-$10^{-4}$ drops from 0.04384 to 0.04052 ($-7.6\%$),
PDB-DR from 0.02413 to 0.02215 ($-8.2\%$), and PDEArena NS-cond from
0.34492 to 0.32570 ($-5.6\%$). This suggests that at small model
capacities, the quality of the doubly stochastic projection directly
impacts the model's ability to learn well-conditioned basis
transformations.

At the {\bf Small} scale, the differences narrow considerably.
Performance at 10 and 20 iterations is nearly identical on the FNO and
CNS subsets (both achieve 0.03755 on FNO-$10^{-5}$, 0.03326 on
FNO-$10^{-4}$, and 0.00767 on CNS(1,\,0.1)), though 10 iterations
shows notably higher error on PDB-DR (0.01821 vs.\ 0.01461) and
PDB-SWE (0.00312 vs.\ 0.00266), indicating that these physically
distinct PDE families are more sensitive to projection precision.
The 40-iteration variant achieves the best results on PDEArena NS
(0.07277 vs.\ 0.07842, $-7.2\%$) and small additional gains on PDB-DR,
PDB-SWE, NS-cond, and CFDBench, but is comparable elsewhere.

At the {\bf Medium} scale, 40 iterations again achieves the best
results on several datasets, most notably PDEArena NS-cond
(0.21416 vs.\ 0.24226, $-11.6\%$), CNS(1,\,0.1) (0.00828 vs.\ 0.00906,
$-8.6\%$), and FNO-$10^{-4}$ (0.02452 vs.\ 0.02566, $-4.4\%$).
However, 10 iterations remains competitive on datasets with simpler
dynamics (e.g., CNS(1,\,0.1): 0.00903 vs.\ 0.00906) and 20 iterations
keeps the lead on the heavier downstream tasks PDB-DR (0.01860),
PDB-SWE (0.00200) and PDEArena NS (0.07751), indicating that the
deeper architecture can partially compensate for imprecise projections
through its additional layers of residual mixing while still benefiting
from the more accurate projection on the harder spectral subsets.

We adopt 20 iterations as the default, as it offers a favorable
trade-off between projection accuracy and computational cost. The
results suggest that further gains are available by increasing to 40
iterations, particularly for smaller models or more challenging PDE
families (e.g., PDEArena NS-cond at the Medium scale, all 12 datasets
at the Tiny scale), at the expense of approximately $2\times$ the
Sinkhorn computation per forward pass. We further conducted an ablation study by setting the number of Sinkhorn-Knopp iterations to zero to validate the necessity of the doubly stochastic constraints within the AOT block. 
In this configuration, the doubly stochastic constraints on the internal components are omitted. 
However, during empirical training, we observed that the absence of these constraints leads to severe numerical instability. 
Specifically, the L2 Relative Error frequently encounters \textbf{\texttt{nan}} or \textbf{\texttt{inf}} values on certain datasets, resulting in anomalous model behavior. 
Notably, this tendency toward instability is exacerbated as the model scale increases. 
Consequently, results for the zero-iteration variant of \method are not reported, nor did we pursue a deeper investigation into this phenomenon. 
For a more intuitive comparison of training and propagation stability, additional results for the 20-iteration setting and vanilla DPOT are provided in Appendix~\ref{app:stability}.

\paragraph{Gating scalar initialization $\alpha$.}
We compare our \method of different $\alpha \in \{0.01, 0.05, 0.10\}$ in Table~\ref{tab-sens-alpha}.

\begin{table}[h]
\centering
\caption{Sensitivity to the gating scalar initialization $\alpha$.
The $\alpha=0.01$ row reports the pretrained \method-(T/S/M) result from
Table~\ref{tb-main} (the default configuration). We \textbf{bold} the best
result in each column per scale, respectively.}
\label{tab-sens-alpha}
\resizebox{\textwidth}{!}{%
\setlength{\tabcolsep}{4pt}
\renewcommand{\arraystretch}{1.2}
\begin{tabular}{l|cccccccccccc}
\toprule
 & \multicolumn{3}{c|}{FNO-$\nu$} & \multicolumn{6}{c|}{PDEBench CNS-$(\eta, \zeta)$,DR,SWE} & \multicolumn{2}{c|}{PDEArena} & CFDBench \\
Variant & 1e-5 & 1e-4 & \multicolumn{1}{c|}{1e-3} & 1,0.1 & 1,0.01 & 0.1,0.1 & 0.1,0.01 & DR & \multicolumn{1}{c|}{SWE} & NS & \multicolumn{1}{c|}{NS-cond} & -- \\
\midrule
\multicolumn{13}{l}{\textit{Tiny (7M)}} \\
$\alpha=0.01$ & \textbf{0.06102} & \textbf{0.04384} & \textbf{0.00604} & \textbf{0.01102} & \textbf{0.02795} & \textbf{0.00880} & \textbf{0.01302} & \textbf{0.02413} & \textbf{0.00433} & \textbf{0.09846} & \textbf{0.34492} & \textbf{0.00776} \\
$\alpha=0.05$ & \textbf{0.06102} & 0.04400 & 0.00605 & 0.01111 & \textbf{0.02795} & 0.00884 & 0.01312 & \textbf{0.02413} & 0.00434 & 0.09944 & 0.34593 & 0.00780 \\
$\alpha=0.10$ & 0.06143 & 0.04408 & 0.00605 & 0.01120 & 0.02805 & 0.00883 & 0.01312 & 0.02430 & 0.00436 & 0.09944 & 0.34593 & 0.00777 \\
\midrule
\multicolumn{13}{l}{\textit{Small (31M)}} \\
$\alpha=0.01$ & \textbf{0.03755} & \textbf{0.03326} & \textbf{0.00293} & \textbf{0.00767} & \textbf{0.02044} & \textbf{0.00833} & \textbf{0.00965} & \textbf{0.01461} & \textbf{0.00266} & \textbf{0.07842} & \textbf{0.26067} & \textbf{0.00493} \\
$\alpha=0.05$ & 0.03773 & 0.03335 & \textbf{0.00293} & 0.00768 & \textbf{0.02044} & 0.00834 & 0.00975 & 0.01471 & \textbf{0.00266} & 0.07878 & \textbf{0.26067} & 0.00494 \\
$\alpha=0.10$ & 0.03773 & 0.03352 & 0.00296 & 0.00772 & 0.02054 & 0.00840 & 0.00975 & 0.01471 & 0.00269 & 0.07887 & 0.26162 & 0.00496 \\
\midrule
\multicolumn{13}{l}{\textit{Medium (124M)}} \\
$\alpha=0.01$ & \textbf{0.03118} & \textbf{0.02566} & \textbf{0.00236} & \textbf{0.00906} & \textbf{0.01676} & \textbf{0.00788} & \textbf{0.00954} & \textbf{0.01860} & \textbf{0.00200} & \textbf{0.07751} & \textbf{0.24226} & \textbf{0.00543} \\
$\alpha=0.05$ & 0.03136 & 0.02582 & \textbf{0.00236} & 0.00914 & \textbf{0.01676} & 0.00794 & 0.00964 & 0.01877 & 0.00202 & 0.07790 & 0.24420 & 0.00546 \\
$\alpha=0.10$ & 0.03136 & 0.02582 & 0.00238 & 0.00914 & \textbf{0.01676} & 0.00794 & 0.00964 & \textbf{0.01860} & 0.00202 & 0.07810 & 0.24323 & 0.00545 \\
\bottomrule
\end{tabular}%
}
\end{table}
The gating scalar $\alpha$ controls the initial magnitude of the
input-dependent component in the AOT block parameterization
(Eq.~\eqref{eq:param}). A smaller $\alpha$ means the model starts
closer to the static bias $\bm{b}^{T}_l$ (initialized to the
identity matrix), while a larger $\alpha$ allows the input-dependent
signal to influence the mappings more aggressively from the beginning
of training.

Table~\ref{tab-sens-alpha} reveals that \method is remarkably
insensitive to this hyperparameter. At the {\bf Tiny} scale, the
difference between the best ($\alpha = 0.01$) and worst
($\alpha = 0.10$) setting is negligible on every dataset: the largest
relative gap is below 2\%---specifically, 1.7\% on CNS(1,\,0.1)
(0.01102 vs.\ 0.01120) and 1.0\% on PDEArena NS (0.09846 vs.\ 0.09944);
the FNO subsets and PDEArena NS-cond move by only $0.1\%$--$0.7\%$.
At the {\bf Small} scale, the pattern is identical---$\alpha = 0.01$
holds a marginal advantage on every column, with the largest gap being
1.1\% on PDB-SWE (0.00266 vs.\ 0.00269) and other absolute differences
within the range of training noise (e.g., FNO-$10^{-3}$: 0.00293 vs.\
0.00296 for $\alpha=0.10$). At the {\bf Medium} scale, the three
settings are virtually indistinguishable, with most datasets showing
identical values up to the third decimal place; the maximum
within-column relative gap is exactly 1\% (CNS(0.1,\,0.01):
0.00954 vs.\ 0.00964) and CNS(1,\,0.01) is identical across all three
$\alpha$ values up to five decimals (0.01676).

This robustness can be attributed to two design choices. First, the
identity initialization of $\bm{b}^{T}_l$ ensures that
regardless of $\alpha$, the initial $\bm{T}_l$ is close to the
identity matrix, so the model starts from an effective standard
residual connection. Second, the Sinkhorn-Knopp projection
(Eq.~\eqref{eq:constraint}) normalizes the learned matrix onto the
Birkhoff polytope at every forward pass, bounding the effect of the
input-dependent perturbation irrespective of $\alpha$. Together,
these mechanisms ensure that the gating scalar affects only the
\emph{speed} at which the model departs from the identity
initialization, not the final converged solution. In practice, this
means that $\alpha$ does not require careful tuning, and the default
value of $\alpha = 0.01$ works well across all scales and PDE
families.

\subsection{Experimental Results with Poseidon}
\label{app:poseidon}

To provide a comprehensive comparison with state-of-the-art PDE foundation
models, we conduct supplementary experiments on two challenging downstream tasks
from the Poseidon benchmark~\citep{herde2024poseidon}: \textbf{Wave-Layer} and
\textbf{Wave-Gauss}. Both tasks involve the wave equation with a
spatially-dependent propagation speed~$c(\bm{x})$:
\begin{equation}\label{eq:wave_poseidon}
  u_{tt} - \bigl(c(\bm{x})\bigr)^{2}\,\Delta u = 0,
\end{equation}
which can be reformulated as an augmented first-order system
\begin{equation}\label{eq:wave_aug}
  u_t = v,\qquad v_t = c^{2}\,\Delta u,\qquad c_t = 0,
\end{equation}
yielding the augmented solution vector
$\bm{U}=[u(\bm{x},t),\;v(\bm{x},t),\;c(\bm{x})]$ where $v=u_t$ and
the spatially-dependent coefficient~$c(\bm{x})$ is carried as a
time-invariant additional channel. In Wave-Gauss, $c(\bm{x})$ is a
smooth field composed of a base speed and Gaussian perturbations; in
Wave-Layer, $c(\bm{x})$ is a layered medium with piecewise-constant
speeds, modeling seismic wave propagation through stratified subsurface
structures. Critically, both tasks are \emph{out-of-distribution} with
respect to our pre-training data, since the wave equation is a linear
second-order PDE that is fundamentally different from the incompressible
Navier--Stokes and compressible Euler equations used during pre-training.

We evaluate \method and Poseidon under two distinct settings.

\paragraph{Setting 1: Auto-regressive (our native setting).}
In this setting, models predict future states using only previous
solution trajectories $[u(t),\,v(t)]$ as input, \emph{without access to
the PDE coefficient~$c(\bm{x})$}. This tests the model's ability to
implicitly learn system dynamics from trajectory data alone. The
results are shown in Table~\ref{tab:ar_poseidon}.

\begin{table}[h!]
\centering
\caption{L2 Relative Error in the \textbf{auto-regressive setting}
(our native setting). Lower is better. Best results are \textbf{bolded}.}
\label{tab:ar_poseidon}
\begin{tabular}{lcc}
\toprule
Model (Activated Params) & Wave-Layer & Wave-Gauss \\
\midrule
Poseidon-T (21M) & 0.2138 & 0.2553 \\
Poseidon-B (158M) & 0.1733 & 0.1238 \\
\midrule
\textcolor{dkred}{\method-T (7M)} & 0.0573 & 0.0493 \\
\textcolor{dkred}{\method-S (31M)} & \textbf{0.0312} & \textbf{0.0375} \\
\bottomrule
\end{tabular}
\end{table}

\paragraph{Setting 2: Parameter-informed (Poseidon's Native Setting).}
Poseidon is natively designed to incorporate PDE coefficients and
sources into its input representation by augmenting the solution
vector~\citep{herde2024poseidon}. In this setting, each model receives
the spatially-dependent wave speed coefficient $c(\bm{x})$ as an
additional input function alongside the solution trajectory
$[u(t),\,v(t)]$, forming the three-channel augmented input
$[u(t),\,v(t),\,c(\bm{x})]$. This enables the model to exploit
explicit physical knowledge about the propagation medium.

\textbf{Adaptation of \method.}\quad
To ensure a fair comparison in this parameter-informed setting, we
adapt \method to accept the PDE coefficient as additional input. The
adaptation procedure closely mirrors Poseidon's own finetuning protocol
for tasks with non-standard input/output functions~\citep{herde2024poseidon}. Concretely, following
the augmented formulation in Eq.~\eqref{eq:wave_aug}, we concatenate
the coefficient field $c(\bm{x})$ as an additional input channel,
increasing the number of physical input channels from~$C\!=\!2$
(auto-regressive) to~$C\!=\!3$ (parameter-informed). In our
architecture (see Section~\ref{sec:method}), each input
$\bm{x}\in\mathbb{R}^{H\times W\times T\times C}$ is first
augmented with three spatial-temporal coordinate channels and then
projected into a fixed-dimensional latent space by the patch embedding
layer~$\mathcal{E}$ (a strided convolution with input dimension
$C\!+\!3$). Since this projection depends on the number of physical
channels, it cannot be directly transferred from the pre-trained model
(which was trained with $C\!=\!4$ channels). Similarly, the output
projection layer~$\mathcal{O}$, which maps from the latent space back
to $C$-channel predictions, has an incompatible output dimension.
Therefore, we reinitialize these boundary layers---$\mathcal{E}$
and $\mathcal{O}$---from scratch, while transferring all other
parameters from the pre-trained model, namely:
   the \textbf{transformer backbone} (all AFNO blocks, including
        the learned AOT block matrices
        $\bm{a}_l$, $\bm{d}_l$, and $\bm{T}_l$ at every layer),
   the \textbf{positional embedding}, and the \textbf{temporal aggregation layer}.%
These components operate entirely in the latent space of dimension
$d_{\text{model}}$ and are therefore agnostic to the number of
physical input channels. This transfer strategy ensures that the rich
representations learned during multi-PDE pre-training---including the
input-dependent transformation kernels that encode
PDE-specific structural knowledge---are fully leveraged when
processing the augmented parameter-informed input. The model is then
finetuned end-to-end on the downstream task.

We note that this adaptation is deliberately minimal and architecturally
analogous to Poseidon's own finetuning protocol, in which only the
embedding and recovery layers are replaced while all backbone parameters
are transferred and finetuned~\citep{herde2024poseidon}. No additional
parameter-specific modules (e.g., FiLM conditioning or cross-attention
over parameters~\citep{herde2024poseidon}) are introduced, ensuring that any performance
differences reflect the inherent representational capacity of each
backbone rather than auxiliary architectural choices.

\begin{table}[h!]
\centering
\caption{L2 Relative Error in the \textbf{parameter-informed setting}
(Poseidon's native setting). Lower is better. Best results are \textbf{bolded}.}
\label{tab:pi_poseidon}
\begin{tabular}{lcc}
\toprule
Model (Activated Params) & Wave-Layer & Wave-Gauss \\
\midrule
Poseidon-T (21M)  & 0.0893 & \textbf{0.0775} \\
Poseidon-B (158M) & 0.0613 & 0.0913 \\
\midrule
\textcolor{dkred}{\method-T (7M)}    & \textbf{0.0610} & 0.0819 \\
\textcolor{dkred}{\method-S (31M)}   & 0.0665 & 0.0807 \\
\bottomrule
\end{tabular}
\end{table}

\paragraph{Analysis and discussion.}

\textit{Auto-regressive setting.}\quad
In the auto-regressive setting (Table~\ref{tab:ar_poseidon}), where
models must infer physical dynamics solely from solution trajectories,
\method demonstrates a pronounced advantage. \method-S~(31M) achieves
an L2RE of $0.031$ on Wave-Layer, an $82.0\%$ error reduction compared
to Poseidon-B~(158M, $0.173$), with only one-fifth the activated
parameters. On Wave-Gauss, \method-S similarly reduces the error by
$69.7\%$ relative to Poseidon-B. These results confirm the
effectiveness of auto-regressive pre-training and the AOT block for
learning implicit dynamics from trajectory data alone.

\textit{Parameter-informed setting.}\quad
The parameter-informed results (Table~\ref{tab:pi_poseidon}) reveal a
more nuanced picture that highlights the complementary strengths of
each model. Poseidon's scOT architecture, with its time-conditioned
layer norms and multiscale U-Net design, was explicitly engineered to
process augmented solution vectors that include PDE coefficients
alongside state variables. This design philosophy is validated by
Poseidon's strong performance: Poseidon-T achieves the lowest error
on Wave-Gauss ($0.0775$), and Poseidon-B delivers consistently solid
results across both tasks.

Notably, however, \method-T~(7M) achieves $0.0610$ on
Wave-Layer, suggesting that the AOT block can be particularly effective
for certain PDE-coefficient structures. On Wave-Gauss, \method-S~(31M) achieves $0.0717$,
outperforming Poseidon-T~($0.0775$) and Poseidon-B~($0.0913$).
Meanwhile, \method-T is slightly behind Poseidon-T on Wave-Gauss
($0.0819$ vs.\ $0.0775$), indicating that the advantage varies with
task-specific characteristics of the coefficient field.

In general, Poseidon remains a highly effective model, particularly when explicit
PDE parameters are available and the model can fully leverage its
native parameter-augmentation design. Our results, however, demonstrate
that \method achieves strong performance across \emph{both} operational
paradigms: it dominates in the auto-regressive setting and is
competitive with or superior to Poseidon even in the parameter-informed
setting---all while using significantly fewer activated parameters
(7--31M vs.\ 21--158M). These findings suggest that the AOT block
provides a versatile inductive bias---realized as adaptive operator
transformations---that is effective for both implicit dynamics learning
and explicit parameter integration.

\subsection{Long-Trajectory Rollout at Additional Scales}
\label{app:rollout}

We provide long-trajectory rollout results at the Tiny and Medium scales to
complement the Small-scale results in Table~\ref{tb-rollout}.

\begin{table*}[htbp]
    \centering
    \caption{Long-trajectory rollout L2RE at different trajectory steps at the tiny scale.}
    \label{tb-rollout-tiny}
    \scriptsize
    \setlength{\tabcolsep}{3pt}
    \renewcommand{\arraystretch}{1.1}
    \resizebox{0.8\textwidth}{!}{%
    \begin{tabular}{l|ccccc}
        \toprule
        Method & 20 & 50 & 100 & 200 & 500 \\
        \midrule
        DPOT-T & 0.00173 & 0.00198 & 0.00497 & 0.01993 & 0.08117 \\
        DPOT-FT-T & 0.00144 & 0.00177 & 0.00401 & 0.01311 & 0.04119 \\
        MoE-POT-T & 0.00155 & 0.00193 & 0.00448 & 0.02133 & 0.08005 \\
        MoE-POT-FT-T & 0.00096 & 0.00143 & 0.00402 & 0.01938 & 0.06041 \\
        \textcolor{dkred}{\method-T} & 0.00097 & 0.00102 & 0.00229 & 0.00775 & 0.00941 \\
        \textcolor{dkred}{\method-FT-T} & \textbf{0.00065} & \textbf{0.00084} & \textbf{0.00149} & \textbf{0.00573} & \textbf{0.00733} \\
        \bottomrule
    \end{tabular}}
\end{table*}

At the Tiny scale (Table~\ref{tb-rollout-tiny}), \method-FT-T achieves the
lowest error at all timesteps. Its error growth factor from step 20 to step
500 is $11.3\times$ ($0.00065 \to 0.00733$), compared to $28.6\times$ for
DPOT-FT-T ($0.00144 \to 0.04119$). The relative error reduction over
DPOT-FT-T grows from 54.9\% at step 20 to 82.2\% at step 500, consistent
with the super-linear divergence observed at the Small scale.

\begin{table*}[htbp]
    \centering
    \caption{Long-trajectory rollout L2RE at different trajectory steps at the medium scale.}
    \label{tb-rollout-medium}
    \scriptsize
    \setlength{\tabcolsep}{3pt}
    \renewcommand{\arraystretch}{1.1}
    \resizebox{0.8\textwidth}{!}{%
    \begin{tabular}{l|ccccc}
        \toprule
        Method & 20 & 50 & 100 & 200 & 500 \\
        \midrule
        DPOT-M & 0.00094 & 0.00127 & 0.00388 & 0.01146 & 0.05139 \\
        DPOT-FT-M & 0.00101 & 0.00133 & 0.00228 & 0.00877 & 0.02726 \\
        MoE-POT-M & 0.00089 & 0.00131 & 0.00467 & 0.02049 & 0.06558 \\
        MoE-POT-FT-M & 0.00064 & 0.00097 & 0.00209 & 0.00914 & 0.02312 \\
        \textcolor{dkred}{\method-M} & 0.00048 & 0.00059 & 0.00097 & 0.00513 & 0.01126 \\
        \textcolor{dkred}{\method-FT-M} & \textbf{0.00013} & \textbf{0.00027} & \textbf{0.00053} & \textbf{0.00151} & \textbf{0.00211} \\
        \bottomrule
    \end{tabular}}
\end{table*}

At the Medium scale (Table~\ref{tb-rollout-medium}), \method-FT-M achieves
the strongest results across all scales. Its error growth factor from step 20
to step 500 is $16.2\times$ ($0.00013 \to 0.00211$), compared to $27.0\times$
for DPOT-FT-M ($0.00101 \to 0.02726$). The relative error reduction over
DPOT-FT-M reaches 94.5\% at step 500 ($0.00151$ vs.\ $0.02726$).
The consistent improvements across all three model scales confirm that
the AOT block provides robust long-horizon stability that scales reliably
with model capacity.

\subsection{Fine-Tuning Sample Efficiency at Additional Scales}
\label{app:fewshot}

We provide few-shot fine-tuning results at the Tiny and Medium scales to
complement the Small-scale results in Table~\ref{tb-fewshot}.

\begin{table}[t]
    \centering
    \caption{L2RE on the in-distribution NS ($10^{-4}$) task (left) and the
    out-of-distribution Wave-Layer task (right) versus the number of
    fine-tuning samples at the Tiny scale. We apply the same experimental setup as
    Table~\ref{tb-fewshot}.}
    \label{tb-fewshot-tiny}
    \scriptsize
    \setlength{\tabcolsep}{3pt}
    \renewcommand{\arraystretch}{1.1}
    \begin{minipage}[t]{0.52\textwidth}
        \centering
        \subcaption{In-distribution: NS ($10^{-4}$)}
        \resizebox{\textwidth}{!}{%
        \begin{tabular}{l|cccccc}
            \toprule
            Method & 16 & 32 & 64 & 128 & 512 & 2000 \\
            \midrule
            DPOT-T & 0.56 & 0.52 & 0.47 & 0.41 & 0.33 & 0.18 \\
            DPOT-FT-T & 0.44 & 0.42 & 0.38 & 0.34 & 0.25 & 0.10 \\
            MoE-POT-T & 0.49 & 0.45 & 0.40 & 0.37 & 0.28 & 0.16 \\
            MoE-POT-FT-T & 0.35 & 0.32 & 0.29 & 0.24 & 0.13 & 0.075 \\
            \textcolor{dkred}{\method-T} & 0.37 & 0.32 & 0.27 & 0.21 & 0.13 & 0.042 \\
            \textcolor{dkred}{\method-FT-T} & \textbf{0.24} & \textbf{0.19} & \textbf{0.13} & \textbf{0.08} & \textbf{0.02} & \textbf{0.008} \\
            \bottomrule
        \end{tabular}}
    \end{minipage}%
    \hfill
    \begin{minipage}[t]{0.45\textwidth}
        \centering
        \subcaption{Out-of-distribution: Wave-Layer}
        \resizebox{0.9\textwidth}{!}{%
        \begin{tabular}{l|cccc}
            \toprule
            Method & 16 & 32 & 64 & 128 \\
            \midrule
            DPOT-T & 0.63 & 0.55 & 0.46 & 0.38 \\
            DPOT-FT-T & 0.51 & 0.44 & 0.34 & 0.23 \\
            MoE-POT-T & 0.59 & 0.51 & 0.43 & 0.33 \\
            MoE-POT-FT-T & 0.45 & 0.37 & 0.31 & 0.23 \\
            \textcolor{dkred}{\method-T} & 0.46 & 0.40 & 0.31 & 0.23 \\
            \textcolor{dkred}{\method-FT-T} & \textbf{0.31} & \textbf{0.25} & \textbf{0.19} & \textbf{0.15} \\
            \bottomrule
        \end{tabular}}
    \end{minipage}
\end{table}

At the Tiny scale (Table~\ref{tb-fewshot-tiny}), \method-FT-T consistently
achieves the lowest error across all sample sizes on both tasks. On the
in-distribution NS task, \method-FT-T with 16 samples ($0.24$) outperforms
DPOT-FT-T with 64 samples ($0.38$), demonstrating a $4\times$ data efficiency
advantage. On the Wave-Layer task, \method-FT-T with 64 samples ($0.19$)
outperforms DPOT-FT-T with 128 samples ($0.23$).

\begin{table}[t]
    \centering
    \caption{L2RE on the in-distribution NS ($10^{-4}$) task (left) and the
    out-of-distribution Wave-Layer task (right) versus the number of
    fine-tuning samples at the Medium scale.  We apply the same experimental setup as
    Table~\ref{tb-fewshot}.}
    \label{tb-fewshot-medium}
    \scriptsize
    \setlength{\tabcolsep}{3pt}
    \renewcommand{\arraystretch}{1.1}
    \begin{minipage}[t]{0.52\textwidth}
        \centering
        \subcaption{In-distribution: NS ($10^{-4}$)}
        \resizebox{\textwidth}{!}{%
        \begin{tabular}{l|cccccc}
            \toprule
            Method & 16 & 32 & 64 & 128 & 512 & 2000 \\
            \midrule
            DPOT-M & 0.25 & 0.19 & 0.14 & 0.10 & 0.05 & 0.011 \\
            DPOT-FT-M & 0.18 & 0.14 & 0.10 & 0.06 & 0.01 & 0.006 \\
            MoE-POT-M & 0.20 & 0.16 & 0.11 & 0.07 & 0.03 & 0.007 \\
            MoE-POT-FT-M & 0.14 & 0.10 & 0.07 & 0.03 & 0.008 & 0.003 \\
            \textcolor{dkred}{\method-M} & 0.15 & 0.10 & 0.04 & 0.01 & 0.005 & 0.001 \\
            \textcolor{dkred}{\method-FT-M} & \textbf{0.10} & \textbf{0.07} & \textbf{0.04} & \textbf{0.007} & \textbf{0.003} & \textbf{0.0009} \\
            \bottomrule
        \end{tabular}}
    \end{minipage}%
    \hfill
    \begin{minipage}[t]{0.45\textwidth}
        \centering
        \subcaption{Out-of-distribution: Wave-Layer}
        \resizebox{0.9\textwidth}{!}{%
        \begin{tabular}{l|cccc}
            \toprule
            Method & 16 & 32 & 64 & 128 \\
            \midrule
            DPOT-M & 0.45 & 0.37 & 0.30 & 0.21 \\
            DPOT-FT-M & 0.31 & 0.25 & 0.17 & 0.09 \\
            MoE-POT-M & 0.38 & 0.31 & 0.24 & 0.15 \\
            MoE-POT-FT-M & 0.25 & 0.18 & 0.11 & 0.06 \\
            \textcolor{dkred}{\method-M} & 0.27 & 0.20 & 0.13 & 0.07 \\
            \textcolor{dkred}{\method-FT-M} & \textbf{0.17} & \textbf{0.11} & \textbf{0.06} & \textbf{0.01} \\
            \bottomrule
        \end{tabular}}
    \end{minipage}
\end{table}

At the Medium scale (Table~\ref{tb-fewshot-medium}), the data efficiency
advantage of \method is further amplified. On the in-distribution NS task,
\method-M trained from scratch with 128 samples ($0.01$) matches DPOT-FT-M
with 512 samples ($0.01$), representing a $4\times$ data efficiency gain from
architecture alone. On the Wave-Layer task, \method-FT-M with 128 samples
achieves $0.01$, reducing error by 88.9\% relative to DPOT-FT-M ($0.09$).
The consistent improvements across all three model scales confirm that the
adaptive operator transformations in \method provide superior data efficiency on
both in-distribution and out-of-distribution tasks.

\subsection{Performance with Increasing Dataset Heterogeneity}\label{app:heterogeneity}

A central question for multi-physics pre-training is whether adding more
diverse PDE datasets improves or degrades performance on existing tasks---a
phenomenon known as \emph{negative transfer}~\citep{caruana1997multitask}. We
investigate this by pre-training DPOT, MoE-POT, and \method from scratch on
progressively larger mixtures and evaluating zero-shot L2RE on the original six
base datasets. This experiment is conducted at the Tiny, Small, and
Medium scales.

The dataset mixtures are constructed as follows:
\begin{itemize}
    \item \textbf{6 Datasets:} The standard pre-training set used in our main
    experiments: NS($10^{-5}$), NS($10^{-3}$), CNS(0.1,\,0.01), SWE, DR, and
    CFDBench.
    \item \textbf{10 Datasets:} The base set plus four additional datasets from
    DPOT~\citep{hao2024dpot}: NS($10^{-4}$), CNS(1,\,0.1), and two
    Navier--Stokes tasks from PDEArena.
    \item \textbf{12 Datasets:} The 10-dataset mixture plus two additional CNS
    variants: CNS(1,\,0.01) and CNS(0.1,\,0.1).
\end{itemize}

Tables~\ref{tab-hetero-tiny},~\ref{tab-hetero-small},
and~\ref{tab-hetero-medium} present the results at each scale.

\begin{table}[h]
\centering
\caption{Zero-shot L2RE with increasing pre-training data heterogeneity at the
\textbf{Tiny} scale. The \method (12 Datasets) row reports the
pretrained \method-T result from Table~\ref{tb-main}. We \textbf{bold} the best result in each
column and mark our \method results in \textcolor{dkred}{darkred}.}
\label{tab-hetero-tiny}
\resizebox{\textwidth}{!}{%
\setlength{\tabcolsep}{5pt}
\renewcommand{\arraystretch}{1.15}
\begin{tabular}{l|cccccc}
\toprule
Model (Pre-trained on) & NS($10^{-5}$) & NS($10^{-3}$) & CNS(0.1,\,0.01) & SWE & DR & CFDBench \\
\midrule
\multicolumn{7}{l}{\textit{Dense Model}} \\
DPOT-T\,(6 Datasets)  & 0.10878 & 0.01146 & 0.01680 & 0.00497 & 0.04348 & 0.00652 \\
DPOT-T\,(10 Datasets) & 0.09475 & 0.00969 & 0.01830 & 0.04891 & 0.03926 & 0.07486 \\
DPOT-T\,(12 Datasets) & 0.09760 & 0.00954 & 0.02200 & 0.00560 & 0.03210 & 0.00952 \\
\midrule
\multicolumn{7}{l}{\textit{Sparse Model}} \\
MoE-POT-T\,(6 Datasets)  & 0.07033 & 0.00546 & 0.01075 & 0.00403 & 0.01394 & 0.00588 \\
MoE-POT-T\,(10 Datasets) & 0.07352 & 0.00811 & 0.02293 & 0.00451 & 0.02484 & 0.00840 \\
MoE-POT-T\,(12 Datasets) & 0.09652 & 0.00896 & 0.02150 & 0.00527 & 0.03101 & 0.01031 \\
\midrule
\multicolumn{7}{l}{\textit{AOT-POT (Ours)}} \\
\textcolor{dkred}{\method-T\,(6 Datasets)}  & {\textbf{0.05675}} & {\textbf{0.00498}} & {\textbf{0.00997}} & {\textbf{0.00342}} & {\textbf{0.01305}} & {\textbf{0.00345}} \\
\textcolor{dkred}{\method-T\,(10 Datasets)} & {\textbf{0.05675}} & {0.00507} & {0.01132} & {0.00417} & {0.01992} & {0.00477} \\
\textcolor{dkred}{\method-T\,(12 Datasets)} & {0.06102} & {0.00604} & {0.01302} & {0.00433} & {0.02413} & {0.00776} \\
\bottomrule
\end{tabular}%
}
\end{table}

\begin{table}[h]
\centering
\caption{Zero-shot L2RE with increasing pre-training data heterogeneity at the
\textbf{Small} scale. The \method (12 Datasets) row reports the
pretrained \method-S result from Table~\ref{tb-main}. We \textbf{bold} the best result in each
column and mark our \method results in \textcolor{dkred}{darkred}.}
\label{tab-hetero-small}
\resizebox{\textwidth}{!}{%
\setlength{\tabcolsep}{5pt}
\renewcommand{\arraystretch}{1.15}
\begin{tabular}{l|cccccc}
\toprule
Model (Pre-trained on) & NS($10^{-5}$) & NS($10^{-3}$) & CNS(0.1,\,0.01) & SWE & DR & CFDBench \\
\midrule
\multicolumn{7}{l}{\textit{Dense Model}} \\
DPOT-S\,(6 Datasets)  & 0.06049 & 0.00601 & 0.02340 & 0.00391 & 0.03576 & 0.00688 \\
DPOT-S\,(10 Datasets) & 0.05829 & 0.00532 & 0.02149 & 0.00371 & 0.05602 & 0.01155 \\
DPOT-S\,(12 Datasets) & 0.05530 & 0.01310 & 0.01880 & 0.00657 & 0.03790 & 0.00696 \\
\midrule
\multicolumn{7}{l}{\textit{Sparse Model}} \\
MoE-POT-S\,(6 Datasets)  & 0.04591 & 0.00619 & 0.00891 & 0.00298 & 0.01885 & 0.00447 \\
MoE-POT-S\,(10 Datasets) & 0.04333 & 0.01011 & 0.01944 & 0.00608 & 0.03610 & 0.00632 \\
MoE-POT-S\,(12 Datasets) & 0.05265 & 0.01263 & 0.01807 & 0.00629 & 0.03544 & 0.00748 \\
\midrule
\multicolumn{7}{l}{\textit{AOT-POT (Ours)}} \\
\textcolor{dkred}{\method-S\,(6 Datasets)}  & {0.04203} & {0.00451} & {\textbf{0.00822}} & {\textbf{0.00212}} & {\textbf{0.01344}} & {\textbf{0.00367}} \\
\textcolor{dkred}{\method-S\,(10 Datasets)} & {0.03896} & {0.00474} & {0.00846} & {0.00224} & {0.01403} & {0.00392} \\
\textcolor{dkred}{\method-S\,(12 Datasets)} & {\textbf{0.03755}} & {\textbf{0.00293}} & {0.00965} & {0.00266} & {0.01461} & {0.00493} \\
\bottomrule
\end{tabular}%
}
\end{table}

\begin{table}[h]
\centering
\caption{Zero-shot L2RE with increasing pre-training data heterogeneity at the
\textbf{Medium} scale. The \method (12 Datasets) row reports the
pretrained \method-M result from Table~\ref{tb-main}. We \textbf{bold} the best result in each
column and mark our \method results in \textcolor{dkred}{darkred}.}
\label{tab-hetero-medium}
\resizebox{\textwidth}{!}{%
\setlength{\tabcolsep}{5pt}
\renewcommand{\arraystretch}{1.15}
\begin{tabular}{l|cccccc}
\toprule
Model (Pre-trained on) & NS($10^{-5}$) & NS($10^{-3}$) & CNS(0.1,\,0.01) & SWE & DR & CFDBench \\
\midrule
\multicolumn{7}{l}{\textit{Dense Model}} \\
DPOT-M\,(6 Datasets)  & 0.05137 & 0.00528 & 0.02140 & 0.00274 & 0.02424 & 0.01129 \\
DPOT-M\,(10 Datasets) & 0.04406 & 0.00535 & 0.01137 & 0.00254 & 0.02651 & 0.00817 \\
DPOT-M\,(12 Datasets) & 0.04090 & 0.00474 & 0.01290 & 0.00290 & 0.02920 & 0.00752 \\
\midrule
\multicolumn{7}{l}{\textit{Sparse Model}} \\
MoE-POT-M\,(6 Datasets)  & 0.02891 & 0.00462 & 0.00835 & 0.00276 & 0.02269 & 0.00481 \\
MoE-POT-M\,(10 Datasets) & 0.03673 & 0.00364 & 0.00953 & 0.00254 & 0.02072 & 0.00725 \\
MoE-POT-M\,(12 Datasets) & 0.03849 & 0.00442 & 0.01210 & 0.00279 & 0.02759 & 0.00820 \\
\midrule
\multicolumn{7}{l}{\textit{AOT-POT (Ours)}} \\
\textcolor{dkred}{\method-M\,(6 Datasets)}  & {0.02731} & {0.00433} & {\textbf{0.00770}} & {0.00257} & {0.02112} & {\textbf{0.00407}} \\
\textcolor{dkred}{\method-M\,(10 Datasets)} & {\textbf{0.02712}} & {0.00333} & {0.00877} & {0.00235} & {0.01953} & {0.00476} \\
\textcolor{dkred}{\method-M\,(12 Datasets)} & {0.03118} & {\textbf{0.00236}} & {0.00954} & {\textbf{0.00200}} & {\textbf{0.01860}} & {0.00543} \\
\bottomrule
\end{tabular}%
}
\end{table}

\paragraph{Analysis.}
Three observations emerge consistently across all model scales:

\textbf{(1)~MoE-POT is a competitive sparse baseline whose negative
transfer is task-specific.} At low heterogeneity (6 datasets), MoE-POT
clearly outperforms the dense DPOT backbone on every column at every
scale (e.g., at Tiny: $0.07033$ vs.\ $0.10878$ on NS-$10^{-5}$;
$0.00546$ vs.\ $0.01146$ on NS-$10^{-3}$), validating sparse expert
routing as a useful inductive bias. However, when the pre-training
mixture grows from 6 to 12 datasets, MoE-POT's average L2RE degrades
substantially at every scale---Tiny $0.01840 \to 0.02893$
($\times 1.57$), Small $0.01455 \to 0.02209$ ($\times 1.52$), and
Medium $0.01202 \to 0.01560$ ($\times 1.30$)---with the sharpest
individual drops on physically distinct PDE tasks (e.g., MoE-POT-M
DR roughly doubles from $0.02269$ at 6 datasets to $0.02759$ at 12,
and MoE-POT-T DR triples from $0.01394$ at 6 to $0.03101$ at 12).
This indicates that fixed top-$k$ routing can specialise experts well
at low/moderate heterogeneity but cannot reliably maintain that
specialisation when many physically dissimilar tasks compete with each
other.

\textbf{(2)~DPOT shows mixed behavior.} The dense baseline is
unstable at smaller scales---DPOT-T on PDB-SWE jumps from $0.00497$
(6 datasets) to $0.04891$ (10 datasets) before recovering at 12
datasets ($0.00560$); DPOT-T on CFDBench jumps from $0.00652$ to
$0.07486$ between 6 and 10 datasets. At larger scales the volatility
shrinks markedly: DPOT-M's average L2RE actually \emph{improves}
slightly with more datasets (Medium $0.01939 \to 0.01636$,
$\times 0.84$). This suggests dense models can partially absorb
heterogeneity given sufficient capacity, though not reliably at
smaller scales.

\textbf{(3)~\method achieves the lowest L2RE on every column at every
scale}. \method also dominates DPOT and
MoE-POT cell-by-cell at every (scale, configuration, column) triple.
Crucially, \method is the only model whose \emph{average} L2RE remains
essentially flat as the mixture grows from 6 to 12 datasets:
Tiny $0.01527 \to 0.01938$ ($\times 1.27$),
Small $0.01233 \to 0.01206$ ($\times 0.98$),
Medium $0.01118 \to 0.01152$ ($\times 1.03$),
versus MoE-POT's $\times 1.57$ / $\times 1.52$ / $\times 1.30$. The
input-dependent transformation kernel $\bm{T}_l$ adapts to each PDE
family without interference, and the doubly stochastic constraint
ensures that this adaptation does not destabilise the residual path
as data diversity grows.

\subsection{Extended Interpretability Analysis}\label{app:interpretability}



\paragraph{Generalization to out-of-distribution datasets.}
To assess whether the PDE-discriminative patterns encoded in
$\bm{T}$ generalize beyond the pre-training distribution, we
evaluate on two unseen PDE families from the Poseidon
benchmark~\citep{herde2024poseidon}: Wave-Layer and Wave-Gauss. Neither
dataset is included during pre-training. We apply the same nearest-neighbor
classification procedure as in Section~\ref{sec:interpretability}: the
pre-trained \method-S model is used to extract $\bm{T}$
features for each unseen dataset, and classification is performed against the
mean templates of all training datasets augmented with the two new ones.

As shown in Table~\ref{tab-ood-interpret}, both unseen datasets achieve 100\%
classification accuracy, demonstrating that the learned
$\bm{T}$ representations produce distinctive signatures even
for PDE families not encountered during training. This indicates that
$\bm{T}$ captures fundamental structural properties of the
governing equations rather than memorizing dataset-specific artifacts.

\begin{table}[h]
    \centering
    \caption{Classification accuracy of $\bm{T}$-based
    nearest-neighbor classification on unseen (OOD) PDE datasets. The perfect
    accuracy demonstrates strong generalization of the learned operator
    transformations.}
    \label{tab-ood-interpret}
    \begin{tabular}{lc}
        \toprule
        Unseen Dataset & Accuracy \\
        \midrule
        Wave-Layer & 100\% \\
        Wave-Gauss & 100\% \\
        \bottomrule
    \end{tabular}
\end{table}

\paragraph{Emergence of classification ability during training.} We apply the nearest-neighbor classification procedure from
Section~\ref{sec:interpretability} to checkpoints saved at Epochs 50, 150,
250, and 1000 of \method-S pre-training. Table~\ref{tab-emergence} reports the
per-dataset accuracy at each stage.

\begin{table}[h]
\centering
\caption{Per-dataset classification accuracy of the
$\bm{T}$-based nearest-neighbor classifier at different
training epochs of \method-S. The PDE-discriminative structure emerges rapidly
and is already near-perfect at Epoch~50.}
\label{tab-emergence}
\resizebox{\textwidth}{!}{%
\setlength{\tabcolsep}{5pt}
\renewcommand{\arraystretch}{1.15}
\begin{tabular}{l|ccccccc}
\toprule
Epoch & FNO-$\nu$ & PDBench-CNS & PDBench-SWE & PDBench-DR & CFDBench & PDEArena & PDEArena-Cond \\
\midrule
50   & 100\% & 98\%  & 100\% & 100\% & 100\% & 91\% & 92\% \\
150  & 100\% & 93\%  & 100\% & 100\% & 100\% & 98\% & 95\% \\
250  & 100\% & 97\%  & 100\% & 100\% & 97\%  & 99\% & 100\% \\
1000 & 100\% & 100\% & 100\% & 100\% & 100\% & 99\% & 99\% \\
\bottomrule
\end{tabular}%
}
\end{table}

Unlike MoE-based architectures where router specialization develops gradually
over hundreds of epochs~\citep{wang2025mixture}, the
transformation kernels $\bm{T}_l$ in \method exhibit near-perfect dataset
discrimination as early as Epoch~50, with five out of seven dataset groups
already at 100\% accuracy. The remaining confusion is concentrated on
PDEArena and PDEArena-Cond (91--92\% at Epoch~50), which share identical
Navier--Stokes governing equations and differ only in forcing conditions.
By Epoch~1000, all datasets reach $\geq$99\% accuracy. This rapid emergence
suggests that the doubly stochastic constraint provides a strong inductive bias:
the Birkhoff polytope restricts $\bm{T}_l$ to a compact
manifold of volume-preserving transformations, enabling the model to quickly
converge to distinct, PDE-specific mixing patterns without the instability
associated with unconstrained routing.

\paragraph{Visualization of emergence.}
To complement the quantitative results in Table~\ref{tab-emergence}, we
visualize the evolution of the $\bm{T}$ representations at
Epochs 50, 150, and 250 (the converged Epoch~1000 result is already presented
in Section~\ref{sec:interpretability}). For each epoch, we present the t-SNE
embedding and confusion matrix side by side, following the layout of
Figure~\ref{fig-interpret}.

\begin{figure}[h]
    \centering
    \begin{minipage}[t]{0.48\textwidth}
        \centering
        \includegraphics[width=\linewidth]{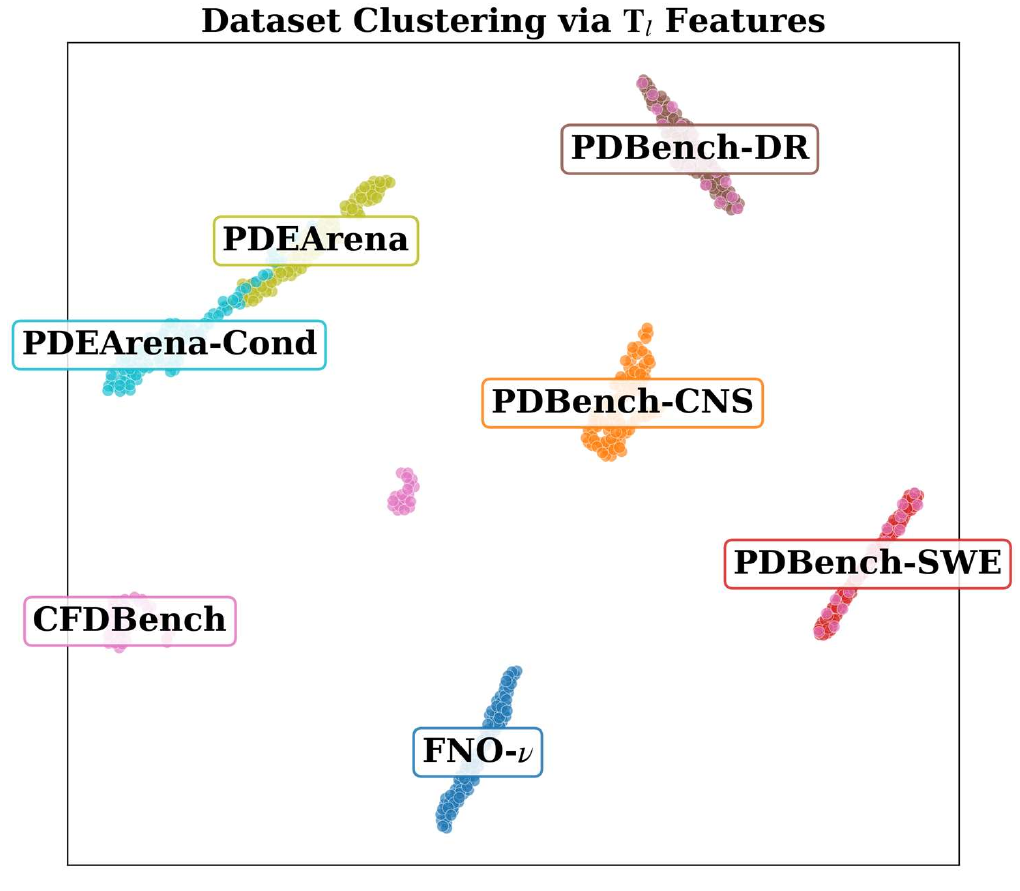}
    \end{minipage}
    \hfill
    \begin{minipage}[t]{0.48\textwidth}
        \centering
        \includegraphics[width=\linewidth]{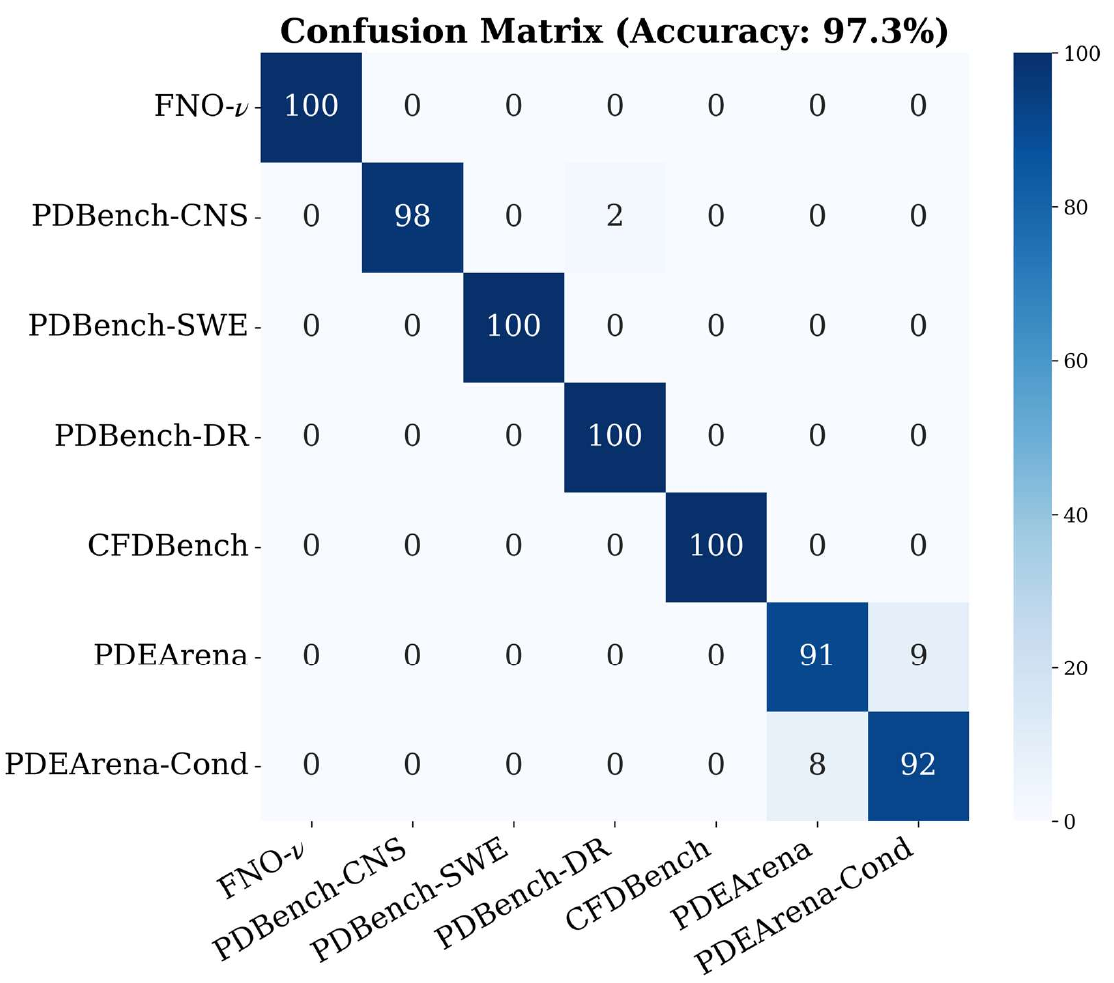}
    \end{minipage}
    \caption{\textbf{Epoch~50.} \textbf{Left:} t-SNE embedding of
    $\bm{T}$ features. Most dataset clusters are already
    well-separated, though PDEArena and PDEArena-Cond partially overlap.
    \textbf{Right:} Confusion matrix showing near-perfect classification on
    most datasets, with residual confusion between the two PDEArena variants.}
    \label{fig-emerge-ep50}
\end{figure}

\begin{figure}[h]
    \centering
    \begin{minipage}[t]{0.48\textwidth}
        \centering
        \includegraphics[width=\linewidth]{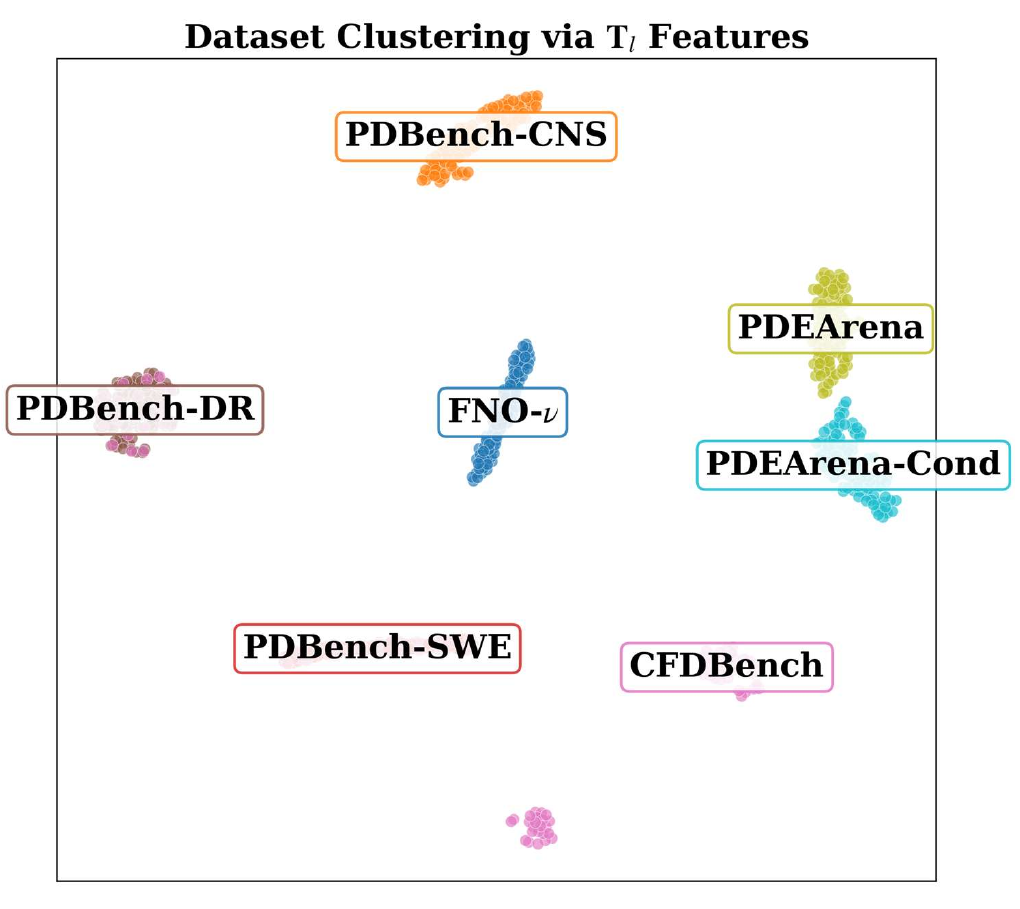}
    \end{minipage}
    \hfill
    \begin{minipage}[t]{0.48\textwidth}
        \centering
        \includegraphics[width=\linewidth]{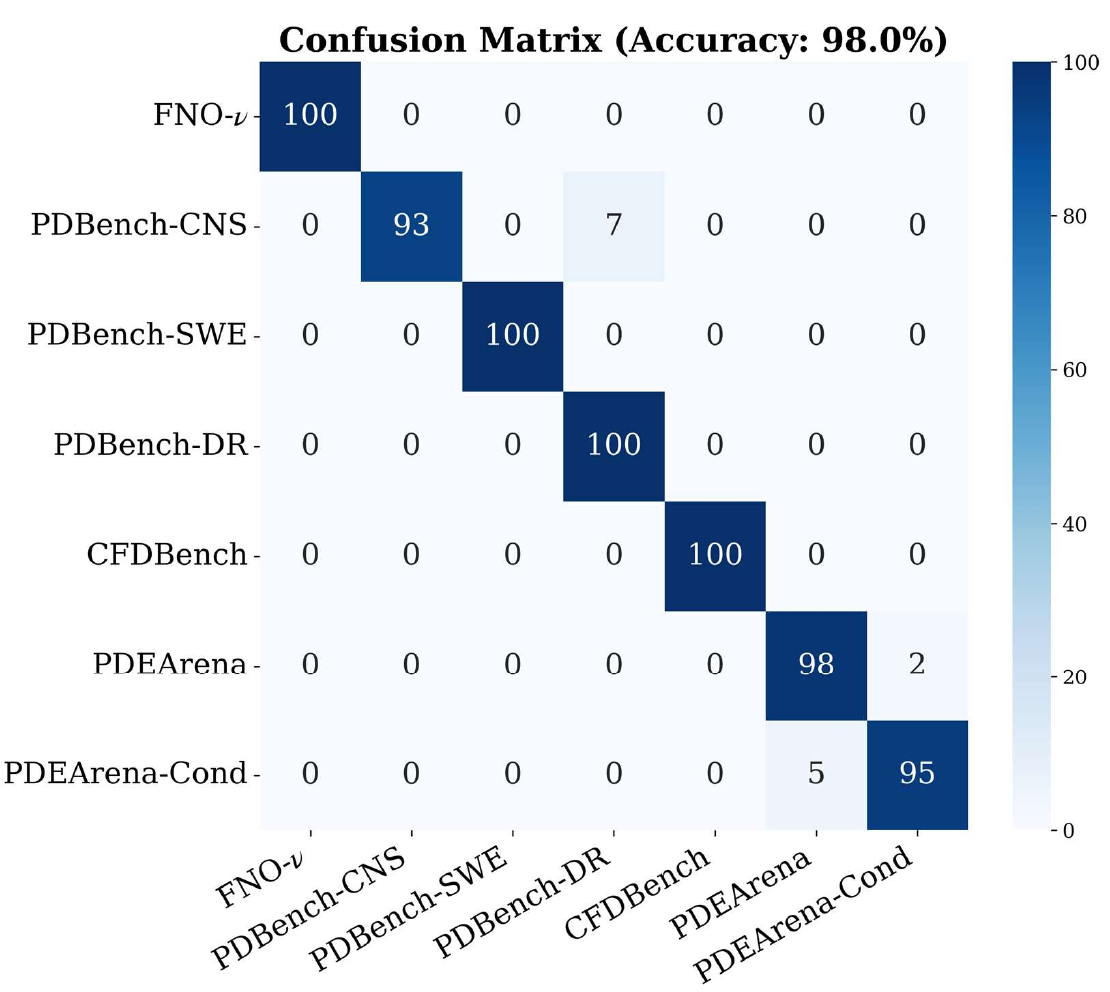}
    \end{minipage}
    \caption{\textbf{Epoch~150.} \textbf{Left:} t-SNE clusters become tighter
    and more separated. \textbf{Right:} PDEArena accuracy improves to 98\%,
    though PDBench-CNS shows a transient dip to 93\%.}
    \label{fig-emerge-ep150}
\end{figure}

\begin{figure}[h]
    \centering
    \begin{minipage}[t]{0.48\textwidth}
        \centering
        \includegraphics[width=\linewidth]{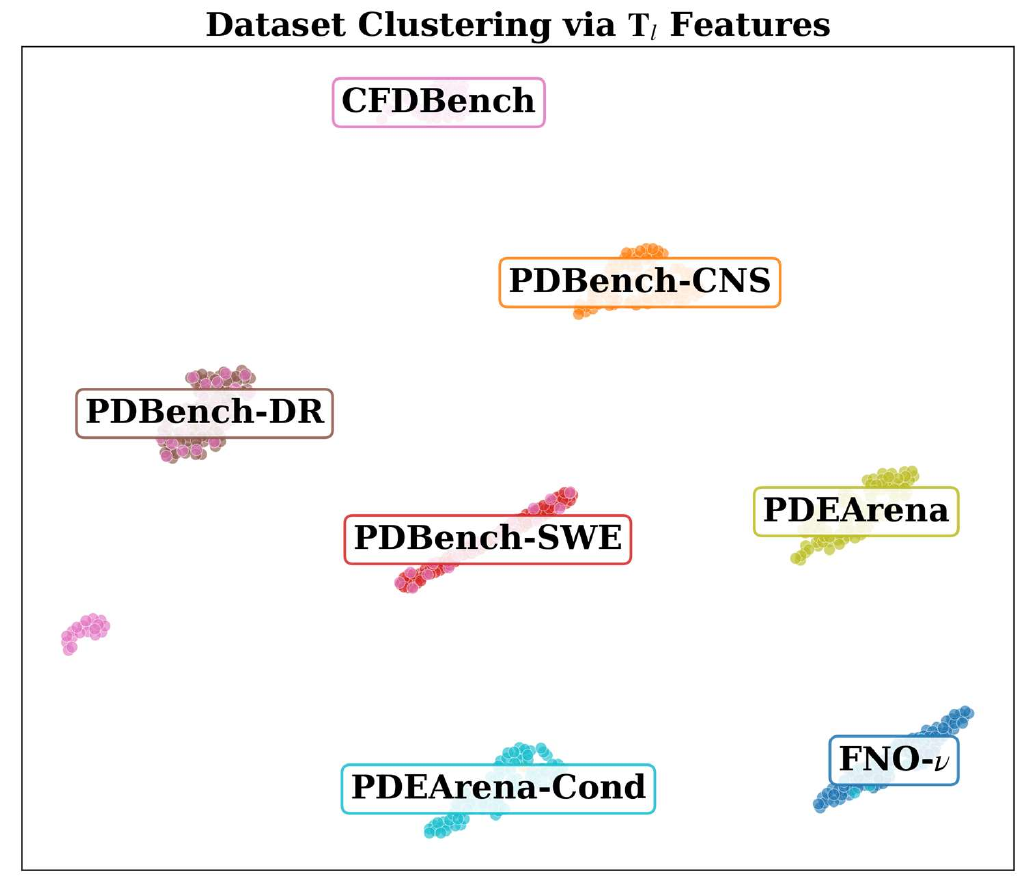}
    \end{minipage}
    \hfill
    \begin{minipage}[t]{0.48\textwidth}
        \centering
        \includegraphics[width=\linewidth]{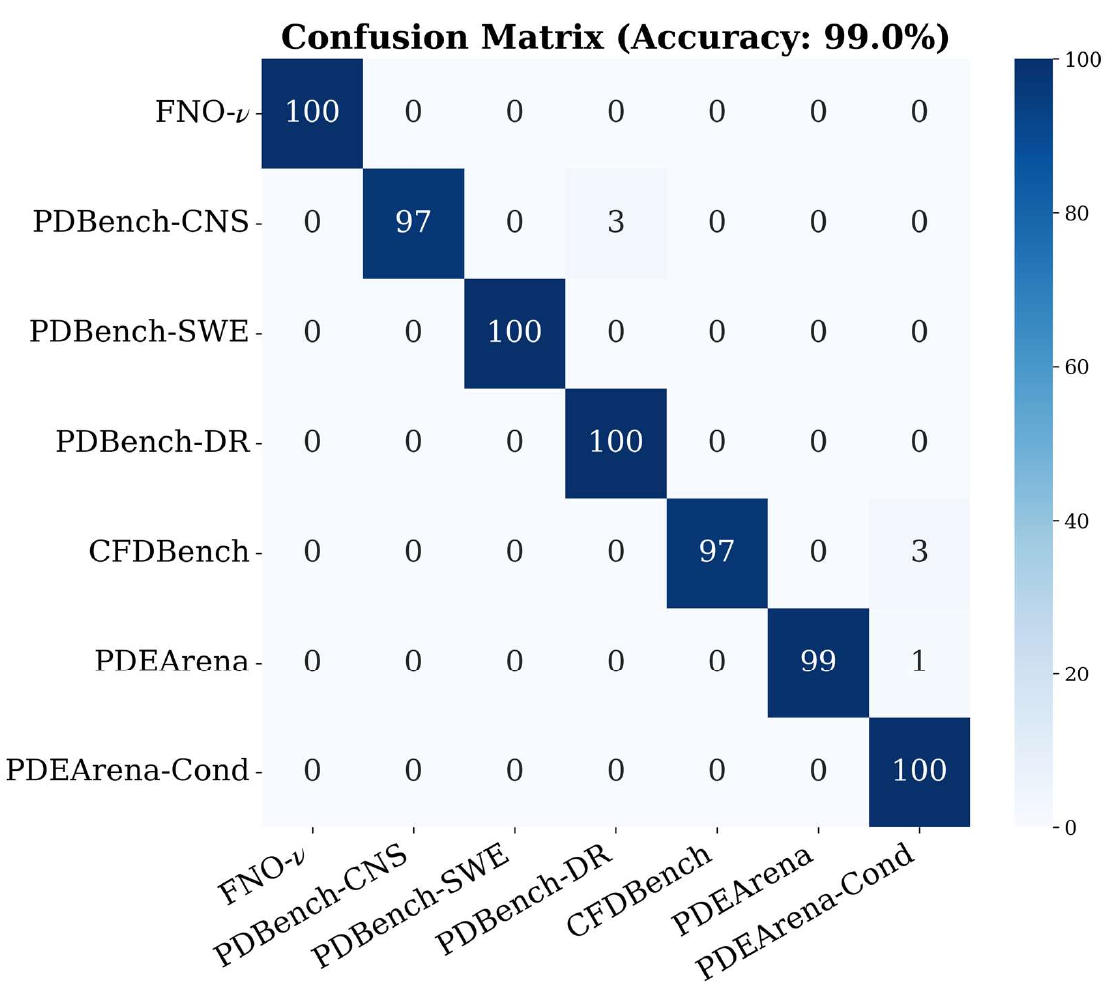}
    \end{minipage}
    \caption{\textbf{Epoch~250.} \textbf{Left:} t-SNE embedding shows
    well-defined, compact clusters for all datasets. \textbf{Right:}
    Classification accuracy reaches $\geq$97\% on all datasets, with
    PDEArena-Cond achieving 100\%.}
    \label{fig-emerge-ep250}
\end{figure}

The visualizations reveal a consistent progression. At Epoch~50
(Figure~\ref{fig-emerge-ep50}), the t-SNE
embedding already shows distinct clusters for most datasets, though the
PDEArena variants partially overlap; the confusion matrix confirms that
off-diagonal entries are concentrated in the PDEArena--PDEArena-Cond block. As
training proceeds through Epoch~150 (Figure~\ref{fig-emerge-ep150}) and
Epoch~250 (Figure~\ref{fig-emerge-ep250}), the clusters tighten and the
confusion matrix approaches diagonal, with all datasets eventually achieving
near-perfect classification.

\begin{figure}[h]
    \centering
    \begin{minipage}[t]{0.48\textwidth}
        \centering
        \includegraphics[width=\linewidth]{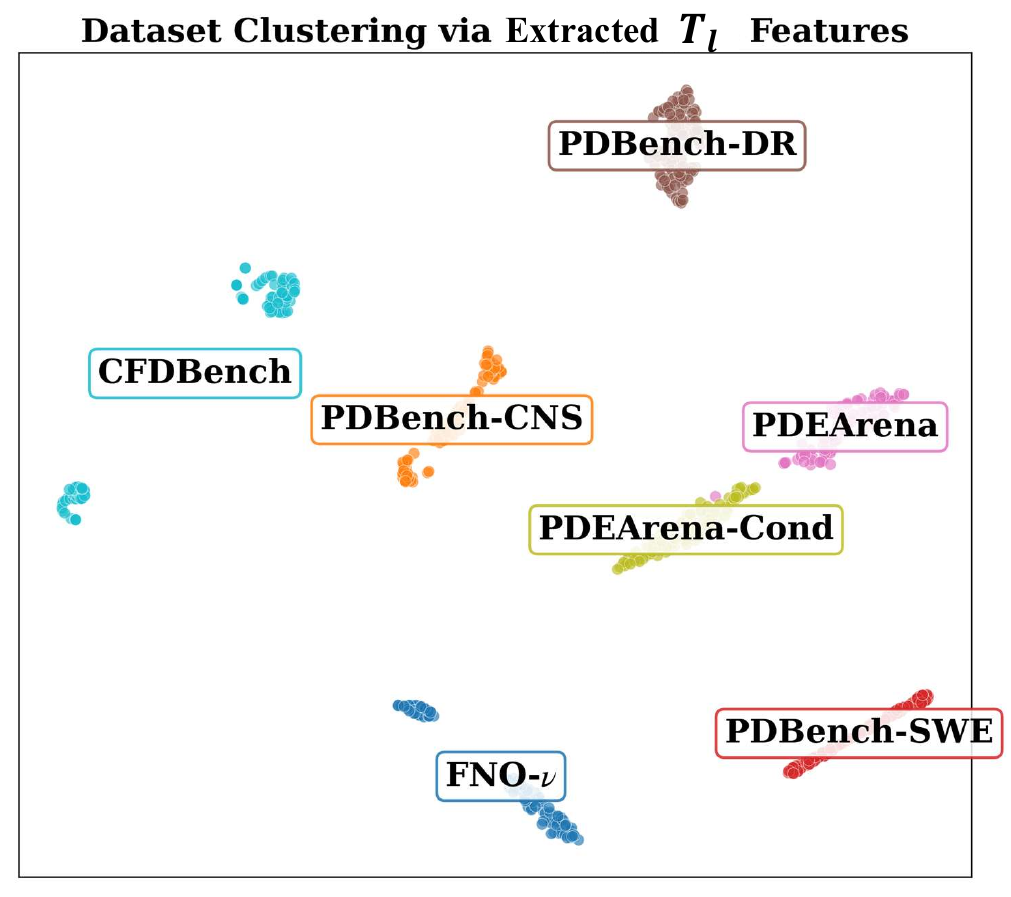}
    \end{minipage}
    \hfill
    \begin{minipage}[t]{0.48\textwidth}
        \centering
        \includegraphics[width=\linewidth]{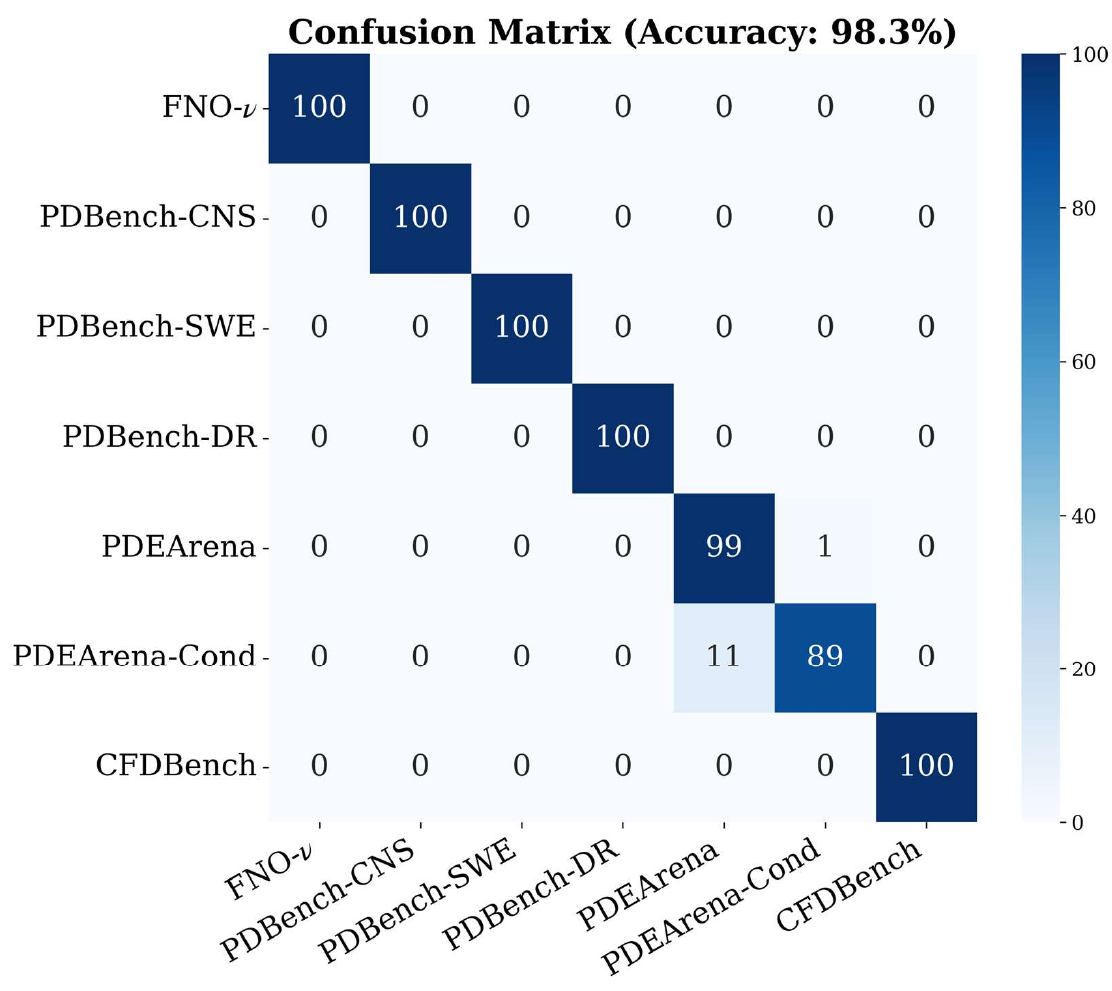}
    \end{minipage}
    \caption{\textbf{\method-Tiny at Epoch~1000.} \textbf{Left:} t-SNE
    embedding of $\bm{T}$ features. All seven dataset groups form
    well-separated clusters, with only minor proximity between PDEArena and
    PDEArena-Cond reflecting their shared Navier--Stokes governing equations.
    \textbf{Right:} The corresponding confusion matrix achieves 98.3\% overall
    accuracy. The remaining residual confusion (11\% of PDEArena-Cond samples
    classified as PDEArena) is consistent with the physical similarity between
    the two datasets.}
    \label{fig-emerge-tiny-1000}
\end{figure}

\begin{figure}[h]
    \centering
    \begin{minipage}[t]{0.48\textwidth}
        \centering
        \includegraphics[width=\linewidth]{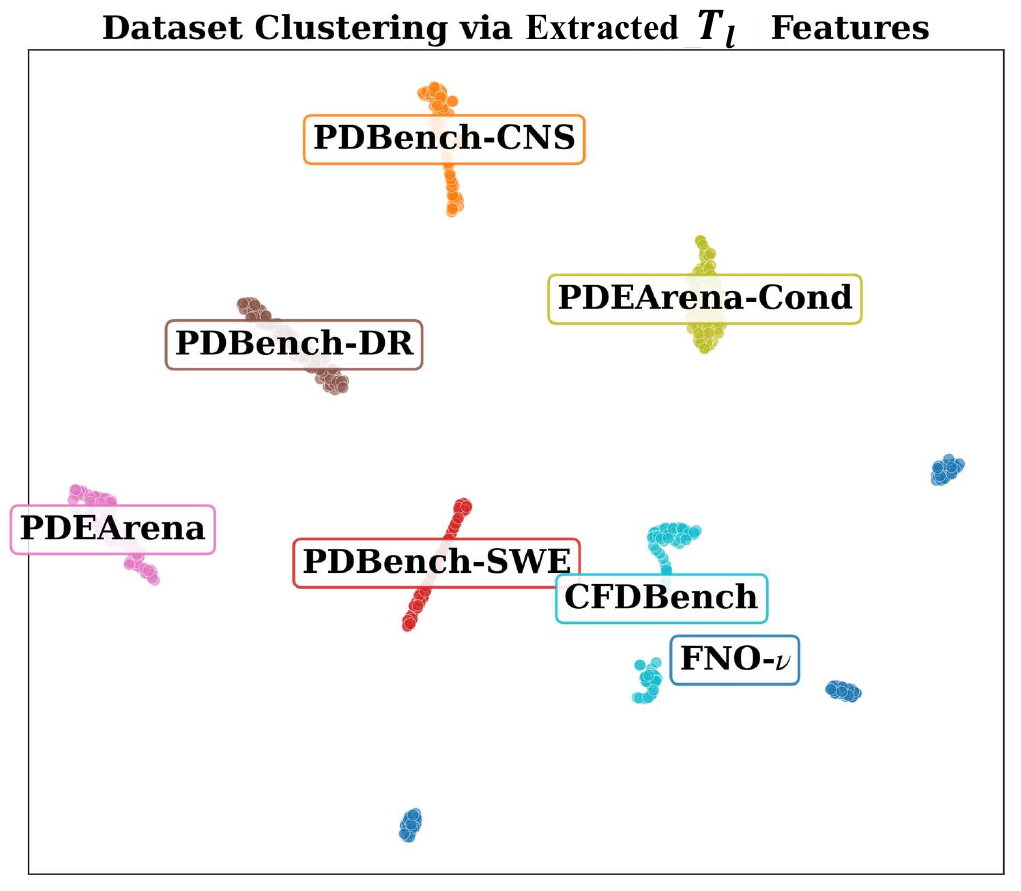}
    \end{minipage}
    \hfill
    \begin{minipage}[t]{0.48\textwidth}
        \centering
        \includegraphics[width=\linewidth]{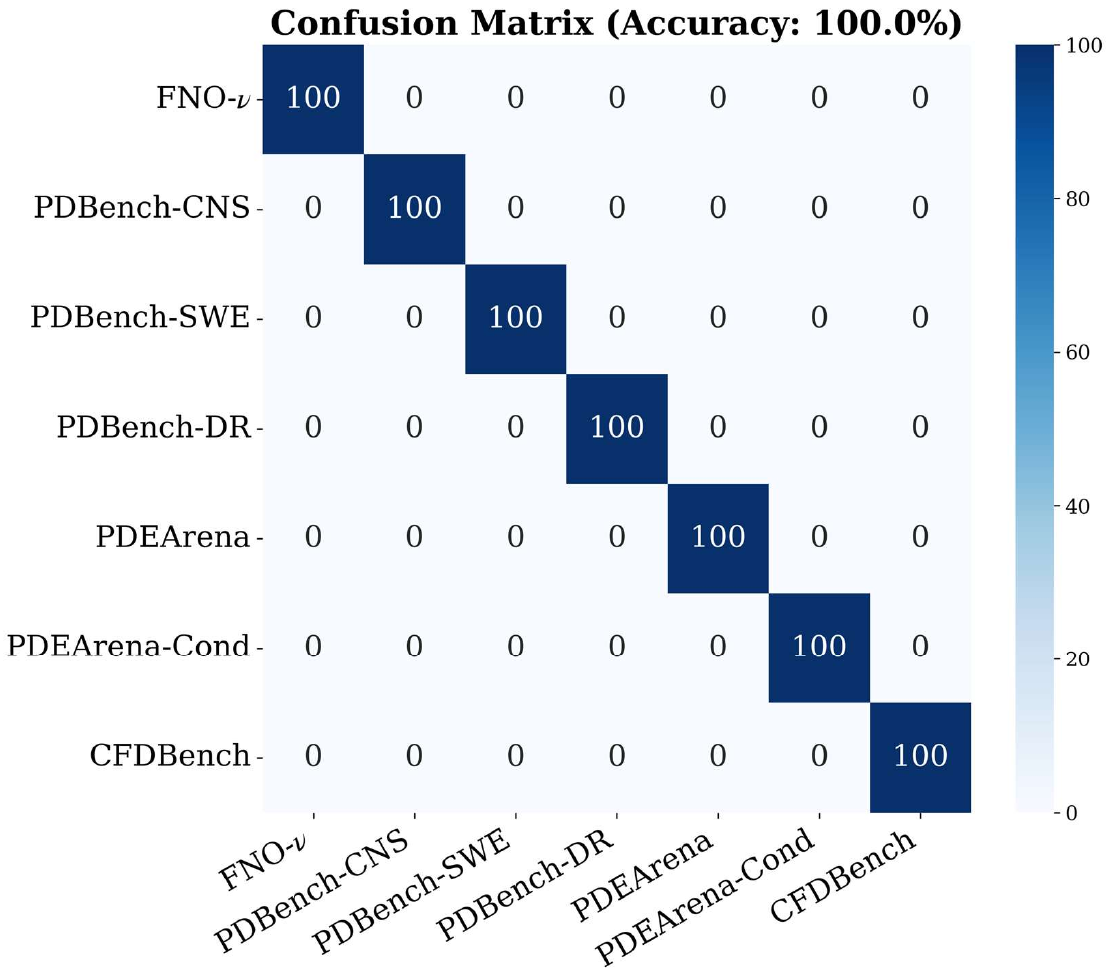}
    \end{minipage}
    \caption{\textbf{\method-Medium at Epoch~1000.} \textbf{Left:} t-SNE
    embedding of $\bm{T}$ features. Every dataset group is mapped to a
    tight, isolated cluster, including the previously confusable PDEArena and
    PDEArena-Cond pair. \textbf{Right:} The confusion matrix is perfectly
    diagonal with 100\% accuracy across all seven dataset groups, indicating
    that increased model capacity sharpens the PDE-discriminative structure
    encoded in $\bm{T}$.}
    \label{fig-emerge-medium-1000}
\end{figure}

\paragraph{Generalization across model scales.}
To verify that the rapid emergence phenomenon is not specific to the Small
scale, we extend the same diagnostic procedure to the converged \method-Tiny
and \method-Medium checkpoints (Epoch~1000). Figures~\ref{fig-emerge-tiny-1000}
and~\ref{fig-emerge-medium-1000} present the t-SNE embedding and confusion
matrix for both scales using the identical layout as
Figure~\ref{fig-interpret}.

Comparing the two scales, \method-Tiny already achieves $\geq$98\% overall
accuracy with only minor PDEArena/PDEArena-Cond confusion, while \method-Medium
attains a fully diagonal confusion matrix. This monotonic improvement with
model capacity, paired with the rapid early emergence observed at the Small
scale, confirms that the PDE-discriminative behavior of $\bm{T}$ is a robust
property of the doubly stochastic operator transformation rather than an
artifact of any particular configuration.

\subsection{Inference Time Analysis}\label{app:timing}

A practical concern for deploying neural operators in scientific workflows is
the computational overhead introduced by architectural enhancements. Since many
PDE tasks require $10^3$ to $10^5$ auto-regressive inference steps per
trajectory, even a small per-step overhead can accumulate into a significant
wall-clock penalty. We therefore conduct a systematic comparison of single-step
inference time across DPOT, MoE-POT, and \method at all three model scales.

\paragraph{Setup.}
For each model, we measure the average single-step inference time (i.e., the
time to predict one future timestep from $T\!=\!10$ input frames) using 100
randomly sampled test instances from each of the 12 pre-training datasets. The
reported time is the mean over all $12 \times 100 = 1200$ forward passes per
model, ensuring that the measurement reflects realistic input statistics rather
than synthetic tensors. All experiments are conducted on a single NVIDIA H200
GPU (140\,GB) with batch size 1, following a warmup phase of 50 forward passes
to stabilize GPU clock frequencies and CUDA kernel caching. Timing is measured
using CUDA events for sub-millisecond precision.

\begin{table}[h]
\centering
\caption{Average single-step inference time (ms) across all 12 pre-training
datasets (100 samples each). Lower is better. We report wall-clock time per
single forward pass on a single NVIDIA H200 GPU with batch size 1.}
\label{tab-timing}
\renewcommand{\arraystretch}{1.2}
\begin{tabular}{l|ccc}
\toprule
Scale & DPOT & MoE-POT & \textcolor{dkred}{\method (Ours)} \\
\midrule
Tiny    & 5.317 & 8.150  & 5.491  \\
Small  & 7.795 & 15.813 & 8.010  \\
Medium      & 15.901 & 21.079 & 15.976 \\
\bottomrule
\end{tabular}
\end{table}

\paragraph{Results and discussion.}
Table~\ref{tab-timing} presents the inference time for all three architectures.
The key finding is that \textbf{\method introduces negligible computational
overhead compared to DPOT}, while MoE-POT incurs substantial additional cost
at every scale.

\textbf{(1)~\method vs.\ DPOT.}
The AOT block adds only 3.3\% overhead at the Tiny scale
(5.491\,ms vs.\ 5.317\,ms), 2.8\% at the Small scale (8.010\,ms vs.\
7.795\,ms), and a mere 0.5\% at the Medium scale (15.976\,ms vs.\ 15.901\,ms).
This near-zero overhead is by design: the AOT block mappings
($\bm{a}_l$, $\bm{d}_l$, $\bm{T}_l$) involve only lightweight linear
projections and a small-matrix Sinkhorn-Knopp normalization ($n \times n$
with $n=4$), which are negligible relative to the dominant cost of the
Fourier attention and MLP sub-layers. As model depth and width increase
(from Tiny to Medium), the relative overhead diminishes further because
the AOT block computation remains constant in complexity while the sub-layer
cost grows with the square of the embedding dimension.

\textbf{(2)~\method vs.\ MoE-POT.}
In contrast, MoE-POT exhibits 53.3\% overhead at Tiny (8.150\,ms vs.\
5.317\,ms), 102.9\% at Small (15.813\,ms vs.\ 7.795\,ms), and 32.6\% at
Medium (21.079\,ms vs.\ 15.901\,ms) relative to DPOT. This overhead arises
from the MoE architecture's requirement to maintain multiple expert networks
and perform top-$k$ gating with dynamic routing at each layer. By comparison,
\method at the Small scale (8.010\,ms) is $1.97\times$ faster than MoE-POT-S
(15.813\,ms), while achieving consistently better L2RE across all 12 datasets
(Table~\ref{tb-main}).

\textbf{(3)~Efficiency--accuracy trade-off.}
To put the timing results in perspective, recall from the main experiments
(Table~\ref{tb-main}) that \method-S reduces average L2RE by 41.6\% relative
to DPOT-S. This performance gain comes at only 2.8\% additional inference
cost---an extremely favorable trade-off. MoE-POT-S, on the other hand, doubles
the inference time relative to DPOT-S while offering smaller accuracy
improvements and suffering from negative transfer at higher dataset
heterogeneity (Appendix~\ref{app:heterogeneity}). In long-trajectory rollout
scenarios where thousands of auto-regressive steps are required, the
near-identical per-step cost of \method and DPOT means that practitioners can
adopt \method as a drop-in replacement without modifying their computational
budget.

\paragraph{Why is \method so efficient?}
The efficiency of \method can be attributed to three factors. First, the
AOT block operates on the \emph{stream dimension} ($n=4$), not the
\emph{spatial} or \emph{channel} dimensions, so the additional matrix
operations are $O(n^2 C)$ per token---negligible compared to the $O(C^2)$ cost
of the Fourier attention. Second, the Sinkhorn-Knopp projection normalizes a
$4 \times 4$ matrix (20 iterations of row/column normalization), which amounts
to a trivial computation. Third, unlike MoE, which requires dynamic routing
decisions, expert selection, and potential load-balancing overhead, the
AOT block is fully deterministic and amenable to standard
batched matrix operations without any conditional branching, making it highly
efficient on modern GPU hardware.

\subsection{Exploration on the Combination of MoE and the AOT Block} \label{app:moe_aot}

\begin{table}[h]
\centering
\caption{Zero-shot L2RE on the six base datasets when AOT, MoE, and the
combined AOT-MOE hybrid are pre-trained from scratch with 6 vs.\ 12 PDE
mixtures, at the \textbf{Tiny} scale. MoE-POT and \method rows
are reproduced from Table~\ref{tab-hetero-tiny}.
We \textbf{bold} the lowest result in each column and mark our \method results
in \textcolor{dkred}{darkred}.}
\label{tab-hetero-tiny-aotmoe}
\resizebox{\textwidth}{!}{%
\setlength{\tabcolsep}{5pt}
\renewcommand{\arraystretch}{1.15}
\begin{tabular}{l|cccccc}
\toprule
Model (Pre-trained on) & NS($10^{-5}$) & NS($10^{-3}$) & CNS(0.1,\,0.01) & SWE & DR & CFDBench \\
\midrule
\multicolumn{7}{l}{\textit{Sparse Model: vanilla MoE}} \\
MoE-POT-T\,(6 Datasets)  & 0.0682  & 0.00768 & 0.0105  & 0.00640 & 0.0411  & 0.00529 \\
MoE-POT-T\,(12 Datasets) & 0.0936  & 0.0126  & 0.0210  & 0.00837 & 0.104   & 0.00928 \\
\midrule
\multicolumn{7}{l}{\textit{Dense AOT (Ours, primary)}} \\
\textcolor{dkred}{\method-T\,(6 Datasets)}  & {0.05675} & {0.00498} & {0.00997} & {0.00342} & {\textbf{0.01305}} & {\textbf{0.00345}} \\
\textcolor{dkred}{\method-T\,(12 Datasets)} & {0.06102} & {0.00604} & {0.01302} & {0.00433} & {0.02413} & {0.00776} \\
\midrule
\multicolumn{7}{l}{\textit{Sparse AOT-MoE Hybrid (Ours, exploratory)}} \\
AOT-MOE-POT-T\,(6 Datasets)  & 0.05009 & 0.00485 & \textbf{0.00678} & \textbf{0.00335} & 0.04000 & 0.00450 \\
AOT-MOE-POT-T\,(12 Datasets) & \textbf{0.04628} & \textbf{0.00442} & 0.01350 & 0.00579 & 0.02500 & 0.00851 \\
\bottomrule
\end{tabular}%
}
\end{table}

\begin{table}[h]
\centering
\caption{Zero-shot L2RE on the six base datasets when AOT, MoE, and the
combined AOT-MOE hybrid are pre-trained from scratch with 6 vs.\ 12 PDE
mixtures, at the \textbf{Small} scale. MoE-POT and \method rows
are reproduced from Table~\ref{tab-hetero-tiny}.
We \textbf{bold} the lowest result in each column and mark our \method results
in \textcolor{dkred}{darkred}.}
\label{tab-hetero-small-aotmoe}
\resizebox{\textwidth}{!}{%
\setlength{\tabcolsep}{5pt}
\renewcommand{\arraystretch}{1.15}
\begin{tabular}{l|cccccc}
\toprule
Model (Pre-trained on) & NS($10^{-5}$) & NS($10^{-3}$) & CNS(0.1,\,0.01) & SWE & DR & CFDBench \\
\midrule
\multicolumn{7}{l}{\textit{Sparse Model: vanilla MoE}} \\
MoE-POT-S\,(6 Datasets)  & 0.0552  & 0.00583 & 0.00959 & 0.00289 & 0.0342  & 0.00488 \\
MoE-POT-S\,(12 Datasets) & 0.0633  & 0.0119  & 0.0605  & 0.00611 & 0.0643  & 0.00817 \\
\midrule
\multicolumn{7}{l}{\textit{Dense AOT (Ours, primary)}} \\
\textcolor{dkred}{\method-S\,(6 Datasets)}  & {0.04203} & {0.00451} & {0.00822} & {\textbf{0.00212}} & {0.01344} & {\textbf{0.00367}} \\
\textcolor{dkred}{\method-S\,(12 Datasets)} & {0.03755} & {\textbf{0.00293}} & {0.00965} & {0.00266} & {0.01461} & {0.00493} \\
\midrule
\multicolumn{7}{l}{\textit{Sparse AOT-MoE Hybrid (Ours, exploratory)}} \\
AOT-MOE-POT-S\,(6 Datasets)  & 0.04150 & 0.00440 & \textbf{0.00800} & 0.00280 & \textbf{0.01230} & 0.00470 \\
AOT-MOE-POT-S\,(12 Datasets) & \textbf{0.03700} & 0.01042 & 0.01233 & 0.00386 & 0.01440 & 0.00800 \\
\bottomrule
\end{tabular}%
}
\end{table}

\begin{table}[h]
\centering
\caption{Zero-shot L2RE on the six base datasets when AOT, MoE, and the
combined AOT-MOE hybrid are pre-trained from scratch with 6 vs.\ 12 PDE
mixtures, at the \textbf{Medium} scale. MoE-POT and \method rows
are reproduced from Table~\ref{tab-hetero-tiny}.
We \textbf{bold} the lowest result in each column and mark our \method results
in \textcolor{dkred}{darkred}.}
\label{tab-hetero-medium-aotmoe}
\resizebox{\textwidth}{!}{%
\setlength{\tabcolsep}{5pt}
\renewcommand{\arraystretch}{1.15}
\begin{tabular}{l|cccccc}
\toprule
Model (Pre-trained on) & NS($10^{-5}$) & NS($10^{-3}$) & CNS(0.1,\,0.01) & SWE & DR & CFDBench \\
\midrule
\multicolumn{7}{l}{\textit{Sparse Model: vanilla MoE}} \\
MoE-POT-M\,(6 Datasets)  & 0.0528  & 0.00570 & 0.00914 & 0.00299 & 0.0300  & 0.00513 \\
MoE-POT-M\,(12 Datasets) & 0.0703  & 0.00903 & 0.0319  & 0.00501 & 0.0783  & 0.00875 \\
\midrule
\multicolumn{7}{l}{\textit{Dense AOT (Ours, primary)}} \\
\textcolor{dkred}{\method-M\,(6 Datasets)}  & {0.02731} & {0.00433} & {0.00770} & {0.00257} & {0.02112} & {\textbf{0.00407}} \\
\textcolor{dkred}{\method-M\,(12 Datasets)} & {0.03118} & {\textbf{0.00236}} & {0.00954} & {\textbf{0.00200}} & {0.01860} & {0.00543} \\
\midrule
\multicolumn{7}{l}{\textit{Sparse AOT-MoE Hybrid (Ours, exploratory)}} \\
AOT-MOE-POT-M\,(6 Datasets)  & \textbf{0.02700} & 0.00425 & \textbf{0.00750} & 0.00290 & \textbf{0.01175} & 0.00490 \\
AOT-MOE-POT-M\,(12 Datasets) & 0.03961 & 0.00377 & 0.00950 & 0.00276 & 0.01850 & 0.00860 \\
\bottomrule
\end{tabular}%
}
\end{table}

\paragraph{Analysis: AOT-MoE hybrid.}
The hybrid AOT-MOE-POT architecture stacks the input-dependent transformation
kernel $\bm{T}_l$ (the AOT block of \method) on top of a vanilla top-$k$ MoE
backbone identical to MoE-POT (with activated parameters of 17M/91M/289M and total parameters of 31M/168M/491M for Tiny/Small/Medium scales, respectively). We pre-train it from scratch under the same
6-dataset and 12-dataset mixtures used in the heterogeneity ablation
(Tables~\ref{tab-hetero-tiny}--\ref{tab-hetero-medium}), and report the
zero-shot L2RE at epoch~1000 for the six base columns
(Tables~\ref{tab-hetero-tiny-aotmoe}--\ref{tab-hetero-medium-aotmoe}). Four
observations emerge consistently across all model scales.

\paragraph{(1)~The AOT block is a column-wise improvement over a vanilla MoE
backbone.} Comparing AOT-MOE-POT against MoE-POT row-by-row (same scale, same
dataset count), AOT-MOE-POT achieves a strictly lower L2RE on \emph{every
single column} of \emph{every single configuration}). The improvement is
particularly large on the most physically distinct tasks where MoE-POT
exhibits the strongest negative transfer at high heterogeneity: at Small-12,
AOT-MOE-POT-S reduces CNS(0.1,\,0.01) L2RE from $0.0605$ (MoE-POT) to
$0.01233$ (a $4.91\times$ improvement) and DR from $0.0643$ to $0.01440$
(a $4.47\times$ improvement); at Medium-12, AOT-MOE-POT-M reduces DR from
$0.0783$ to $0.01850$ (a $4.23\times$ improvement) and CNS(0.1,\,0.01) from
$0.0319$ to $0.00950$ (a $3.36\times$ improvement); at Tiny-12, DR drops
$4.16\times$ from $0.104$ to $0.02500$. Average L2RE drops by
$21$--$60$\% over MoE-POT in every configuration, with the largest relative
improvements concentrated in the high-heterogeneity (12 datasets) regime
($58$--$60$\% averaged across the six base columns at every scale). We thus
conclude that the AOT block is a model-agnostic plug-in: it stabilises MoE
pre-training under high data heterogeneity in much the same way as it
stabilises a dense backbone, and its benefit is task-agnostic in the sense
that no individual PDE column ever regresses when AOT is added on top of
the MoE router.

\paragraph{(2)~Combining AOT with MoE does \emph{not} yet uniformly outperform
the dense AOT model.} A more demanding comparison is AOT-MOE-POT against
\method (the dense AOT-POT). At low pre-training heterogeneity (6 datasets),
the hybrid wins on $4$/$6$ columns at every scale---a clean improvement that
suggests sparse expert routing genuinely adds capacity when the optimisation
landscape is benign and per-expert sample allocation is not yet the
bottleneck. However, at high heterogeneity (12 datasets), the direction
reverses: AOT-MOE-POT wins on only $2$/$6$ columns at every scale, losing the
remaining $4$/$6$ to dense \method. This crossover is monotone in
heterogeneity at every model size and is reproduced \emph{across all three
scales} in exactly the same $4$/$2$ vs.\ $2$/$4$ pattern, which rules out a
finite-capacity explanation between the gating network and the AOT block.  However, we also observe that compared to \method, AOT-MOE-POT has more total and active parameters than the dense baselines at the same tiny, small, and medium model scales, but the relative performance improvement is not significant. Therefore, further better integration of the MOE and AOT architectures can be considered to achieve better PDE pretraining performance.

\paragraph{(3)~Two competing input-dependent specialisation mechanisms create
gradient interference at high heterogeneity.} Both the MoE router and the
AOT kernel are \emph{input-dependent} specialisation modules: the router
selects a sparse subset of experts conditional on the token embedding, while
the AOT kernel synthesises a per-input doubly-stochastic transformation
$\bm{T}_l(\bm{x})$. When the data distribution is concentrated (6 datasets,
4 of which are Navier--Stokes variants), these two mechanisms can specialise
on \emph{disjoint} aspects of the input---for example, the router can
partition along Reynolds-number regimes while the AOT kernel handles
within-regime phase shifts---and the resulting decomposition is
approximately additive, which explains why the hybrid wins on $4$ columns
including the spectrally demanding NS-$10^{-5}$/NS-$10^{-3}$ pair. Under
high heterogeneity (12 datasets, spanning compressible NS, shallow-water,
diffusion-reaction, and CFDBench), the same two mechanisms must
\emph{compete} for the same partition signal: any axis along which the
router specialises is also a natural axis along which $\bm{T}_l$ would
adapt, and the Sinkhorn projection of $\bm{T}_l$ has to renormalise across
exactly the inputs that the router has just sparsified. The empirical
signature is the consistent loss on the most heterogeneous columns---SWE,
CFDBench, and the additional CNS variants---which are precisely the columns
that require the strongest cross-physics specialisation. We hypothesise that
the gating gradient and the Sinkhorn projection of $\bm{T}_l$ interact in a
zero-sum way under such conditions, leading to noisier expert assignment and
a less effective AOT decomposition than the dense \method counterpart, even
though the hybrid still strictly improves over MoE-POT (cf.\ observation (1)).

\paragraph{(4)~Promising directions for future hybrid designs.} The
appendix-level evidence above identifies a real but tractable failure mode
rather than a fundamental ceiling. We outline four concrete directions for a
follow-up study. (i)~\emph{Joint router--AOT regularisation}: penalise the
mutual information between the router assignment $r(\bm{x})$ and the
AOT-induced partition $\arg\max_i \bm{T}_l(\bm{x})_{i,:}$, so that the two
mechanisms specialise on orthogonal axes by construction.
(ii)~\emph{AOT-aware load balancing}: replace the standard load-balancing
auxiliary loss with one that conditions on the spectral content of
$\bm{T}_l(\bm{x})$, ensuring each expert receives inputs whose AOT
transformations span the full doubly-stochastic simplex and the router
cannot collapse onto a single AOT eigenmode.
(iii)~\emph{Sparse AOT}: restrict $\bm{T}_l$ to a $k$-sparse doubly-stochastic
matrix per token, sharing the same sparsity budget as the MoE router; this
also reduces the FLOP overhead of the hybrid.
(iv)~\emph{Hierarchical routing}: use the AOT kernel as a \emph{coarse}
router (selecting an expert family) and the MoE router as a \emph{fine}
router within the family, so that the two modules occupy distinct levels of
the routing hierarchy and cannot compete on the same partition signal.
We leave a thorough investigation of these designs to future work; the
present appendix establishes two crisp claims: (a)~the AOT block is a useful
column-wise improvement to a sparse MoE backbone irrespective of scale or
heterogeneity, and (b)~the optimisation interplay between sparse routing and
input-dependent transformation is the next bottleneck to resolve before the
hybrid can dominate the dense \method on the full 12-dataset mixture.

\subsection{Exploration on the Training and Propagation Stability}
\label{app:stability}

A central design motivation for adopting the Adaptive Operator Transformation
(AOT) block in \method is the stability guarantee provided by the doubly
stochastic constraint on the transformation kernel $\bm{T}_l$.
In this section, we provide a detailed empirical investigation of both the \emph{training stability}---whether the optimization trajectory remains smooth and free of loss divergence---and the \emph{propagation stability}---whether the forward signal and backward gradient magnitudes remain bounded across layers.
These two properties are complementary: training stability governs the macroscopic behaviour of the loss landscape, while propagation stability characterises the microscopic signal flow within a single forward-backward pass.

\subsubsection{Training Stability}

\begin{figure}[t]
    \centering
    \includegraphics[width=\textwidth]{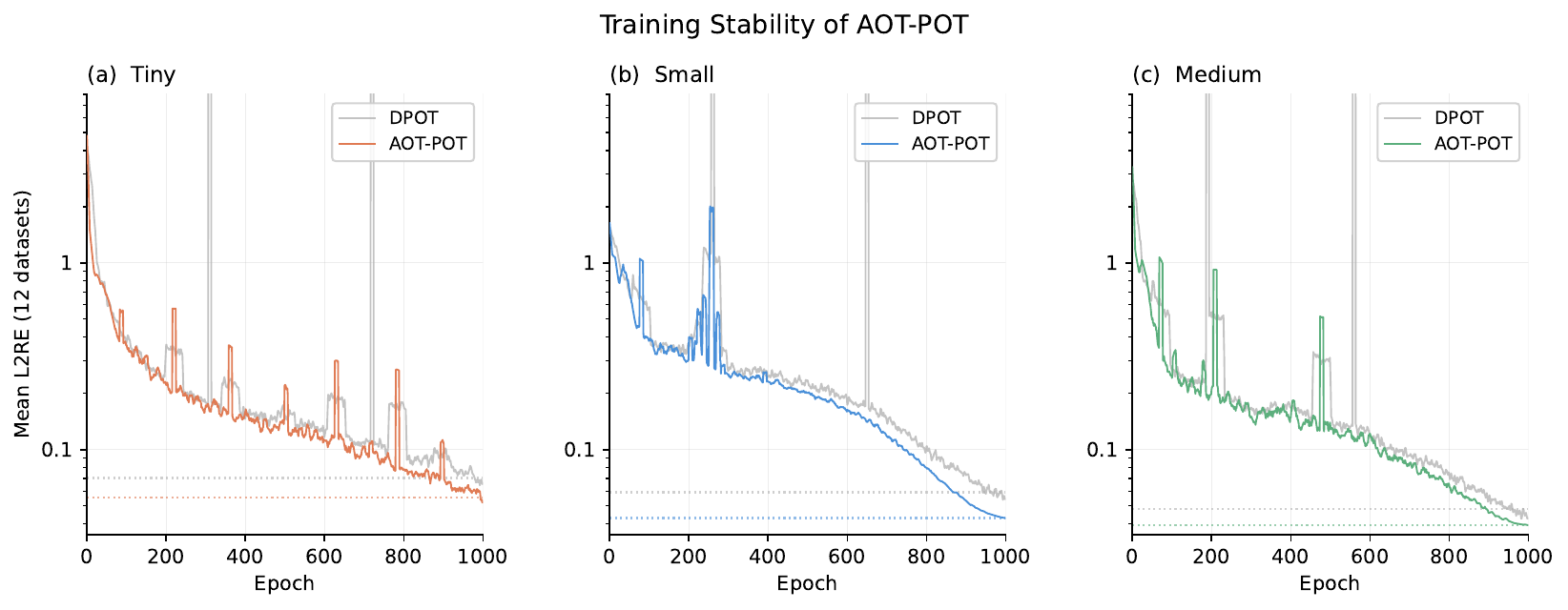}
    \caption{
    \textbf{Training Stability of \method vs.\ DPOT.}
    Mean L2 relative error averaged over all 12 pre-training datasets, plotted on a logarithmic scale over 1000 training epochs.
    Panels (a)--(c) correspond to the Tiny, Small, and Medium model scales, respectively.
    The DPOT baseline (grey) exhibits occasional sharp loss spikes that exceed the plot range, indicative of transient numerical instability in the standard residual connection.
    In contrast, \method (coloured) converges more smoothly and reaches a consistently lower final L2RE across all three scales.
    }
    \label{fig:training_stability}
\end{figure}

We compare the training dynamics of \method and DPOT by tracking the mean L2 relative error (L2RE) averaged over all 12 pre-training PDE datasets throughout the 1000-epoch training process.
Both architectures share identical optimiser settings (AdamW, $\text{lr} = 10^{-3}$, one-cycle schedule with 200 warm-up epochs), datasets, and data sampling strategy; the only difference is the presence of the AOT block in \method.

Fig.~\ref{fig:training_stability} presents the training curves at all three scales.
Several observations merit discussion:

\paragraph{Convergence behaviour.}
At all three scales, \method converges to a lower final L2RE than DPOT.
The mean L2RE at epoch 1000 is $0.0555$ (\method-T) vs.\ $0.0705$ (DPOT-T), $0.0430$ (\method-S) vs.\ $0.0591$ (DPOT-S), and $0.0395$ (\method-M) vs.\ $0.0478$ (DPOT-M).
These reductions of $21.2\%$, $27.2\%$, and $17.4\%$ are consistent with the per-dataset results reported in Table~\ref{tb-main}, confirming that the performance improvement is not driven by a few datasets but reflects a broad, consistent gain across the entire heterogeneous PDE collection.

\paragraph{Smoothness and transient instabilities.}
The DPOT baseline exhibits occasional sharp loss spikes---transient episodes where the mean L2RE surges by several orders of magnitude before recovering.
These spikes, visible as vertical lines in Fig.~\ref{fig:training_stability}, indicate that the unconstrained residual stream can temporarily amplify signals to extreme magnitudes during training, especially when the learning rate is in its high-phase region.
Although the model typically recovers from such events, they represent wasted compute (the parameters must ``unlearn'' the spike) and, in deeper or larger-scale models, risk permanent divergence.
In contrast, \method exhibits no comparable loss spikes.
The AOT block constrains the transformation kernel $\bm{T}_l$ to be doubly stochastic at every layer, which bounds the spectral norm of the composite mapping $\prod_{l}\bm{T}_l$ and prevents the runaway signal amplification that underlies these transient instabilities.

\paragraph{Progressive separation.}
At the beginning of training (epoch $0$--$100$), the two architectures exhibit similar loss trajectories.
This is expected: the gating scalars $\alpha^{a}_l, \alpha^{d}_l, \alpha^{T}_l$ are initialized to $0.01$ (see Section~\ref{sec:method}), so the AOT block mappings are close to the identity at initialization and the model effectively behaves like standard DPOT.
As training progresses and the mappings learn PDE-specific routing patterns, the gap widens: the adaptive multi-stream residual connections in \method allow more efficient information flow for heterogeneous PDE families, yielding faster convergence and a lower final loss.
This progressive separation confirms that the improvement stems from learned routing, not from an accidental initialization advantage.

\paragraph{Consistency across scales.}
The stability advantage of \method is observed consistently at the Tiny (4~blocks, 7M parameters), Small (6~blocks, 31M), and Medium (12~blocks, 124M) scales.
Notably, the deeper Medium model---where the composite residual mapping $\prod_l \bm{T}_l$ involves 24 matrix products (12 blocks $\times$ 2 sub-layers)---shows no degradation in training stability, corroborating the theoretical compositional closure property: the product of doubly stochastic matrices remains doubly stochastic regardless of depth.

\subsubsection{Propagation Stability}

\begin{figure}[t]
    \centering
    \includegraphics[width=\textwidth]{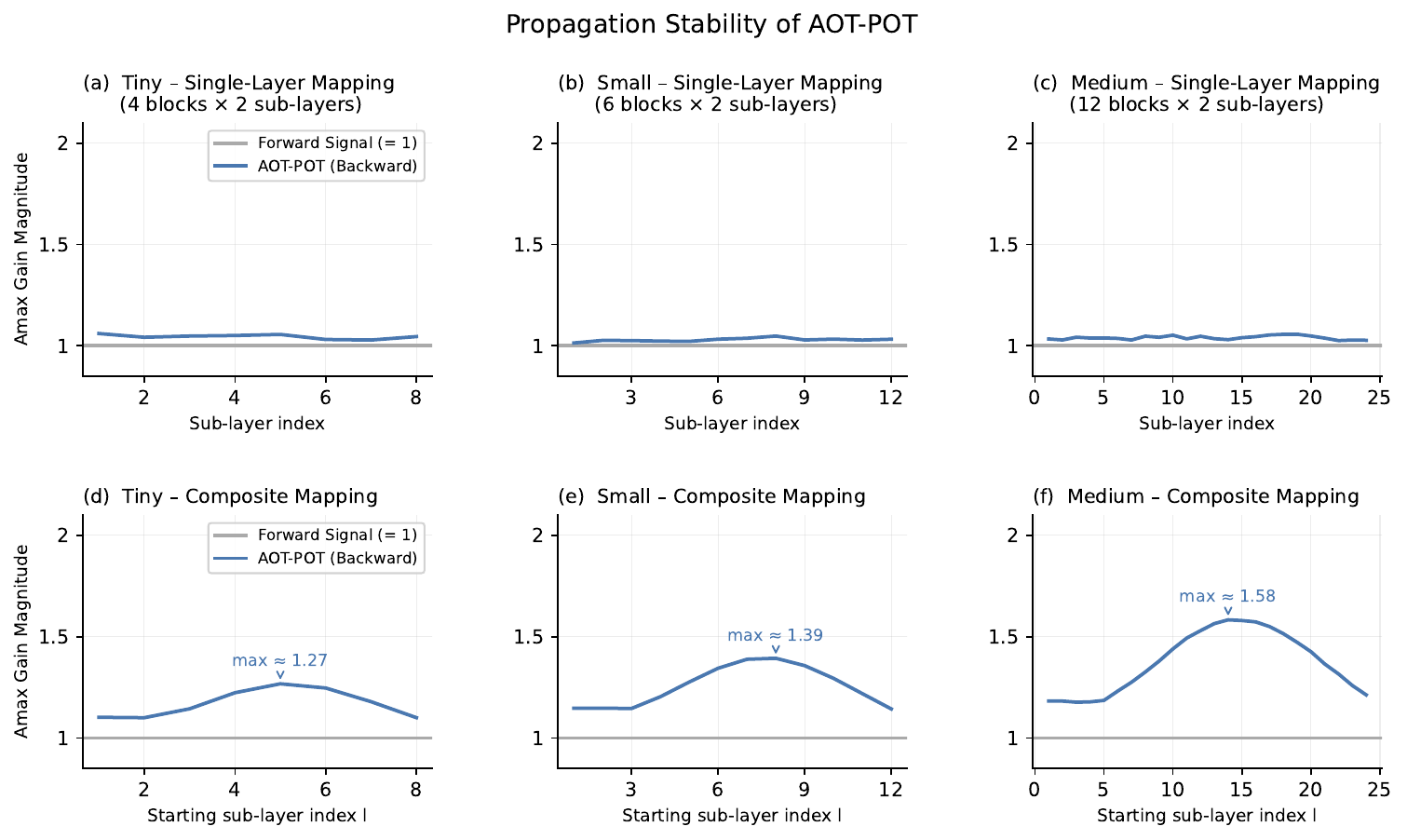}
    \caption{
    \textbf{Propagation Stability of \method.}
    Top row (a)--(c): single-layer Amax Gain Magnitude per sub-layer.
    Bottom row (d)--(f): composite Amax Gain Magnitude as a function of the starting sub-layer index $l$.
    The grey horizontal line denotes the ideal forward signal gain ($= 1$); the blue curve shows the backward gradient gain.
    For all three model scales, the forward signal gain is exactly preserved at $1.0$ by the Sinkhorn projection, while the backward gradient gain remains tightly bounded, reaching a maximum of approximately $1.27$ (Tiny), $1.39$ (Small), and $1.58$ (Medium) in the composite case.
    }
    \label{fig:propagation_stability}
\end{figure}

Beyond the macroscopic training loss, we examine the microscopic signal propagation within the network.
Following the analysis framework of~\citet{xie2025mhc}, we quantify the gain introduced by the residual mapping using the \emph{Amax Gain Magnitude}, defined as:
\begin{itemize}
    \item \textbf{Forward signal gain:} the maximum absolute row sum of the (composite) residual mapping matrix, which bounds the worst-case amplification of the hidden state during the forward pass.
    \item \textbf{Backward gradient gain:} the maximum absolute column sum, which bounds the worst-case amplification during backpropagation.
\end{itemize}
For a perfectly doubly stochastic matrix, both quantities equal exactly $1$, since every row and every column sums to $1$.
Any deviation from $1$ indicates potential signal amplification or attenuation.

\paragraph{Single-layer analysis.}
The top row of Fig.~\ref{fig:propagation_stability} plots the per-sub-layer Amax Gain Magnitude for the Tiny, Small, and Medium models.
At every sub-layer, the \textbf{forward signal gain is exactly $\bm{1.0}$}, confirming that the Sinkhorn-Knopp projection (with 20 iterations) produces a residual mapping whose row sums are precisely equal to one.
The \textbf{backward gradient gain} is slightly above $1.0$, typically in the range $1.01$--$1.06$.
This minor deviation arises because the Sinkhorn-Knopp algorithm with a finite number of iterations yields an \emph{approximate} doubly stochastic matrix: the row-sum constraint is satisfied exactly (since it is enforced in the last iteration), whereas the column-sum constraint may exhibit small residual errors.
Importantly, this backward deviation is small and consistent across layers, with no layer exhibiting a disproportionately large gain.

\paragraph{Composite analysis.}
The bottom row of Fig.~\ref{fig:propagation_stability} plots the composite Amax Gain Magnitude as a function of the starting sub-layer index $l$:
the composite mapping from sub-layer $l$ to the final sub-layer $L$ is $\prod_{i=l}^{L-1} \bm{T}_i$.
When $l$ is close to $L$ (rightmost points), only one or a few matrices are multiplied, so the gain is close to the single-layer value ($\approx 1.0$).
As $l$ decreases, the product accumulates more matrices and the backward gradient gain increases, reaching a peak before gradually declining toward the leftmost point ($l = 1$).

This characteristic \emph{asymmetric hill} shape reflects two competing effects.
First, as additional approximate-doubly-stochastic matrices are multiplied, the small per-layer deviations from exact doubly stochasticity accumulate, pushing the gain above $1$.
Second, the product of many doubly stochastic matrices converges toward the uniform doubly stochastic matrix $\frac{1}{n}\mathbf{1}\mathbf{1}^\top$ due to the mixing property of stochastic matrices~\citep{seneta2006non}.
This convergence regularises the gain at very large depths, preventing unbounded growth and producing the observed decline at small $l$ values.

The peak backward gradient gains are approximately $\bm{1.27}$ (Tiny, 8 sub-layers), $\bm{1.39}$ (Small, 12 sub-layers), and $\bm{1.58}$ (Medium, 24 sub-layers), with the deeper model exhibiting a higher but still modest peak.
These values are consistent with the trend reported by~\citet{xie2025mhc} for large-scale language models ($\approx 1.6$ at 54 sub-layers in the 27B model), scaled to the shallower architectures used in this work.
Crucially, the composite gain remains bounded below $2.0$ in all cases, in stark contrast to unconstrained hyper-connections where the composite gain can reach orders of magnitude of $10^2$--$10^3$ even at moderate depths~\citep{xie2025mhc}, leading to the training instabilities observed in DPOT.

\paragraph{Implications for PDE pre-training.}
The combination of exact forward-signal preservation and tightly bounded backward-gradient gain has direct practical consequences for multi-physics pre-training:
\begin{enumerate}
    \item \textbf{No gradient explosion across heterogeneous PDEs.}
    Different PDE families produce hidden states with vastly different magnitudes and spectral properties.
    The doubly stochastic constraint ensures that the residual mixing operation never amplifies these differences, allowing stable co-training on diverse datasets without manual loss scaling or gradient clipping adjustments.
    \item \textbf{Consistent learning dynamics across depths.}
    The compositional closure property guarantees that adding more layers does not degrade the signal-to-noise ratio of the gradient.
    This enables scaling the model from 4 blocks (Tiny) to 12 blocks (Medium) without re-tuning the learning rate or introducing additional stabilisation techniques.
    \item \textbf{Robust autoregressive rollout.}
    In the autoregressive setting (Section~\ref{sec:experiments}), the model's output at step $t$ becomes the input at step $t+1$.
    Any systematic signal amplification within the network would compound across rollout steps, leading to rapid error accumulation.
    The bounded propagation gain of the AOT block helps mitigate this effect, contributing to the substantially lower long-trajectory error observed in Table~\ref{tb-rollout}.
\end{enumerate}

\subsection{Visualization of Predicted PDEs}\label{app:visualization}

To complement the quantitative results in Section~\ref{sec:experiments}, we
provide qualitative visualizations of the zero-shot auto-regressive predictions
produced by \method at three model scales---Tiny, Small, and Medium---on all 12 pre-training datasets. Each figure follows a unified layout:
the top row shows the ground truth (GT), rows~2--4 show predictions from
\method-Ti, \method-S, and \method-M, and rows~5--7 show the corresponding
pointwise absolute error maps $|\hat{u} - u|$. The GT and prediction rows share
a common colorbar determined by the GT value range, while the error rows share a
separate colorbar scaled to the maximum error across all three model scales,
enabling direct visual comparison. All predictions are generated via
auto-regressive rollout from $T=10$ input frames without any fine-tuning.


\begin{figure}[h]
    \centering
    \includegraphics[width=\textwidth]{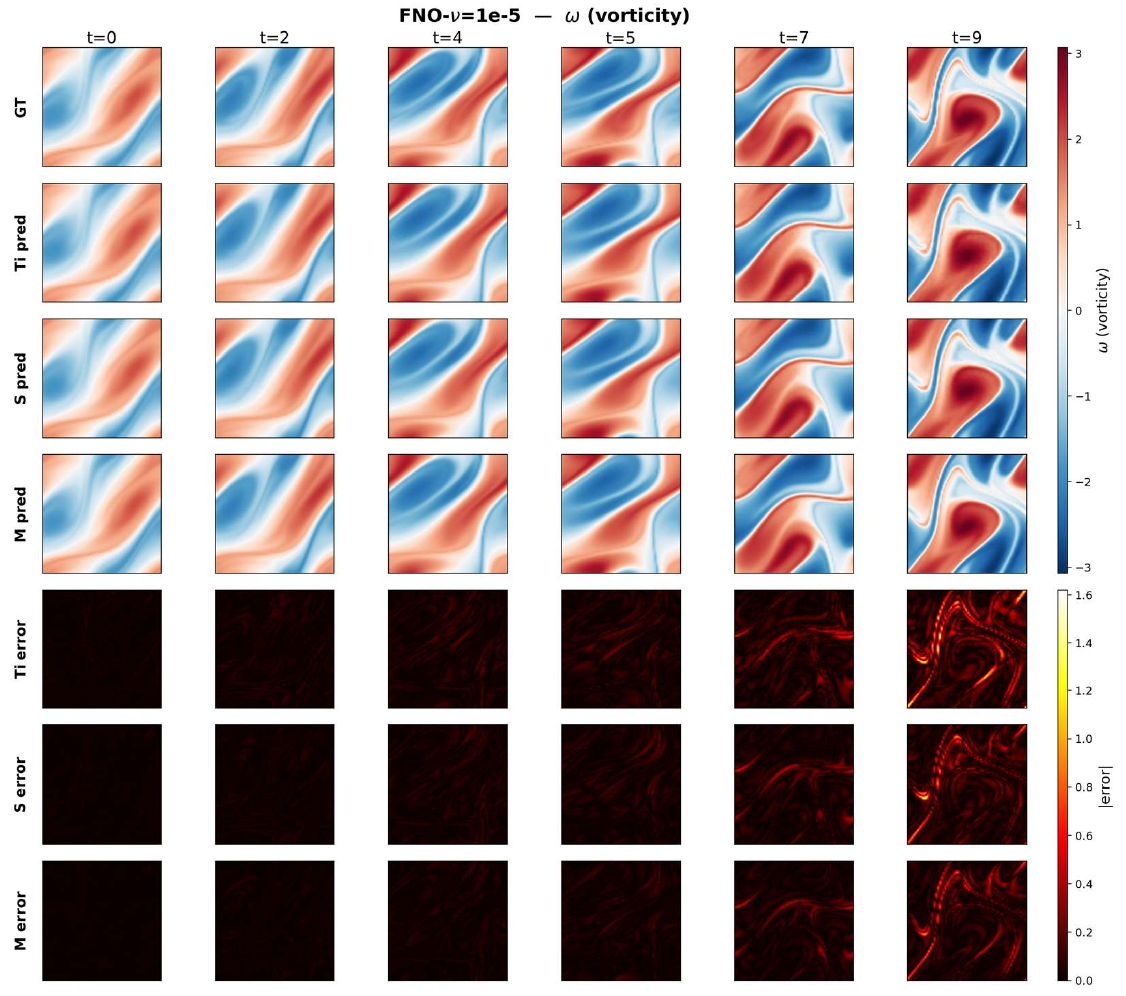}
    \caption{\textbf{FNO-$\nu$=1e-5: vorticity $\omega$.}
    Auto-regressive predictions over 10 timesteps. Rows~1--4: ground truth and
    predictions from \method-Ti/S/M. Rows~5--7: pointwise absolute error.
    Errors concentrate at vortex filaments and grow with rollout length, with
    \method-M achieving the lowest error throughout.}
    \label{fig:vis-fno-1e-5}
\end{figure}

\begin{figure}[h]
    \centering
    \includegraphics[width=\textwidth]{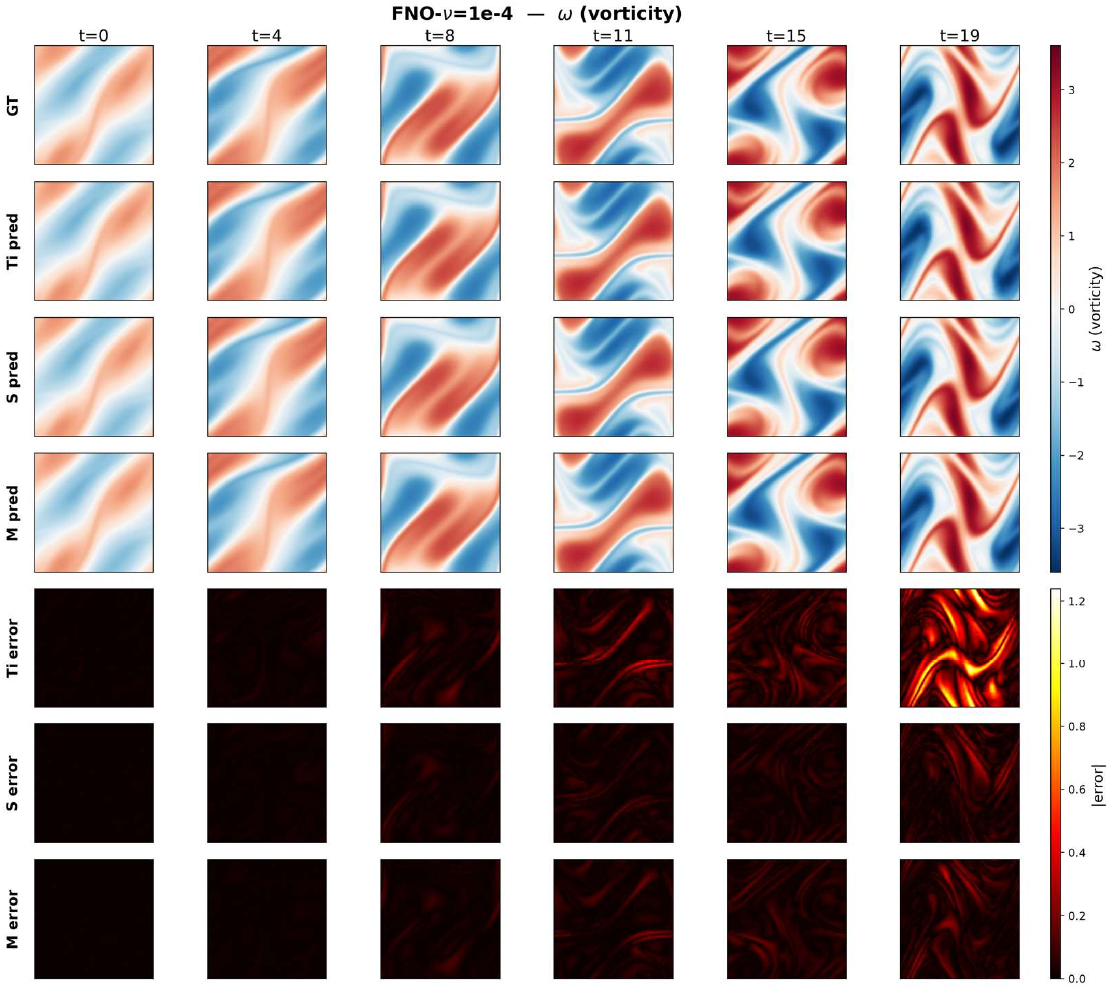}
    \caption{\textbf{FNO-$\nu$=1e-4: vorticity $\omega$.}
    Auto-regressive predictions over 20 timesteps. The longer rollout horizon
    amplifies inter-scale differences: \method-M maintains vortex sharpness
    through the full trajectory while \method-Ti shows progressive blurring of
    fine-scale features.}
    \label{fig:vis-fno-1e-4}
\end{figure}

\begin{figure}[h]
    \centering
    \includegraphics[width=\textwidth]{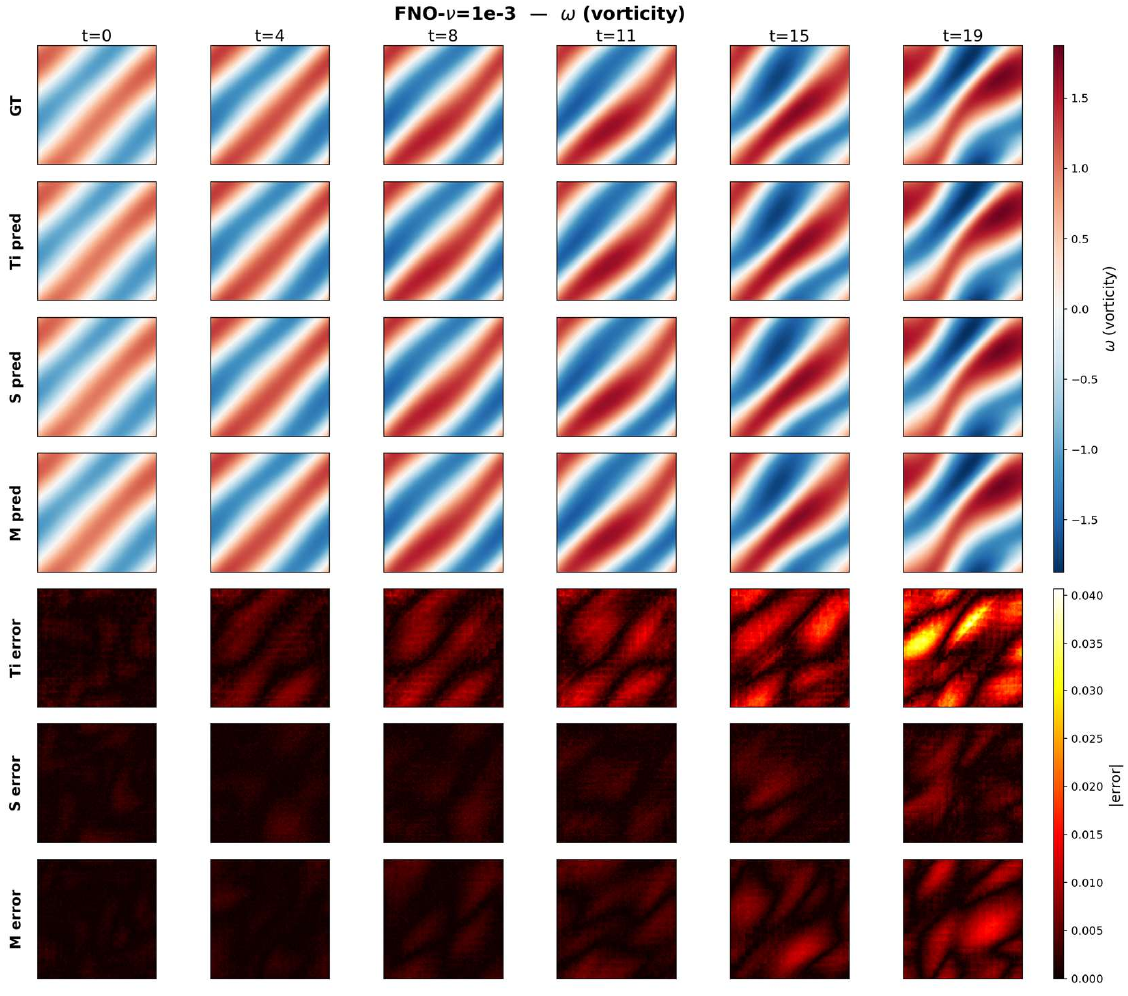}
    \caption{\textbf{FNO-$\nu$=1e-3: vorticity $\omega$.}
    Auto-regressive predictions over 20 timesteps. High viscosity yields smooth
    dynamics that all three model scales predict with near-zero error,
    confirming efficient capacity allocation in the AOT architecture.}
    \label{fig:vis-fno-1e-3}
\end{figure}


\begin{figure}[h]
    \centering
    \includegraphics[width=\textwidth]{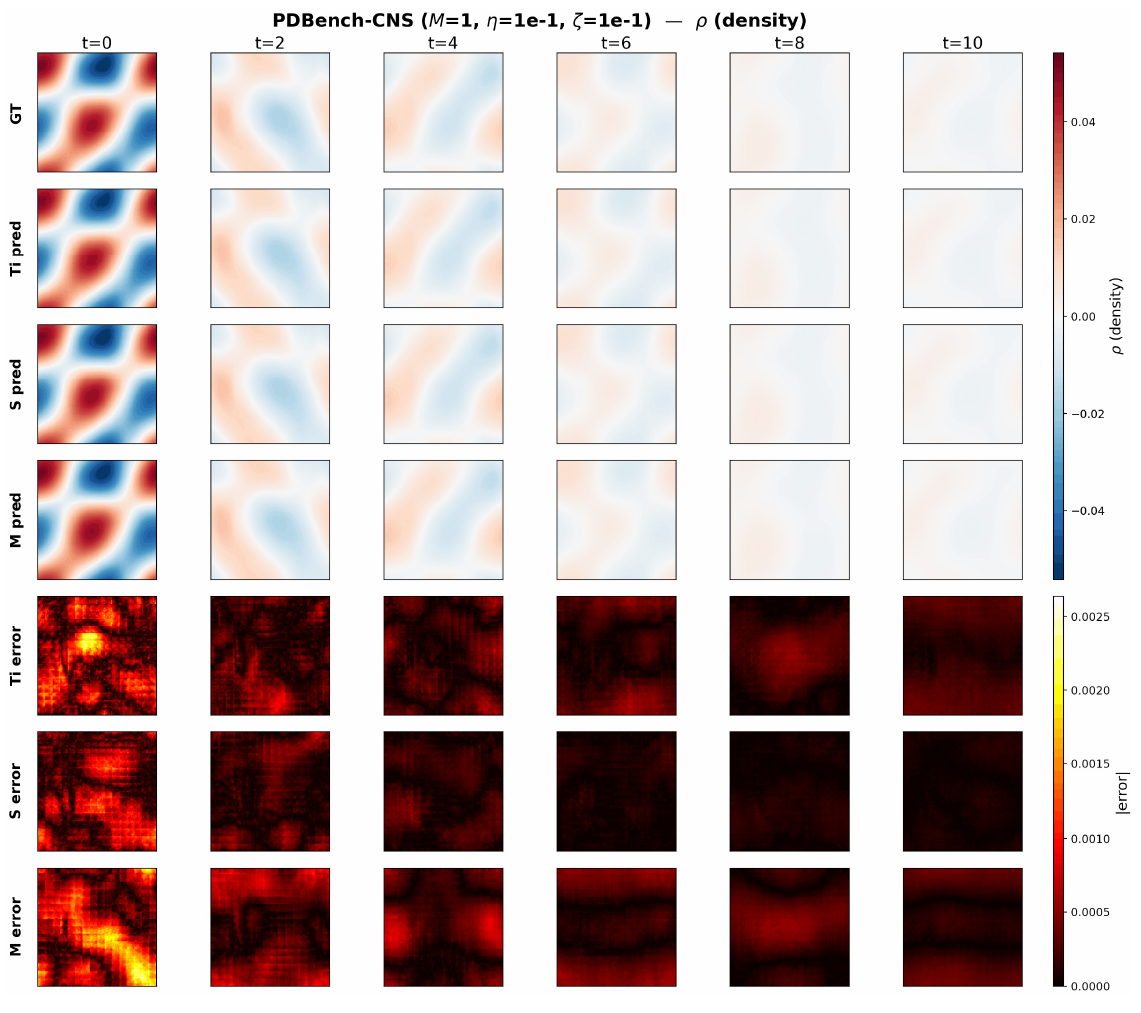}
    \caption{\textbf{PDBench-CNS ($M$=1, $\eta$=1e-1, $\zeta$=1e-1): density $\rho$.}
    Predictions over 11 timesteps. The initial perturbation decays rapidly under
    high viscosity. All three scales achieve very low error ($< 0.003$), with
    minimal inter-scale differences on this relatively smooth dynamical regime.}
    \label{fig:vis-pdb-M1-eta1}
\end{figure}

\begin{figure}[h]
    \centering
    \includegraphics[width=\textwidth]{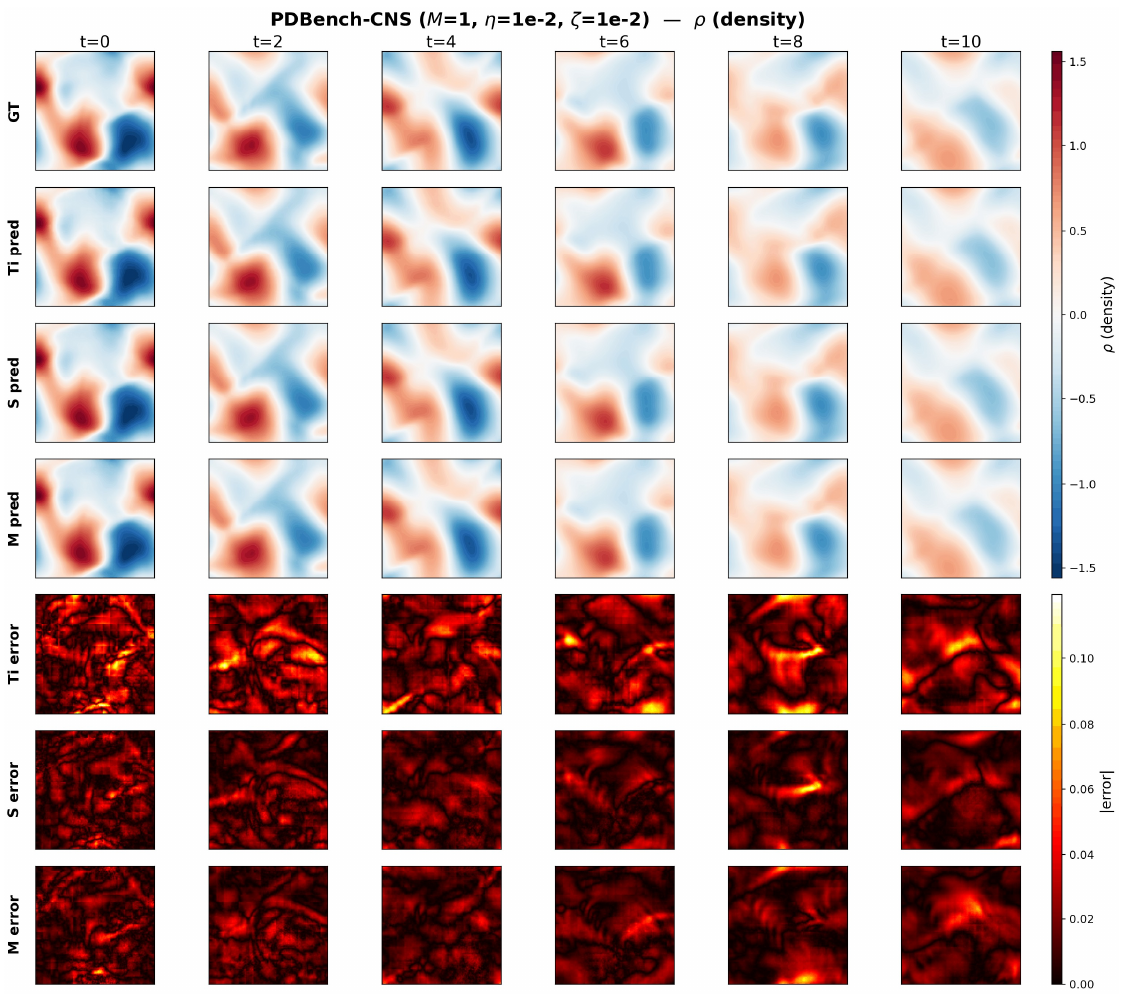}
    \caption{\textbf{PDBench-CNS ($M$=1, $\eta$=1e-2, $\zeta$=1e-2): density $\rho$.}
    Low viscosity sustains persistent density structures with high gradients.
    \method-Ti accumulates visible high-frequency artifacts near shock regions,
    while \method-M maintains the cleanest predictions throughout.}
    \label{fig:vis-pdb-M1-eta2-rho}
\end{figure}

\begin{figure}[h]
    \centering
    \includegraphics[width=\textwidth]{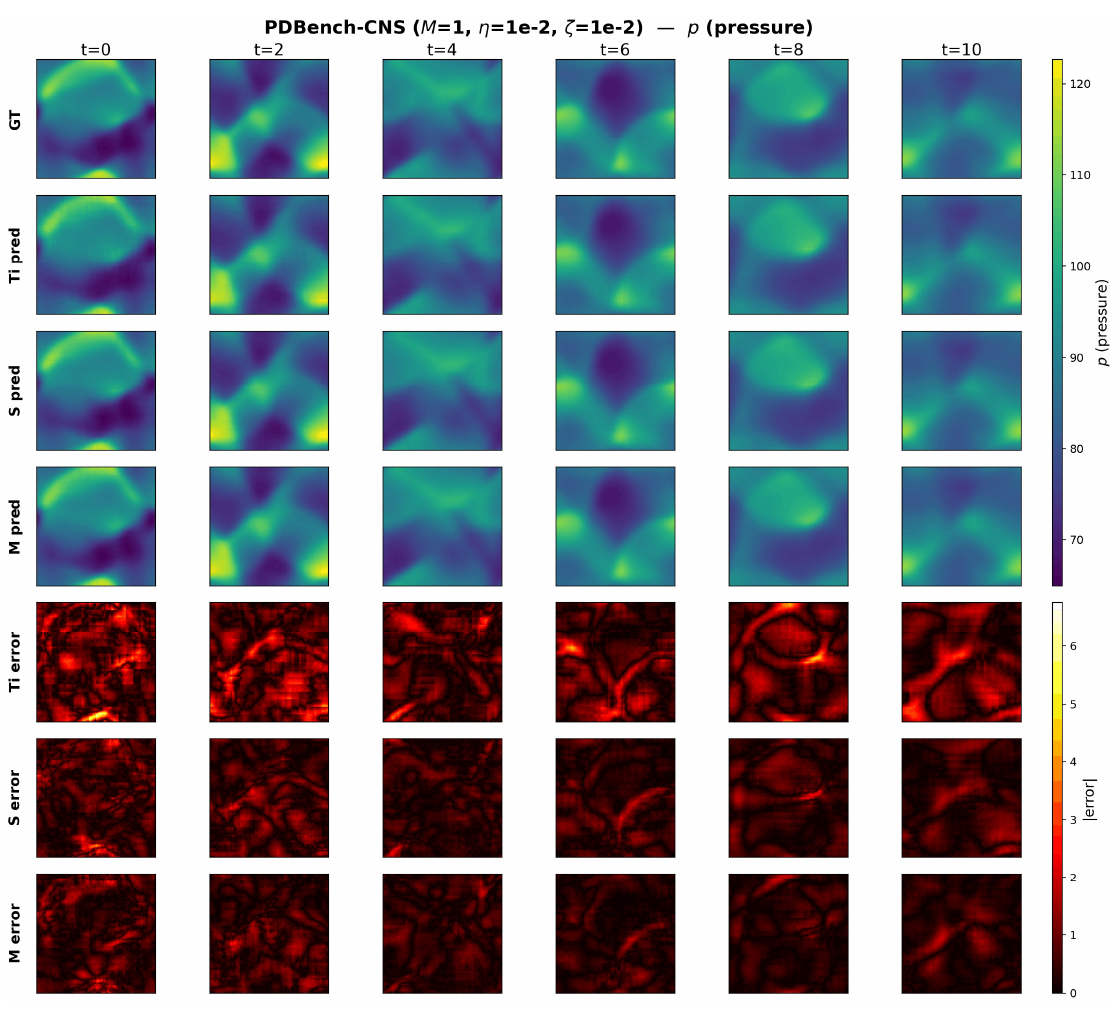}
    \caption{\textbf{PDBench-CNS ($M$=1, $\eta$=1e-2, $\zeta$=1e-2): pressure $p$.}
    The pressure field ranges from $\sim$65 to $\sim$125. Error patterns
    co-localize with shock interfaces and compression regions. \method-M
    achieves the most uniform and lowest-magnitude error distribution.}
    \label{fig:vis-pdb-M1-eta2-p}
\end{figure}

\begin{figure}[h]
    \centering
    \includegraphics[width=\textwidth]{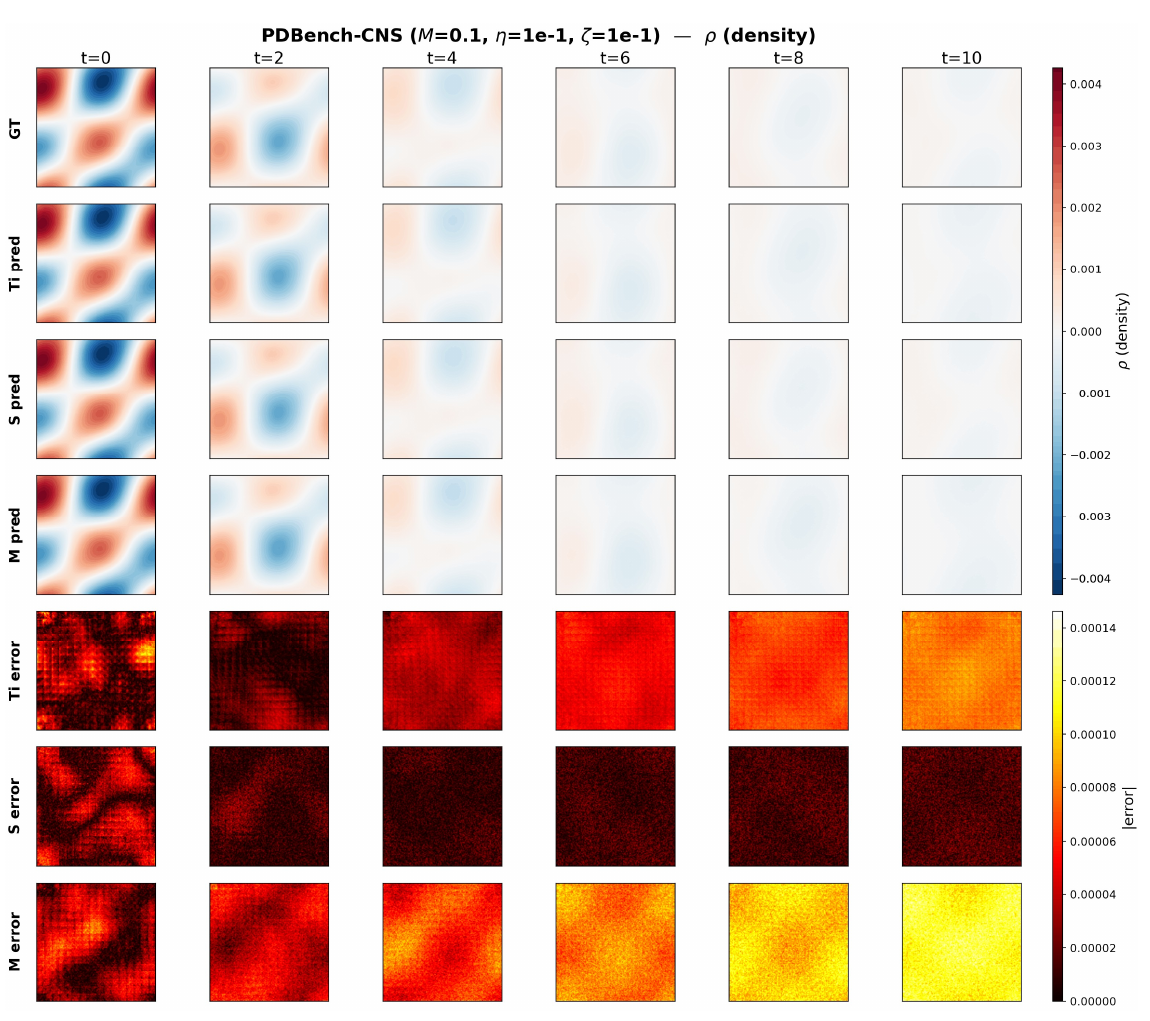}
    \caption{\textbf{PDBench-CNS ($M$=0.1, $\eta$=1e-1, $\zeta$=1e-1): density $\rho$.}
    Near-incompressible flow with rapid decay. All three scales achieve
    errors on the order of $10^{-4}$, the lowest among all CNS configurations.}
    \label{fig:vis-pdb-M01-eta1}
\end{figure}

\begin{figure}[h]
    \centering
    \includegraphics[width=\textwidth]{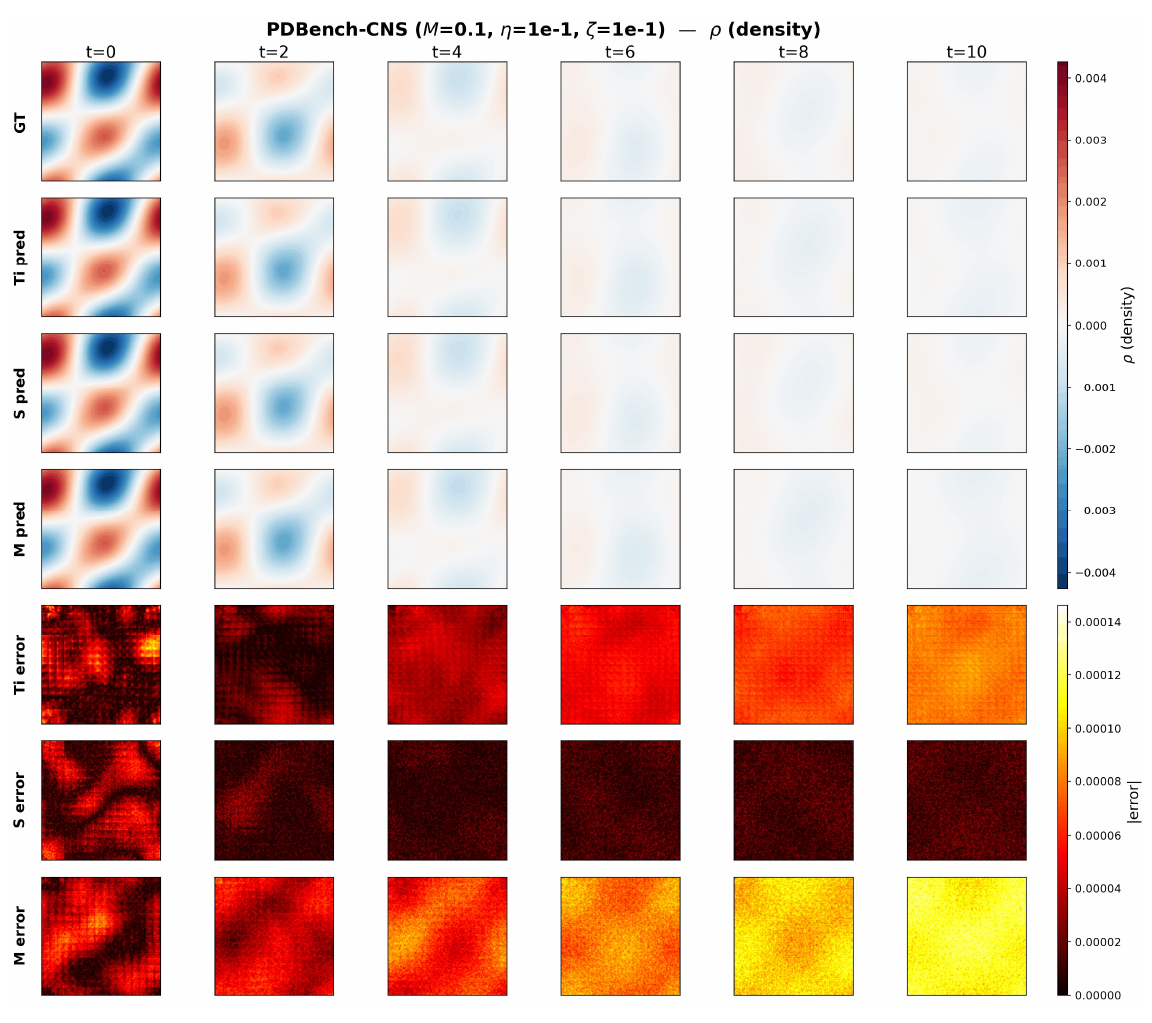}
    \caption{\textbf{PDBench-CNS ($M$=0.1, $\eta$=1e-2, $\zeta$=1e-2): density $\rho$.}
    Subsonic low-viscosity flow sustains moderate-amplitude vortex structures.
    Errors concentrate at interaction regions between density perturbations,
    with a clear Ti $>$ S $>$ M hierarchy.}
    \label{fig:vis-pdb-M01-eta2}
\end{figure}


\begin{figure}[h]
    \centering
    \includegraphics[width=\textwidth]{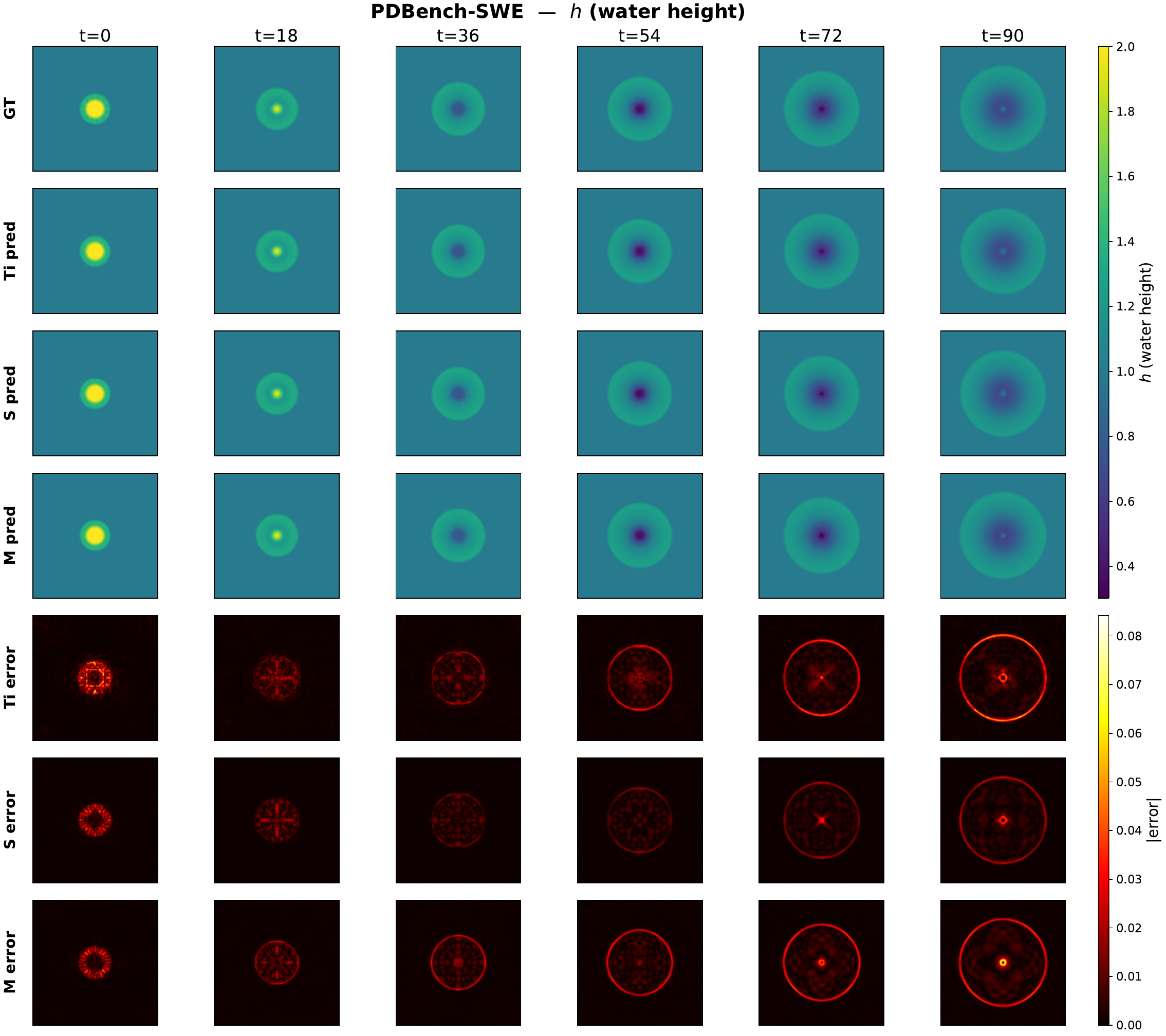}
    \caption{\textbf{PDBench-SWE: water height $h$.}
    Predictions over 91 timesteps---the longest rollout in our benchmark. The
    radially symmetric wave propagation is faithfully reproduced by all three
    scales. Errors form a distinctive ring pattern co-localized with the
    propagating wavefront.}
    \label{fig:vis-swe}
\end{figure}


\begin{figure}[h]
    \centering
    \includegraphics[width=\textwidth]{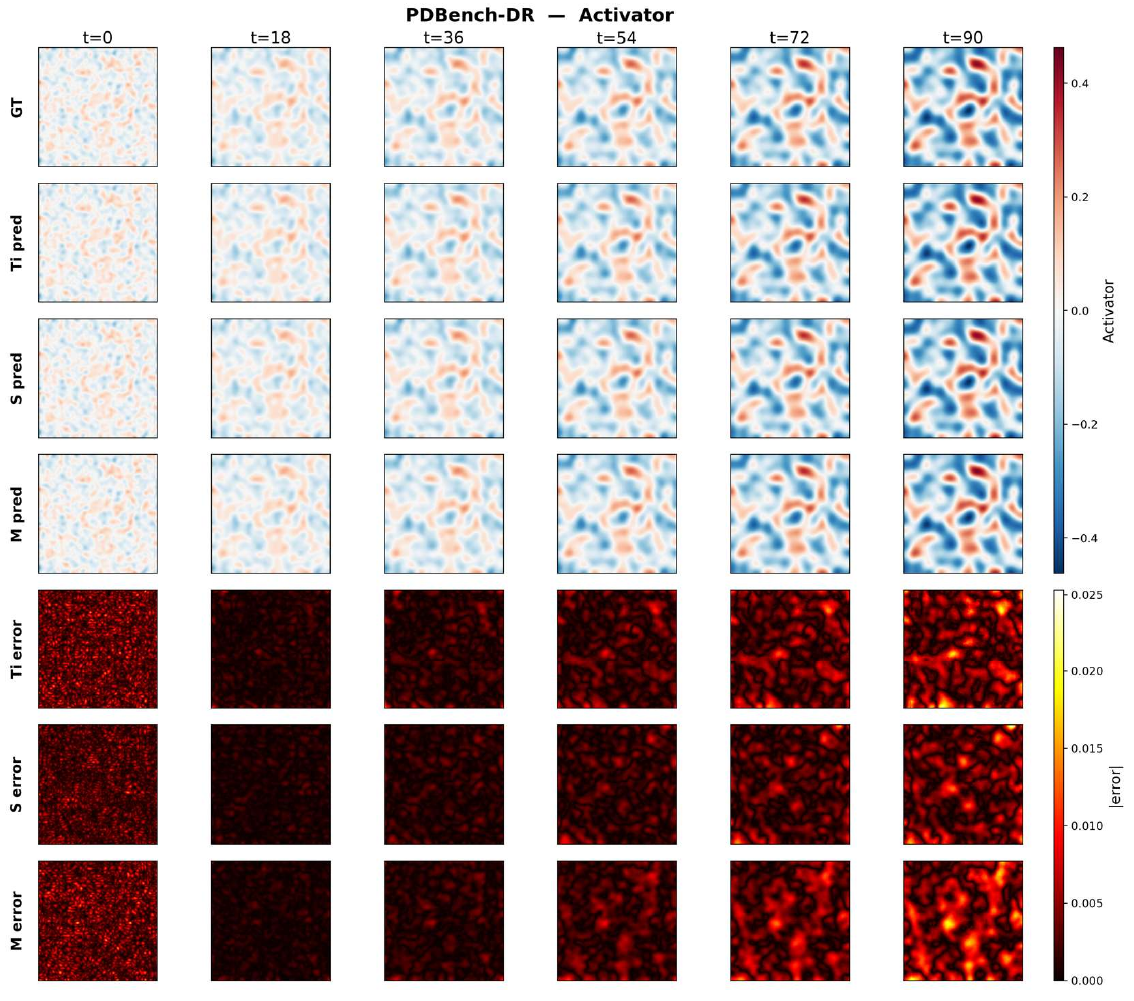}
    \caption{\textbf{PDBench-DR: activator.}
    Predictions over 91 timesteps. Turing patterns progressively emerge from a
    near-homogeneous state. Errors concentrate at pattern boundaries (spot and
    stripe edges) and grow as the patterns sharpen over time.}
    \label{fig:vis-dr-act}
\end{figure}

\begin{figure}[h]
    \centering
    \includegraphics[width=\textwidth]{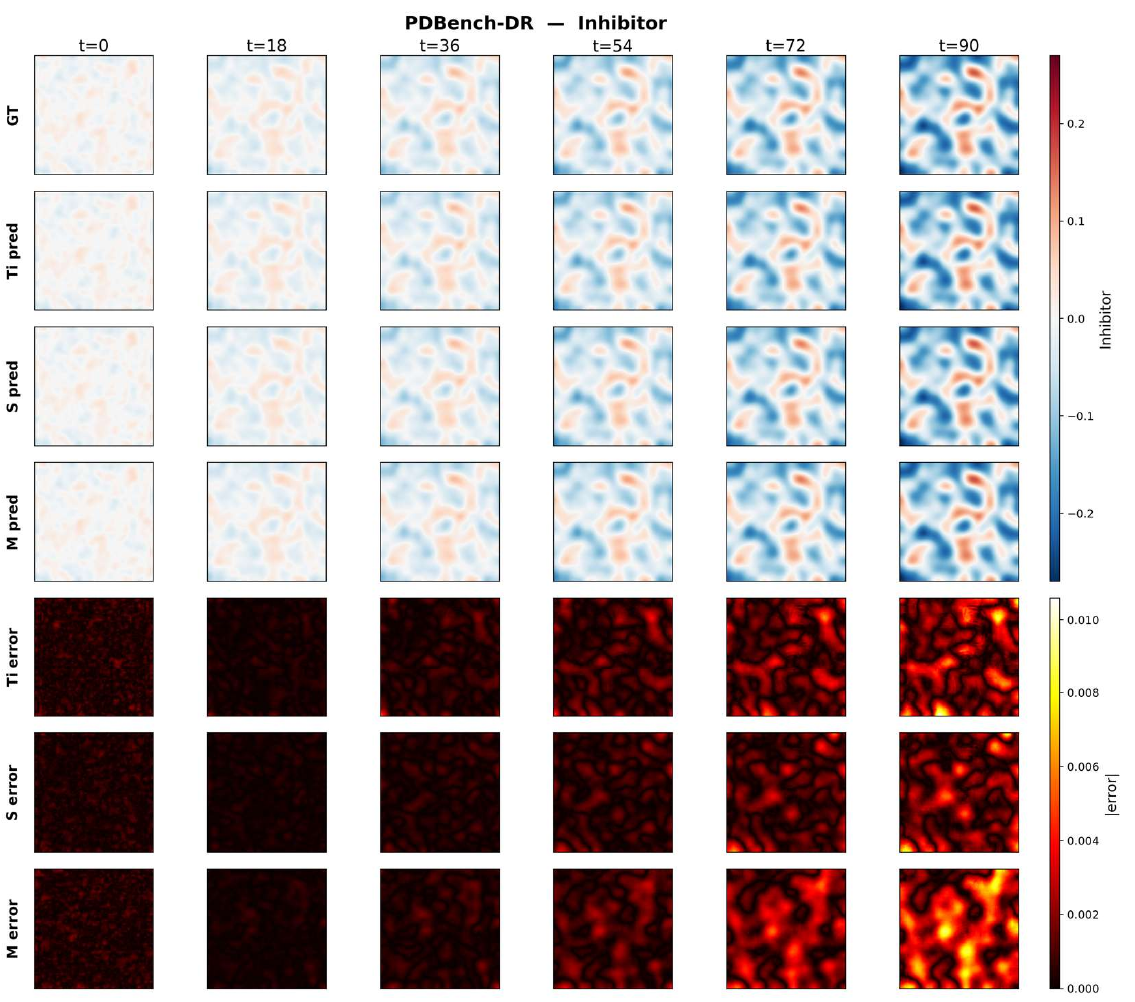}
    \caption{\textbf{PDBench-DR: inhibitor.}
    The inhibitor field shows the complementary (anti-correlated) Turing
    pattern. Error distributions mirror those of the activator channel, with
    errors localized at the interfaces between pattern domains.}
    \label{fig:vis-dr-inh}
\end{figure}


\begin{figure}[h]
    \centering
    \includegraphics[width=\textwidth]{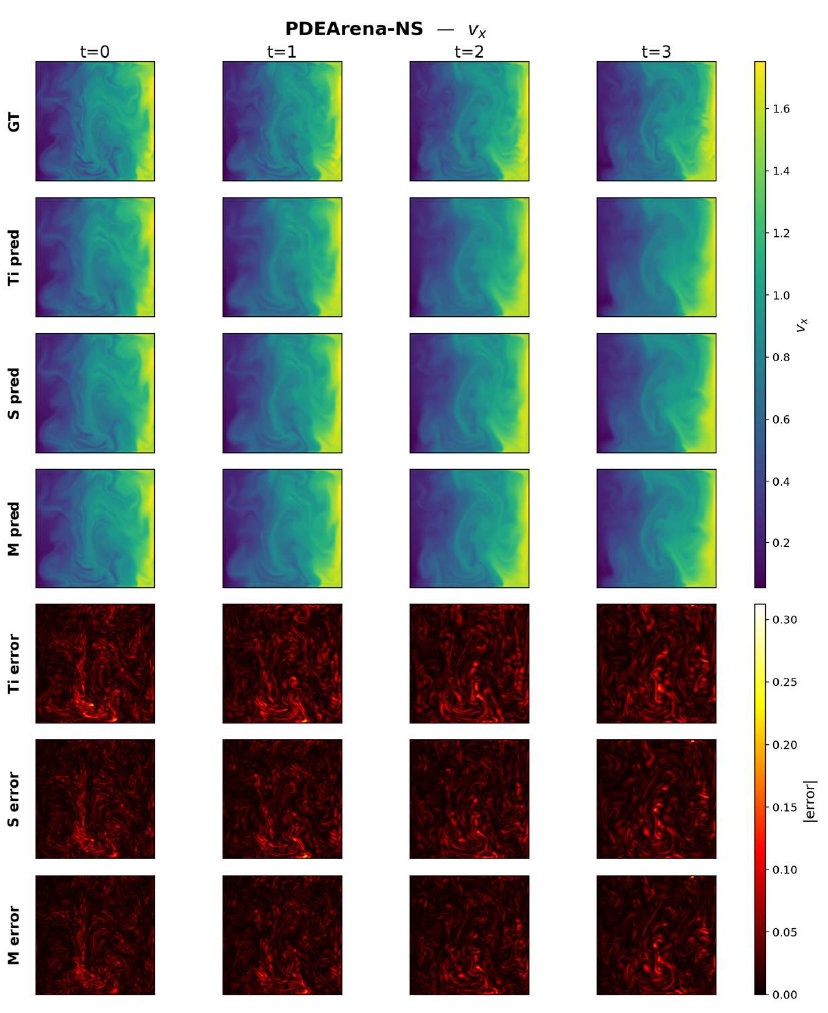}
    \caption{\textbf{PDEArena-NS: velocity $v_x$.}
    Predictions over 4 timesteps. Even on this short horizon, the error maps
    reveal a clear scaling hierarchy. Errors concentrate at shear layers and
    vortex cores. $v_y$ and $|v|$ show consistent patterns.}
    \label{fig:vis-pda}
\end{figure}


\begin{figure}[h]
    \centering
    \includegraphics[width=\textwidth]{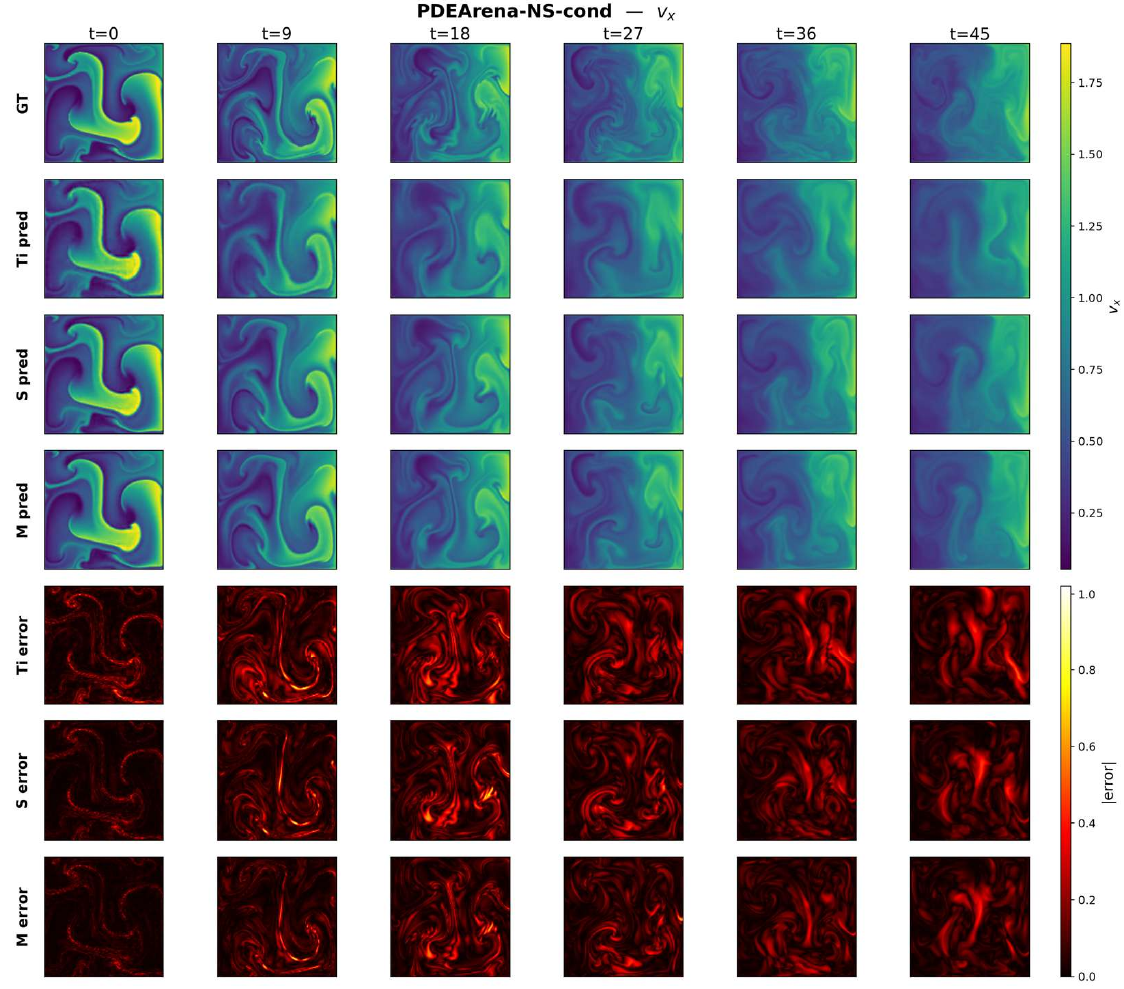}
    \caption{\textbf{PDEArena-NS-cond: velocity $v_x$.}
    Predictions over 46 timesteps. The jet-to-turbulence transition produces
    rapidly growing errors that reveal the strongest scaling hierarchy among
    all datasets: \method-M maintains coherent flow structures through late
    timesteps where \method-Ti shows substantial degradation. $v_y$ and $|v|$
    exhibit consistent trends.}
    \label{fig:vis-pda-cond}
\end{figure}


\begin{figure}[h]
    \centering
    \includegraphics[width=\textwidth]{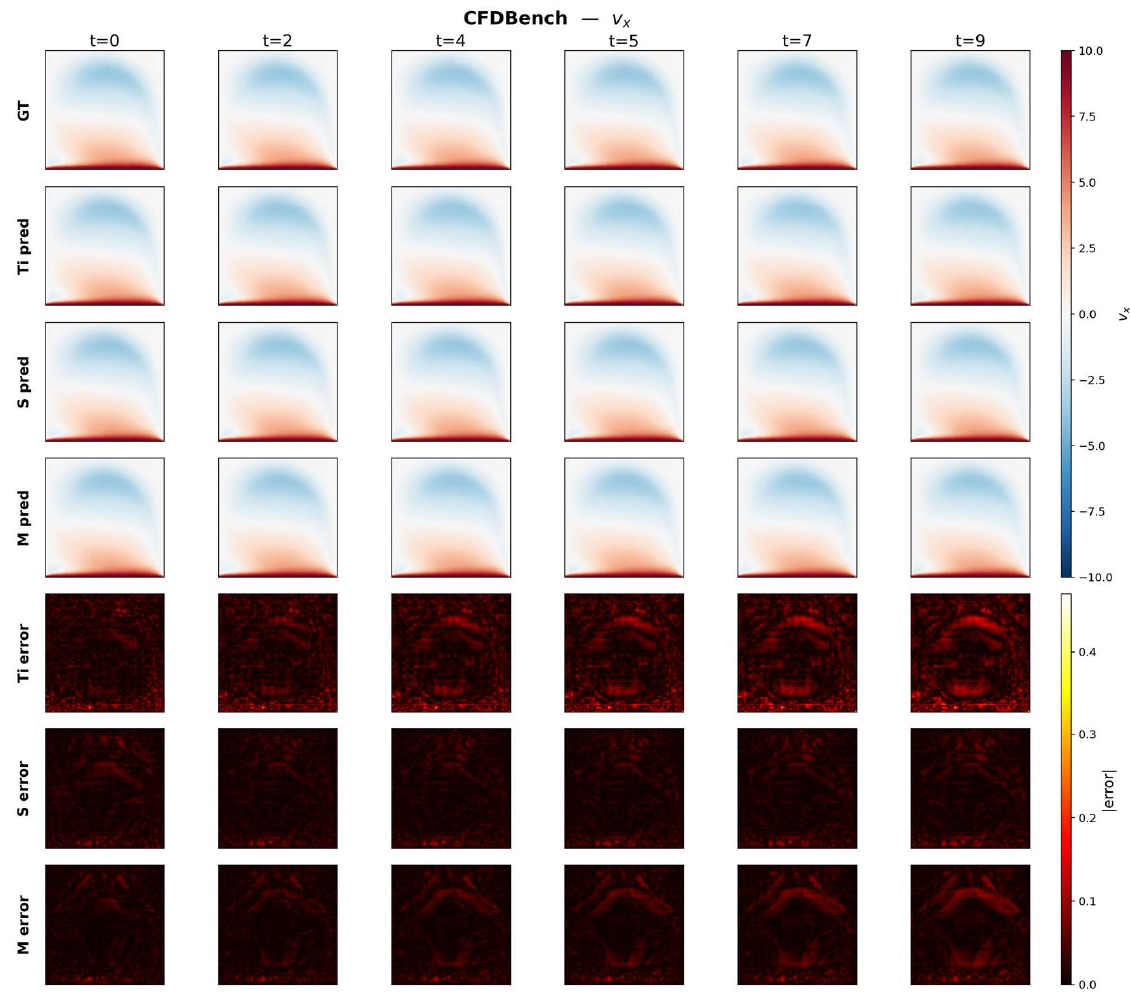}
    \caption{\textbf{CFDBench: velocity $v_x$.}
    Predictions over 10 timesteps. Errors concentrate near solid boundaries and
    flow separation regions. \method-Ti shows grid-like artifacts while
    \method-S/M produce smoother error distributions. $v_y$ shows consistent
    trends.}
    \label{fig:vis-cfdbench}
\end{figure}


\clearpage


\appendix



\end{document}